\documentclass[5p]{elsarticle}

\usepackage{url}

\usepackage{amsthm}
\usepackage{amssymb}
\usepackage{amsmath}
\usepackage{graphicx}
\usepackage[skip=8px,font=footnotesize]{caption} 
\usepackage[skip=8px,font=footnotesize]{subcaption}

\makeatletter
\renewcommand*\subcaption@label{%
  \caption@withoptargs\subcaption@@label}
\makeatother

\usepackage[table]{xcolor}
\usepackage{paralist}
\usepackage{tikz}
\usetikzlibrary{calc}
\usetikzlibrary{matrix}
\usetikzlibrary{fit}
\usepackage{accents} 
\usepackage{mdframed}

\usepackage{mathtools} 
\usepackage{pgfplots}
\usepackage{pgfplotstable}
\usepackage{multirow}
\usepackage{wrapfig}
\usepackage{xspace} 
\usepackage{pifont} 
\def\bcheckmark{\makebox[0pt][l]{$\square$}\raisebox{.15ex}{\hspace{0.1em}$\checkmark$}}
\usepackage{enumitem}
\usepackage{framed}
\usepackage{setspace}
\usepackage{textcomp}
\usepackage{multirow} 
\usepackage{placeins} 

\setlist{nosep}

\usepackage{tabularx}
\newcolumntype{Y}{>{\centering\arraybackslash}X}

\newcommand*{\dittostraight}{------------''------------}

\makeatletter
\DeclareRobustCommand\onedot{\futurelet\@let@token\@onedot}
\def\@onedot{\ifx\@let@token.\else.\null\fi\xspace}
\def\eg{{e.g}\onedot} 
\def\ie{{i.e}\onedot}

\def\etal{{et al}\onedot}

\def\NAlgorithms{28\xspace}

\def\BSDS{BSDS500\xspace}
\def\NYU{NYUV2\xspace}
\def\Fash{Fash\xspace}
\def\SBD{SBD\xspace}
\def\SUNRGBD{SUNRGBD\xspace}

\def\W{\textbf{W}\xspace}
\def\EAMS{\textbf{EAMS}\xspace}
\def\NC{\textbf{NC}\xspace}
\def\FH{\textbf{FH}\xspace}
\def\reFH{\textbf{reFH}\xspace}
\def\RW{\textbf{RW}\xspace}
\def\QS{\textbf{QS}\xspace}
\def\TP{\textbf{TP}\xspace}
\def\PF{\textbf{PF}\xspace}
\def\CIS{\textbf{CIS}\xspace}

\def\SLIC{\textbf{SLIC}\xspace}
\def\vlSLIC{\textbf{vlSLIC}\xspace}
\def\CRS{\textbf{CRS}\xspace}
\def\ERS{\textbf{ERS}\xspace}
\def\PB{\textbf{PB}\xspace}
\def\DASP{\textbf{DASP}\xspace}
\def\SEEDS{\textbf{SEEDS}\xspace}
\def\reSEEDS{\textbf{reSEEDS}\xspace}
\def\VC{\textbf{VC}\xspace}
\def\TPS{\textbf{TPS}\xspace}
\def\CCS{\textbf{CCS}\xspace}
\def\VCCS{\textbf{VCCS}\xspace}
\def\CW{\textbf{CW}\xspace}
\def\preSLIC{\textbf{preSLIC}\xspace}
\def\MSS{\textbf{MSS}\xspace}
\def\WP{\textbf{WP}\xspace}

\def\ERGC{\textbf{ERGC}\xspace}
\def\LSC{\textbf{LSC}\xspace}
\def\ETPS{\textbf{ETPS}\xspace}
\def\POISE{\textbf{POISE}\xspace}
\def\SEAW{\textbf{SEAW}\xspace}

\def\Wr{\textbf{W}\xspace}
\def\EAMSr{\textbf{EAMS}\xspace}
\def\NCr{\textbf{NC}\xspace}
\def\FHr{\textbf{FH}\xspace}
\def\reFHr{\textbf{reFH}\xspace}

\def\QSr{\textbf{QS}\xspace}
\def\TPr{\textbf{TP}\xspace}
\def\PFr{\textbf{PF}\xspace}
\def\CISr{\textbf{CIS}\xspace}
\def\SLICr{\textbf{SLIC}\xspace}
\def\vlSLICr{\textbf{vlSLIC}\xspace}
\def\CRSr{\textbf{CRS}\xspace}
\def\ERSr{\textbf{ERS}\xspace}
\def\PBr{\textbf{PB}\xspace}
\def\DASPr{\textbf{DASP}\xspace}
\def\SEEDSr{\textbf{SEEDS}\xspace}
\def\reSEEDSr{\textbf{reSEEDS}\xspace}
\def\VCr{\textbf{VC}\xspace}
\def\TPSr{\textbf{TPS}\xspace}

\def\VCCSr{\textbf{VCCS}\xspace}
\def\CWr{\textbf{CW}\xspace}
\def\preSLICr{\textbf{preSLIC}\xspace}
\def\MSSr{\textbf{MSS}\xspace}
\def\WPr{\textbf{WP}\xspace}

\def\ERGCr{\textbf{ERGC}\xspace}
\def\LSCr{\textbf{LSC}\xspace}
\def\ETPSr{\textbf{ETPS}\xspace}
\def\POISEr{\textbf{POISE}\xspace}
\def\SEAWr{\textbf{SEAW}\xspace}

\def\UEV{\text{UV}\xspace}
\def\MR{\text{MR}\xspace}
\def\AvgRec{Average\allowbreak\xspace Miss Rate\xspace}
\def\AvgUE{Average\allowbreak\xspace Undersegmentation\allowbreak\xspace Error\xspace}
\def\AvgEV{Average\allowbreak\xspace Unexplained\allowbreak\xspace Variation\xspace}
\def\uAvgRec{\underline{A}verage\allowbreak\xspace \underline{M}iss \underline{R}ate\xspace}
\def\uAvgUE{\underline{A}verage\allowbreak\xspace \underline{U}ndersegmentation\allowbreak\xspace \underline{E}rror\xspace}
\def\uAvgEV{\underline{A}verage\allowbreak\xspace \underline{U}nexplained\allowbreak\xspace \underline{V}ariation\xspace}
\def\AUE{\text{AUE}\xspace}
\def\ARec{\text{AMR}\xspace}
\def\AEV{\text{AUV}\xspace}
\def\K{\text{K}\xspace}
\DeclareRobustCommand{\Kd}{%
    \ifmmode
        K_d
    \else
        $K_d$
    \fi
}
\def\UE{\text{UE}\xspace}

\def\Rec{\text{Rec}\xspace}

\DeclareRobustCommand{\UENP}{%
    \ifmmode
        \text{UE}_{\text{NP}}
    \else
        $\text{UE}_{\text{NP}}$
    \fi
}
\DeclareRobustCommand{\UEB}{%
    \ifmmode
        \text{UE}_{\text{Bergh}}
    \else
        $\text{UE}_{\text{Bergh}}$
    \fi
}
\def\UEL{$\text{UE}_{\text{Levin}}$\xspace}
\DeclareRobustCommand{\UEL}{%
    \ifmmode
        \text{UE}_{\text{Levin}}\xspace
    \else
        $\text{UE}_{\text{Levin}}$
    \fi
}
\def\ASA{\text{ASA}\xspace}

\def\CO{\text{CO}\xspace}
\def\EV{\text{EV}\xspace}
\def\MDE{\text{MDE}\xspace}
\def\ICV{\text{ICV}\xspace}
\def\CD{\text{CD}\xspace}

\newcommand*\rot{\rotatebox{90}}

\renewcommand{\arraystretch}{1.25}

\def\fullthreeone{0.325}
\def\fullthreetwo{0.645}
\def\fullthreethree{0.985}
\def\halfthreeone{0.15}
\def\halftwoone{0.225}

\makeatletter
\def\ps@pprintTitle{%
 \let\@oddhead\@empty
 \let\@evenhead\@empty
 \def\@oddfoot{}%
 \let\@evenfoot\@oddfoot}
\makeatother

\begin{document}
\begin{frontmatter}
\title{Superpixels: An Evaluation of the State-of-the-Art}

\journal{Journal of \LaTeX\ Templates}
\author{David Stutz, Alexander Hermans, Bastian Leibe}
\address{Visual Computing Institute, RWTH Aachen University, Germany}

\begin{abstract}
    Superpixels group perceptually similar pixels to create visually meaningful
    entities while heavily reducing the number of primitives for subsequent
    processing steps. As of these properties,
    superpixel algorithms have received much attention since their naming in~2003 \cite{RenMalik:2003}.
    By today, publicly available superpixel algorithms have
    turned into standard tools in low-level vision. As such, and due to their quick
    adoption in a wide range of applications, appropriate benchmarks are crucial
    for algorithm selection and comparison. Until now, the rapidly growing number of
    algorithms as well as varying experimental setups hindered the development
    of a unifying benchmark. We present a comprehensive evaluation of 28 state-of-the-art
    superpixel algorithms utilizing a benchmark focussing on fair comparison and
    designed to provide new insights relevant for applications. To this end, we explicitly discuss
    parameter optimization and the importance of strictly enforcing connectivity.
    Furthermore, by extending well-known metrics,
    we are able to summarize algorithm performance independent of the number of
    generated superpixels, thereby overcoming a major limitation of available benchmarks.
    Furthermore, we discuss runtime, robustness against noise, blur and affine transformations,
    implementation details as well as aspects of visual quality.
    Finally, we present an overall ranking of superpixel algorithms
    which redefines the state-of-the-art and enables researchers to easily select
    appropriate algorithms and the corresponding implementations which themselves
    are made publicly available as part of our benchmark at \url{davidstutz.de/projects/superpixel-benchmark/}.
\end{abstract}
\end{frontmatter}

\pgfplotsset{
    W/.style={blue,solid,mark=diamond,mark options=solid},
    EAMS/.style={red,solid,mark=diamond,mark options=solid},
    NC/.style={yellow!80!black,solid,mark=diamond,mark options=solid},
    FH/.style={pink,solid,mark=diamond,mark options=solid},
    RW/.style={green,solid,mark=diamond,mark options=solid},
    QS/.style={black,solid,mark=diamond,mark options=solid},
    TP/.style={cyan,solid,mark=diamond,mark options=solid},
    PF/.style={violet,solid,mark=diamond,mark options=solid},
    CIS/.style={teal,solid,mark=diamond,mark options=solid},
    SLIC/.style={orange,solid,mark=diamond,mark options=solid},
    CRS/.style={brown!40!black,solid,mark=diamond,mark options=solid},
    ERS/.style={blue,dashed,mark=diamond,mark options=solid},
    PB/.style={red,dashed,mark=diamond,mark options=solid},
    DASP/.style={yellow!80!black,dashed,mark=diamond,mark options=solid},
    SEEDS/.style={green,dashed,mark=diamond,mark options=solid},
    VC/.style={pink,dashed,mark=diamond,mark options=solid},
    TPS/.style={black,dashed,mark=diamond,mark options=solid},
    VCCS/.style={cyan,dashed,mark=diamond,mark options=solid},
    CW/.style={violet,dashed,mark=diamond,mark options=solid},
    preSLIC/.style={teal,dashed,mark=diamond,mark options=solid},
    PRESLIC/.style={teal,dashed,mark=diamond,mark options=solid},
    MSS/.style={orange,dashed,mark=diamond,mark options=solid},
    WP/.style={brown!40!black,dashed,mark=diamond,mark options=solid},
    LRW/.style={blue,dotted,mark=diamond,mark options=solid},
    ERGC/.style={red,dotted,mark=diamond,mark options=solid},
    LSC/.style={yellow!80!black,dotted,mark=diamond,mark options=solid},
    ETPS/.style={green,dotted,mark=diamond,mark options=solid},
    POISE/.style={pink,dotted,mark=diamond,mark options=solid},
    SEAW/.style={black,dotted,mark=diamond,mark options=solid},
    CCS/.style={cyan,dotted,mark=diamond,mark options=solid},
    vlSLIC/.style={orange,solid,mark=x,mark options=solid,transparent},
    VLSLIC/.style={orange,solid,mark=x,mark options=solid,transparent},
    reSEEDS/.style={green,dashed,mark=x,mark options=solid,transparent},
    RESEEDS/.style={green,dashed,mark=x,mark options=solid,transparent},
    reSEEDS3D/.style={green,dashed,mark=otimes,mark options=solid,transparent},
    RESEEDS3D/.style={green,dashed,mark=otimes,mark options=solid,transparent},
    SLIC3D/.style={orange,solid,mark=otimes,mark options=solid,transparent},
    vlSLICI/.style={orange,solid,mark=x,mark options=solid},
    VLSLICI/.style={orange,solid,mark=x,mark options=solid},
    reSEEDSI/.style={green,dashed,mark=x,mark options=solid},
    RESEEDSI/.style={green,dashed,mark=x,mark options=solid},
    reFHI/.style={pink,solid,mark=x,mark options=solid},
    REFHI/.style={pink,solid,mark=x,mark options=solid},
    every tick label/.append style={
        font=\scriptsize,
    },
    every axis/.style={
        yticklabel style={
            /pgf/number format/fixed,
            /pgf/number format/precision=5
        },
        scaled y ticks=false,
        log ticks with fixed point,
		grid=both,
    },
    every axis plot/.append style={
		mark size=0.8,
    },
    every axis y label/.style={
		font=\scriptsize,
		at={(-0.175, 0.5)},
		rotate=90,
    },
    every axis x label/.style={
    	font=\scriptsize,
		at={(0.5, -0.175)},
    },
    POsuperpixelsSP/.style={
    	scaled y ticks=true,
        height=4.2cm,
        width=4.5cm,
        ymin=0,
        ymax=10400,
        x label style={
    		at={(0.5, -0.1)},
        },
        y label style={
    		at={(-0.25, 0.5)},
        },
    },
    POsuperpixelsRec/.style={
        height=4.2cm,
        width=4.5cm,
        ymin=0.55,
        ymax=1,
        x label style={
    		at={(0.5, -0.1)},
        },
        y label style={
    		at={(-0.25, 0.5)},
        },
    },
    POiterationsRec/.style={
        height=4.2cm,
        width=4.5cm,
        ymin=0.4,
        ymax=1,
        x label style={
    		at={(0.5, -0.2)},
        },
        y label style={
    		at={(-0.25, 0.5)},
        },
    },
    POiterationst/.style={
        height=4.2cm,
        width=4.5cm,
        ymin=0,
        ymax=10,
        x label style={
    		at={(0.5, -0.2)},
        },
        y label style={
    		at={(-0.25, 0.5)},
        },
    },
    POcompactnessRec/.style={
        height=4.2cm,
        width=4.5cm,
        ymin=0.45,
        ymax=1,
        x label style={
    		at={(0.5, -0.1)},
        },
        y label style={
    		at={(-0.25, 0.5)},
        },
    },
    POcompactnessCO/.style={
        height=4.2cm,
        width=4.5cm,
        ymin=0.1,
        ymax=0.7,
        x label style={
    		at={(0.5, -0.1)},
        },
        y label style={
    		at={(-0.25, 0.5)},
        },
    },
    EQBSDS500Rec/.style={
        height=5.75cm,
        width=6.5cm,
        ymin=0.5,
        ymax=1,
        xmin=180,
        xmax=8000,
        ylabel=\Rec,
        ytick={0.5,0.6,0.8,1.0},
        xtick={500, 1000, 3000, 6000},
        xlabel=\begin{tabular}{c}$\log\K$\\(a)\end{tabular},
    },
    EQBSDS500UE/.style={
        height=5.75cm,
        width=6.5cm,
        ymin=0.04,
        ymax=0.24,
        xmin=180,
        xmax=8000,
        ylabel=\UE,
        xtick={500, 1000, 3000, 6000},
        xlabel=\begin{tabular}{c}$\log\K$\\(b)\end{tabular},
        title=\BSDS,
    },
    EQBSDS500UE2/.style={
        height=5.75cm,
        width=6.5cm,
        ymin=0.04,
        ymax=0.24,
        xmin=180,
        xmax=8000,
        ylabel=\UE,
        xtick={500, 1000, 3000, 6000},
        xlabel=$\log\K$,
    },
    EQBSDS500EV/.style={
        height=5.75cm,
        width=6.5cm,
        ymin=0.7,
        ymax=0.975,
        xmin=180,
        xmax=8000,
        ylabel=\EV,
        ytick={0.7,0.8,0.9,0.975},
        xtick={500, 1000, 3000, 6000},
        xlabel=\begin{tabular}{c}$\log\K$\\(c)\end{tabular},
    },
    EQBSDS500RecMin/.style={
        height=5.75cm,
        width=6.5cm,
        ymin=0.3,
        ymax=1,
        xmin=180,
        xmax=8000,
        ylabel=$\min\Rec$,
        ytick={0.3,0.4,0.6,0.8,1.0},
        xtick={500, 1000, 3000, 6000},
        xlabel=\begin{tabular}{c}$\log\K$\\(d)\end{tabular},
    },
    EQBSDS500UEMax/.style={
        height=5.75cm,
        width=6.5cm,
        ymin=0.1,
        ymax=0.6,
        xmin=180,
        xmax=8000,
        ylabel=$\max\UE$,
        ytick={0.1,0.2,0.4,0.6},
        xtick={500, 1000, 3000, 6000},
        xlabel=\begin{tabular}{c}$\log\K$\\(e)\end{tabular},
    },
    EQBSDS500EVMin/.style={
        height=5.75cm,
        width=6.5cm,
        ymin=0.2,
        ymax=0.9,
        xmin=180,
        xmax=8000,
        ylabel=$\min\EV$,
        ytick={0.2,0.4,0.6,0.8,0.9},
        xtick={500, 1000, 3000, 6000},
        xlabel=\begin{tabular}{c}$\log\K$\\(f)\end{tabular},
    },
    EQBSDS500Avg/.style={
    	ybar,
    	grid=none,
    	ymajorgrids,
		bar width=0.1cm,
		width=19cm,
		height=3.8cm,
		ymin=0,
		ymax=35,
		ylabel=\begin{tabular}{l}\ref{plot:experiments-quantitative-bsds500-average-rec} \ARec\\\ref{plot:experiments-quantitative-bsds500-average-ue_np} \AUE\\\ref{plot:experiments-quantitative-bsds500-average-ev} \AEV\end{tabular},
		x label style={
            at={(-0.175,0.3)}
        },
        y label style={
        	rotate=270,
        	at={(0.05, 0.7)}
        },
		enlarge x limits=0.03,
		xtick=data,
		x tick label style={
			anchor=east,
			rotate=90,
		},
		title=(a) \BSDS,
    },
    EQBSDS500KMax/.style={
        x label style={
            at={(0.5,-0.3)},
        },
        height=4cm,
        width=6.5cm,
        ymin=200,
        ymax=12000,
        xmin=180,
        xmax=8000,
        ylabel=$\max \K$,
        ytick={2000,4000,6000,8000,10000,12000},
        scaled y ticks={base 10:-3},
        xlabel=\begin{tabular}{c}\K\\(j)\end{tabular},
    },
    EQBSDS500K/.style={
        ybar,
        grid=none,
        ymajorgrids,
		bar width=0.1cm,
		width=12.5cm,
		height=3cm,
		ymin=0,
		ylabel=$\text{std }\K$,
		enlarge x limits=0.03,
		xtick=data,
		x tick label style={
			anchor=east,
			rotate=90,
		},
		y label style={
			at={(-0.075, 0.5)},
		},
    },
    EQBSDS500t/.style={
        max space between ticks=20,
        height=4.5cm,
        width=4.5cm,
        ymin=0,
        ymax=200,
        xmin=180,
        xmax=4500,
        ylabel=$\log t$,
        xtick={400, 1200, 3600},
        xlabel=\K,
        title=\BSDS,
        y label style={
    		at={(-0.25, 0.5)},
        },
        x label style={
    		at={(0.5, -0.2)},
        },
    },
    EQBSDS500ItRec/.style={
        height=3.75cm,
        width=3.5cm,
        ymin=0.4,
        ymax=1,
        xmin=1,
        xmax=25,
        ylabel=\Rec,
        xlabel=iterations,
        xtick={1,3,5,10,25},
        x label style={
    		at={(0.5, -0.25)},
        },
        y label style={
    		at={(-0.4, 0.5)},
        },
    },
    EQBSDS500ItUE/.style={
        height=3.75cm,
        width=3.5cm,
        ymin=0,
        ymax=0.25,
        xmin=1,
        xmax=25,
        ylabel=\UE,
        xlabel=iterations,
        title=\BSDS,
        xtick={1,3,5,10,25},
        x label style={
    		at={(0.5, -0.25)},
        },
        y label style={
            at={(-0.4, 0.5)},
        },
    },
    EQBSDS500Itt/.style={
        height=3.75cm,
        width=3.5cm,
        ymin=0,
        ymax=8,
        xmin=1,
        xmax=25,
        ylabel=$\log t$,
        xlabel=iterations,
        xtick={1,3,5,10,25},
        x label style={
    		at={(0.5, -0.25)},
        },
        y label style={
            at={(-0.4, 0.5)},
        },
    },
    EQBSDS500RobustnessRec/.style={
    	xtick=data,
		width=3.5cm,
		height=3.75cm,
		ymin=0.3,
		ymax=1,
        ylabel=\Rec,
        x label style={
    		at={(0.5, -0.25)},
        },
        y label style={
            at={(-0.4, 0.5)},
        },
    },
    EQBSDS500RobustnessUE/.style={
    	xtick=data,
		width=3.5cm,
		height=3.75cm,
		ymin=0,
		ymax=0.3,
        ylabel=\UE,
        x label style={
    		at={(0.5, -0.25)},
        },
        y label style={
            at={(-0.4, 0.5)},
        },
    },
    EQBSDS500RobustnessK/.style={
    	xtick=data,
		width=3.5cm,
		height=3.75cm,
		ymin=0,
		ymax=10000,
        ylabel=\K,
        scaled y ticks={base 10:-3},
        x label style={
    		at={(0.5, -0.25)},
        },
        y label style={
            at={(-0.4, 0.5)},
        },
    },
    EIBSDS500Rec/.style={
        height=3.75cm,
        width=3.5cm,
        ymin=0.5,
        ymax=1,
        xmin=180,
        xmax=6500,
        ylabel=\Rec,
        xlabel=$\log\K$,
        xtick={500, 1000, 4000},
        ytick={0.5,0.6,0.7,0.8,0.9,1},
        x label style={
    		at={(0.5, -0.25)},
        },
        y label style={
            at={(-0.4, 0.5)},
        },
    },
    EIBSDS500UE/.style={
        height=3.75cm,
        width=3.5cm,
        ymin=0.04,
        ymax=0.24,
        xmin=180,
        xmax=6500,
        ylabel=\UE,
        xlabel=$\log\K$,
        title=\BSDS,
        xtick={500, 1000, 4000},
        x label style={
    		at={(0.5, -0.25)},
        },
        y label style={
            at={(-0.4, 0.5)},
        },
    },
    EQBSDS500CO/.style={
        height=4.5cm,
        width=4.5cm,
        ymin=0,
        ymax=0.65,
        xmin=180,
        xmax=8000,
        ylabel=$\CO$,
        xlabel=$\log\K$,
        title=\BSDS,
        y label style={
    		at={(-0.25, 0.5)},
        },
        x label style={
    		at={(0.5, -0.2)},
        },
    },
    EIBSDS500EV/.style={
        height=3.75cm,
        width=3.5cm,
        ymin=0.7,
        ymax=0.975,
        xmin=180,
        xmax=6500,
        ylabel=\EV,
        xlabel=\K,
    },
    EIBSDS500t/.style={
    	height=3.75cm,
        width=3.5cm,
        ymin=0,
        ymax=0.5,
        xmin=180,
        xmax=4500,
        xtick={1000},
        ylabel=$\log t$,
        xlabel=\K,
        xtick={500, 1000, 4000},
        x label style={
    		at={(0.5, -0.25)},
        },
        y label style={
            at={(-0.4, 0.5)},
        },
    },
    AEBSDS500ItRec/.style={
    	height=3cm,
        width=3cm,
        ymin=0.4,
        ymax=1,
        xmin=1,
        xmax=25,
        ylabel=\Rec,
        xlabel=\K,
    },
    AEBSDS500ItUE/.style={
    	height=3cm,
        width=3cm,
        ymin=0,
        ymax=0.25,
        xmin=1,
        xmax=25,
        ylabel=\UE,
        xlabel=\K,
    },
    AEBSDS500Itt/.style={
    	height=3cm,
        width=3cm,
        ymin=0,
        ymax=8,
        xmin=1,
        xmax=25,
        ylabel=$t$,
        xlabel=\K,
    },
    AEBSDS500RecStd/.style={
        height=5.75cm,
        width=6.5cm,
        ymin=0,
        ymax=0.1,
        xmin=180,
        xmax=8000,
        ylabel=std \Rec,
        xtick={500, 1000, 3000, 6000},
        xlabel=\begin{tabular}{c}\K\\(g)\end{tabular},
    },
    AEBSDS500UEStd/.style={
        height=5.75cm,
        width=6.5cm,
        ymin=0.01,
        ymax=0.1,
        xmin=180,
        xmax=8000,
        ylabel=std \UE,
        xtick={500, 1000, 3000, 6000},
        xlabel=\begin{tabular}{c}\K\\(h)\end{tabular},
    },
    AEBSDS500EVStd/.style={
        height=5.75cm,
        width=6.5cm,
        ymin=0.02,
        ymax=0.15,
        xmin=180,
        xmax=8000,
        ylabel=std \EV,
        ytick={0.02,0.05,0.1,0.15},
        xtick={500, 1000, 3000, 6000},
        xlabel=\begin{tabular}{c}\K\\(i)\end{tabular},
    },
    AEBSDS500UELevinMeanMax/.style={
        height=5.75cm,
        width=6.5cm,
        ymin=4,
        ymax=160,
        xmin=180,
        xmax=8000,
        ylabel=$\UEL$,
        xtick={500, 1000, 3000, 6000},
        xlabel=$\log\K$,
    },
    AEBSDS500ASAMeanMin/.style={
        height=5.75cm,
        width=6.5cm,
        ymin=0.875,
        ymax=0.975,
        xmin=180,
        xmax=8000,
        ylabel=$\ASA$,
        xtick={500, 1000, 3000, 6000},
        xlabel=$\log\K$,
        title=\BSDS,
    },
    AEBSDS500RobustnessRec/.style={
    	xtick=data,
		width=5cm,
		height=5cm,
		ymin=0.3,
		ymax=1,
        ylabel=\Rec,
    },
    AEBSDS500RobustnessUE/.style={
    	xtick=data,
		width=5cm,
		height=5cm,
		ymin=0,
		ymax=0.3,
        ylabel=\UE,
    },
    AEBSDS500RobustnessK/.style={
    	xtick=data,
		width=5cm,
		height=5cm,
		ymin=0,
		ymax=10000,
        ylabel=\K,
        scaled y ticks={base 10:-3},
    },
    EQNYUV2Rec/.style={
        height=5.75cm,
        width=6.5cm,
        ymin=0.6,
        ymax=1,
        xmin=180,
        xmax=8000,
        ylabel=\Rec,
        xtick={500, 1000, 3000, 6000},
        xlabel=\begin{tabular}{c}$\log\K$\\(a)\end{tabular},
    },
    EQNYUV2UE/.style={
        height=5.75cm,
        width=6.5cm,
        ymin=0.05,
        ymax=0.2,
        xmin=180,
        xmax=8000,
        ylabel=\UE,
        xtick={500, 1000, 3000, 6000},
        xlabel=\begin{tabular}{c}$\log\K$\\(b)\end{tabular},
        title=\NYU,
    },
    EQNYUV2UE2/.style={
        height=5.75cm,
        width=6.5cm,
        ymin=0.05,
        ymax=0.2,
        xmin=180,
        xmax=8000,
        ylabel=\UE,
        xtick={500, 1000, 3000, 6000},
        xlabel=$\log\K$,
    },
    EQNYUV2EV/.style={
        height=5.75cm,
        width=6.5cm,
        ymin=0.8,
        ymax=1,
        xmin=180,
        xmax=8000,
        ylabel=\EV,
        xtick={500, 1000, 3000, 6000},
        xlabel=\begin{tabular}{c}$\log\K$\\(c)\end{tabular},
    },
    EQNYUV2RecMin/.style={
        height=5.75cm,
        width=6.5cm,
        ymin=0.25,
        ymax=1,
        xmin=180,
        xmax=8000,
        ytick={0.25,0.4,0.6,0.8,1.0},
        ylabel=$\min\Rec$,
        xtick={500, 1000, 3000, 6000},
        xlabel=\begin{tabular}{c}$\log\K$\\(d)\end{tabular},
    },
    EQNYUV2UEMax/.style={
        height=5.75cm,
        width=6.5cm,
        ymin=0.1,
        ymax=0.6,
        xmin=180,
        xmax=8000,
        ytick={0.1,0.2,0.4,0.6},
        ylabel=$\max\UE$,
        xtick={500, 1000, 3000, 6000},
        xlabel=\begin{tabular}{c}$\log\K$\\(e)\end{tabular},
    },
    EQNYUV2EVMin/.style={
        height=5.75cm,
        width=6.5cm,
        ymin=0.3,
        ymax=1,
        xmin=180,
        xmax=8000,
        ytick={0.3,0.4,0.6,0.8,1},
        ylabel=$\min\EV$,
        xtick={500, 1000, 3000, 6000},
        xlabel=\begin{tabular}{c}$\log\K$\\(f)\end{tabular},
    },
    EQNYUV2Avg/.style={
        ybar,
        grid=none,
        ymajorgrids,
		bar width=0.1cm,
		width=19cm,
		height=3.8cm,
		ymin=0,
		ymax=30,
		ylabel=\begin{tabular}{l}\ref{plot:experiments-quantitative-nyuv2-average-rec} \ARec\\\ref{plot:experiments-quantitative-nyuv2-average-ue_np} \AUE\\\ref{plot:experiments-quantitative-nyuv2-average-ev} \AEV\end{tabular},
		y label style={
            rotate=270,
            at={(0.05, 0.7)}
        },
		enlarge x limits=0.03,
		xtick=data,
		x tick label style={
			anchor=east,
			rotate=90,
		},
		title=(b) \NYU,
    },
    EQNYUV2KMax/.style={
        x label style={
            at={(0.5,-0.3)},
        },
        height=4cm,
        width=6.5cm,
        ymin=200,
        ymax=12000,
        xmin=180,
        xmax=8000,
        ylabel=$\max\K$,
        ytick={2000,4000,6000,8000,10000,12000},
        scaled y ticks={base 10:-3},
        xlabel=\begin{tabular}{c}\K\\(j)\end{tabular},
    },
    EQNYUV2K/.style={
        ybar,
        grid=none,
        ymajorgrids,
		bar width=0.1cm,
		width=12.5cm,
		height=3cm,
		ymin=0,
		ylabel=$\text{std }\K$,
		enlarge x limits=0.03,
		xtick=data,
		x tick label style={
			anchor=east,
			rotate=90,
		},
		y label style={
			at={(-0.075, 0.5)},
		},
    },
    EQNYUV2t/.style={
        height=4.5cm,
        width=4.5cm,
        ymin=0,
        ymax=425,
        xmin=180,
        xmax=4500,
        ylabel=$\log t$,
        xtick={400, 1200, 3600},
        xlabel=\K,
        max space between ticks=20,
        title=\NYU,
        x label style={
    		at={(0.5, -0.2)},
        },
        y label style={
    		at={(-0.25, 0.5)},
        },
    },
    EQNYUV2CO/.style={
        height=4.5cm,
        width=4.5cm,
        ymin=0,
        ymax=0.65,
        xmin=180,
        xmax=8000,
        ylabel=$\CO$,
        xlabel=$\log\K$,
        title=\NYU,
        y label style={
    		at={(-0.25, 0.5)},
        },
        x label style={
    		at={(0.5, -0.2)},
        },
    },
    AENYUV2RecStd/.style={
        height=5.75cm,
        width=6.5cm,
        ymin=0,
        ymax=0.08,
        xmin=180,
        xmax=8000,
        ylabel=std \Rec,
        xtick={500, 1000, 3000, 6000},
        xlabel=\begin{tabular}{c}\K\\(g)\end{tabular},
    },
    AENYUV2UEStd/.style={
        height=5.75cm,
        width=6.5cm,
        ymin=0.01,
        ymax=0.1,
        xmin=180,
        xmax=8000,
        ytick={0.01,0.02,0.04,0.06,0.08,0.1},
        ylabel=std \UE,
        xtick={500, 1000, 3000, 6000},
        xlabel=\begin{tabular}{c}\K\\(h)\end{tabular},
    },
    AENYUV2EVStd/.style={
        height=5.75cm,
        width=6.5cm,
        ymin=0.0,
        ymax=0.1,
        xmin=180,
        xmax=8000,
        ylabel=std \EV,
        xtick={500, 1000, 3000, 6000},
        xlabel=\begin{tabular}{c}\K\\(i)\end{tabular},
    },
    AENYUV2UELevin/.style={
        height=5.75cm,
        width=6.5cm,
        ymin=0,
        ymax=10,
        xmin=180,
        xmax=8000,
        ylabel=$\UEL$,
        xtick={500, 1000, 3000, 6000},
        xlabel=\K,
    },
    AENYUV2ASA/.style={
        height=5.75cm,
        width=6.5cm,
        ymin=0.88,
        ymax=0.98,
        xmin=180,
        xmax=8000,
        ylabel=$\ASA$,
        xtick={500, 1000, 3000, 6000},
        xlabel=\K,
        title=\NYU,
    },
    AENYUV2CO/.style={
        height=6cm,
        width=6.5cm,
        ymin=0,
        ymax=0.6,
        xmin=180,
        xmax=8000,
        ylabel=$\CO$,
        xlabel=$\log\K$,
        x label style={
    		at={(0.5, -0.2)},
        },
        y label style={
    		at={(-0.25, 0.5)},
        },
    },
    EQSBDRec/.style={
        height=6cm,
        width=6.5cm,
        ymin=0.65,
        ymax=1,
        xmin=180,
        xmax=8000,
        ylabel=\Rec,
        xlabel=$\log\K$,
    },
    EQSBDUE/.style={
        height=6cm,
        width=6.5cm,
        ymin=0.04,
        ymax=0.2,
        xmin=180,
        xmax=8000,
        ylabel=\UE,
        xlabel=$\log\K$,
        title=\SBD,
    },
    EQSBDEV/.style={
        height=6cm,
        width=6.5cm,
        ymin=0.75,
        ymax=1,
        xmin=180,
        xmax=8000,
        ylabel=\EV,
        xlabel=$\log\K$,
    },
    EQSBDAvg/.style={
        ybar,
        grid=none,
        ymajorgrids,
		bar width=0.1cm,
		width=19cm,
		height=3.8cm,
		ymin=0,
		ymax=25,
		ylabel=\begin{tabular}{l}\ref{plot:experiments-quantitative-sbd-average-rec} \ARec\\\ref{plot:experiments-quantitative-sbd-average-ue_np} \AUE\\\ref{plot:experiments-quantitative-sbd-average-ev} \AEV\end{tabular},
		x label style={
            at={(-0.175,0.3)}
        },
        y label style={
        		rotate=270,
        		at={(0.05, 0.7)}
        },
		enlarge x limits=0.03,
		xtick=data,
		x tick label style={
			anchor=east,
			rotate=90,
		},
		title=(c) \SBD,
    },
    EQFashRec/.style={
        height=6cm,
        width=6.5cm,
        ymin=0.85,
        ymax=1,
        xmin=180,
        xmax=8000,
        ylabel=\Rec,
        xlabel=$\log\K$,
    },
    EQFashUE/.style={
        height=6cm,
        width=6.5cm,
        ymin=0.02,
        ymax=0.1,
        xmin=180,
        xmax=8000,
        ylabel=\UE,
        xlabel=$\log\K$,
        title=\Fash,
    },
    EQFashEV/.style={
        height=6cm,
        width=6.5cm,
        ymin=0.75,
        ymax=1,
        xmin=180,
        xmax=8000,
        ylabel=\EV,
        xlabel=$\log\K$,
    },
    EQFashAvg/.style={
        ybar,
        grid=none,
        ymajorgrids,
		bar width=0.1cm,
		width=19cm,
		height=3.8cm,
		ymin=0,
		ymax=20,
		ylabel=\begin{tabular}{l}\ref{plot:experiments-quantitative-fash-average-rec} \ARec\\\ref{plot:experiments-quantitative-fash-average-ue_np} \AUE\\\ref{plot:experiments-quantitative-fash-average-ev} \AEV\end{tabular},
		x label style={
            at={(-0.175,0.3)}
        },
        y label style={
        		rotate=270,
        		at={(0.05, 0.7)}
        },
		enlarge x limits=0.03,
		xtick=data,
		x tick label style={
			anchor=east,
			rotate=90,
		},
		title=(e) \Fash,
    },
    EQSUNRGBDRec/.style={
        height=6cm,
        width=6.5cm,
        ymin=0.6,
        ymax=1,
        xmin=180,
        xmax=8000,
        ylabel=\Rec,
        xlabel=$\log\K$,
    },
    EQSUNRGBDUE/.style={
        height=6cm,
        width=6.5cm,
        ymin=0.04,
        ymax=0.175,
        xmin=180,
        xmax=8000,
        ylabel=\UE,
        xlabel=$\log\K$,
        title=\SUNRGBD,
    },
    EQSUNRGBDEV/.style={
        height=6cm,
        width=6.5cm,
        ymin=0.8,
        ymax=1,
        xmin=180,
        xmax=8000,
        ylabel=\EV,
        xlabel=$\log\K$,
    },
    EQSUNRGBDStd/.style={
    		ybar,
		bar width=0.1cm,
		width=18cm,
		height=4cm,
		ymin=0,
		ylabel=std,
		x label style={
            at={(-0.175,0.3)}
        },
		axis x line*=bottom,
        axis y line*=left,
		enlarge x limits=0.03,
		xtick=data,
		x tick label style={
			anchor=east,
			rotate=90,
		},
    },
    EQSUNRGBDAvg/.style={
        ybar,
        grid=none,
        ymajorgrids,
		bar width=0.1cm,
		width=19cm,
		height=3.8cm,
		ymin=0,
		ymax=30,
		ylabel=\begin{tabular}{l}\ref{plot:experiments-quantitative-sunrgbd-average-rec} \ARec\\\ref{plot:experiments-quantitative-sunrgbd-average-ue_np} \AUE\\\ref{plot:experiments-quantitative-sunrgbd-average-ev} \AEV\end{tabular},
		x label style={
            at={(-0.175,0.3)}
        },
        y label style={
        		rotate=270,
        		at={(0.05, 0.7)}
        },
		enlarge x limits=0.03,
		xtick=data,
		x tick label style={
			anchor=east,
			rotate=90,
		},
		title=(d) \SUNRGBD,
    },
    AESBDCO/.style={
        height=6cm,
        width=6.5cm,
        ymin=0,
        ymax=0.7,
        xmin=180,
        xmax=8000,
        ylabel=$\CO$,
        xlabel=$\log\K$,
        title=\SBD,
    },
    AESBDt/.style={
        height=5.75cm,
        width=6.5cm,
        ymin=0,
        ymax=20,
        xmin=180,
        xmax=4500,
        ylabel=$t$,
        xlabel=\K,
        title=\SBD,
    },
    AESUNRGBDCO/.style={
        height=6cm,
        width=6.5cm,
        ymin=0,
        ymax=0.65,
        xmin=180,
        xmax=8000,
        ylabel=$\CO$,
        xlabel=$\log\K$,
        title=\SUNRGBD,
    },
    AESUNRGBDt/.style={
        height=5.75cm,
        width=6.5cm,
        ymin=0,
        ymax=30,
        xmin=180,
        xmax=4500,
        ylabel=$t$,
        xlabel=\K,
        title=\SUNRGBD,
    },
    AEFashCO/.style={
        height=6cm,
        width=6.5cm,
        ymin=0,
        ymax=0.7,
        xmin=180,
        xmax=8000,
        xlabel=$\log\K$,
        xlabel=\K,
        title=\Fash,
    },
    AEFasht/.style={
        height=5.75cm,
        width=6.5cm,
        ymin=0,
        ymax=30,
        xmin=180,
        xmax=4500,
        ylabel=$t$,
        xlabel=\K,
        title=\Fash,
    },
}

\scalebox{0}{
    \begin{tikzpicture}
        \begin{axis}[hide axis]
            \addplot [W,forget plot] coordinates{(1,1)};\label{plot:w} 
            \addplot [EAMS,forget plot] coordinates{(1,1)};\label{plot:eams} 
            \addplot [NC,forget plot] coordinates{(1,1)};\label{plot:nc} 
            \addplot [FH,forget plot] coordinates{(1,1)};\label{plot:fh} 
            \addplot [reFHI,forget plot] coordinates{(1,1)};\label{plot:refh} 
            \addplot [RW,forget plot] coordinates{(1,1)};\label{plot:rw} 
            \addplot [QS,forget plot] coordinates{(1,1)};\label{plot:qs} 
            \addplot [TP,forget plot] coordinates{(1,1)};\label{plot:tp} 
            \addplot [PF,forget plot] coordinates{(1,1)};\label{plot:pf} 
            \addplot [CIS,forget plot] coordinates{(1,1)};\label{plot:cis} 
            \addplot [SLIC,forget plot] coordinates{(1,1)};\label{plot:slic} 
            \addplot [vlSLICI,forget plot] coordinates{(1,1)};\label{plot:vlslic} 
            \addplot [CRS,forget plot] coordinates{(1,1)};\label{plot:crs} 
            \addplot [ERS,forget plot] coordinates{(1,1)};\label{plot:ers} 
            \addplot [PB,forget plot] coordinates{(1,1)};\label{plot:pb} 
            \addplot [DASP,forget plot] coordinates{(1,1)};\label{plot:dasp} 
            \addplot [SEEDS,forget plot] coordinates{(1,1)};\label{plot:seeds} 
            \addplot [reSEEDSI,forget plot] coordinates{(1,1)};\label{plot:reseeds} 
            \addplot [VC,forget plot] coordinates{(1,1)};\label{plot:vc} 
            \addplot [TPS,forget plot] coordinates{(1,1)};\label{plot:tps} 
            \addplot [CCS,forget plot] coordinates{(1,1)};\label{plot:ccs} 
            \addplot [VCCS,forget plot] coordinates{(1,1)};\label{plot:vccs} 
            \addplot [CW,forget plot] coordinates{(1,1)};\label{plot:cw} 
            \addplot [preSLIC,forget plot] coordinates{(1,1)};\label{plot:preslic} 
            \addplot [MSS,forget plot] coordinates{(1,1)};\label{plot:mss} 
            \addplot [WP,forget plot] coordinates{(1,1)};\label{plot:wp} 
            \addplot [LRW,forget plot] coordinates{(1,1)};\label{plot:lrw} 
            \addplot [ERGC,forget plot] coordinates{(1,1)};\label{plot:ergc} 
            \addplot [LSC,forget plot] coordinates{(1,1)};\label{plot:lsc} 
            \addplot [ETPS,forget plot] coordinates{(1,1)};\label{plot:etps} 
            \addplot [POISE,forget plot] coordinates{(1,1)};\label{plot:poise} 
            \addplot [SEAW,forget plot] coordinates{(1,1)};\label{plot:seaw} 
        \end{axis}
    \end{tikzpicture}
}

\section{Introduction}
\label{sec:introduction}

Introduced by Ren and Malik in 2003 \cite{RenMalik:2003}, superpixels group pixels
similar in color and other low-level properties. In this respect, superpixels address
two problems inherent to the processing of digital images~\cite{RenMalik:2003}: firstly, pixels are merely a result of discretization;
and secondly, the high number of pixels in large images prevents many algorithms from being
computationally feasible. Ren and Malik introduce superpixels as more natural entities
-- grouping pixels which perceptually belong together while heavily reducing the number of primitives for subsequent algorithms.

Superpixels have been been used in a wide range of applications -- even before the term
``superpixel'' was coined. As early as 1988, Mester and Franke \cite{MesterFranke:1988}
present segmentation results similar to superpixels. Later, in 1997, early
versions of the watershed algorithm were known to produce superpixel-like segments
\cite{MarcoteguiMeyer:1997}. In the early 2000s, Hoiem \etal \cite{HoiemEfrosHebert:2005,HoiemSteinEfrosHebert:2007}
used the segmentation algorithms of \cite{FelzenswalbHuttenlocher:2004} and \cite{Meyer:1992}
to generate oversegmentations for 3D reconstruction and occlusion boundaries.
Similarly, the normalized cuts algorithm was early adopted for oversegmentation \cite{RenMalik:2003}
and semantic segmentation \cite{GouldRodgersCohenElidanKoller:2008}.
In \cite{HoiemEfrosHebert:2005,HoiemSteinEfrosHebert:2007} and \cite{TigheLazebnik:2010},
superpixels have been used to extract meaningful features for subsequent tasks
-- extensive lists of used features are included. Since the introduction of the
first superpixel algorithms around 2009, they have been applied to many important problems in computer vision:
tracking \cite{ShuWangHuchuanLuFanYangMingHsuanYang:2011,FanYangHuchuanLuMingHsuanYang:2014},
stereo and occlusion \cite{YuhangZhangHartleyMashfordBurn:2011,YamaguchiMcAllesterUrtasun:2014},
3D-reconstruction \cite{BodisSzomoruRiemenschneiderVanGool:2015}, saliency \cite{PerazziKrahenbuhlPritchhornung:2012,HeLauLiuHuangYang:2015},
object detection \cite{ShuDehghanShah:2013,YanYuZhuLeiLi:2015} 
and object proposal detection \cite{ArbelaezPontTusetBarronMarquesMalik:2014,RantalankilaKannalaRahtu:2014},
depth recovery \cite{VanDenBerghCartonVanGool:2013}
and depth estimation \cite{LiuSalzmannHe:2014,LiuShenLin:2014}, semantic segmentation \cite{GouldRodgersCohenElidanKoller:2008,LermaKosecka:2014}, indoor scene understanding \cite{LinFidlerUrtasun:2013,GuptaArbelaezGirshickMalik:2015,GeigerWang:2015},
optical flow \cite{LuYangMinDo:2013}, scene flow \cite{MenzeGeiger:2015}, 
clothes parsing \cite{YamaguchiKiapourOrtizBerg:2012,DongChenXiahuangYan:2013} and as basis for convolutional neural networks 
\cite{GaddeJampaniKiefelGehler:2015,HeLauLiuHuangYang:2015} to name just a few.
Superpixels have also been adopted in domain specific applications such as
medical image segmentation \cite{AndresKotheHelmstaedterDenkHamprecht:2008,LucchiSmithAchantaLepetitFua:2010,LucchiSmithAchantaKnottFua:2012}
or medical image retrieval \cite{HaasDonnerBurnerHolzerLangs:2011}.
Moreover, superpixels have been found useful for dataset annotation \cite{YamaguchiKiapourOrtizBerg:2012,LiuFengDomokosXuHuangHuYan:2014}.
Finally, several superpixel algorithms (among others \cite{VanDenBerghRoigBoixManenVanGool:2013},
\cite{AchantaShajiSmithLucchiFuaSuesstrunk:2012} and \cite{GrundmannKwatraHanEssa:2010}) have
been adapted to videos and image volumes -- a survey and comparison of some of these
so-called supervoxel algorithms can be found in \cite{XuCorso:2012}.

In view of this background, most authors do not make an explicit difference between superpixel algorithms and
oversegmentation algorithms, \ie superpixel algorithms are usually compared with
oversegmentation algorithms and the terms have been used interchangeably
(\eg \cite{LevinshteinStereKutulakosFleetDickinsonSiddiqi:2009,SchickFischerStiefelhagen:2012,NeubertProtzel:2012}).
Veksler \etal \cite{VekslerBoykovMehrani:2010} distinguish superpixel algorithms
from segmentation algorithms running in ``oversegmentation mode''. More recently,
Neubert and Protzel \cite{NeubertProtzel:2013} distinguish superpixel algorithms
from oversegmentation algorithms with respect to their behavior on video sequences.
In general, it is very difficult to draw a clear line between superpixel algorithms
and oversegmentation algorithms. Several oversegmentation algorithms were not intended
to generate superpixels, nevertheless, some of them share many characteristics
with superpixel algorithms. We use the convention that superpixel
algorithms offer control over the number of generated superpixels while segmentation
algorithms running in ``oversegmentation mode'' do not. This covers the observations
made by Veksler \etal and Neubert and Protzel.

In general, most authors (\eg \cite{LevinshteinStereKutulakosFleetDickinsonSiddiqi:2009,LiuTuzelRamalingamChellappa:2011,AchantaShajiSmithLucchiFuaSuesstrunk:2012,SchickFischerStiefelhagen:2012})
agree on the following requirements for superpixels:
\begin{enumerate}[label=--,leftmargin=0.25cm,noitemsep]
    \item Partition. Superpixels should define a partition of the image,
        \ie superpixels should be disjoint and assign a label to every pixel.
    \item Connectivity. Superpixels are expected to represent connected sets of pixels.
    \item Boundary Adherence. Superpixels should preserve image boundaries.
        Here, the appropriate definition of image boundaries may depend on the application.
    \item Compactness, Regularity and Smoothness. In the absence of image boundaries,
        superpixels should be compact, placed regularly and exhibit smooth boundaries.
    \item Efficiency. Superpixels should be generated efficiently.
    \item Controllable Number of Superpixels. The number of generated superpixels should be controllable.
\end{enumerate}

Some of these requirements may be formulated implicitly, \eg Liu \etal \cite{LiuTuzelRamalingamChellappa:2011}
require that superpixels may not lower the achievable performance of subsequent processing steps.
Achanta \etal \cite{AchantaShajiSmithLucchiFuaSuesstrunk:2012} even require superpixels to
increase the performance of subsequent processing steps. Furthermore, the above requirements
should be fulfilled with as few superpixels as possible \cite{LiuTuzelRamalingamChellappa:2011}.

\textbf{Contributions.} We present an extensive evaluation of \NAlgorithms algorithms on
5 datasets regarding visual quality, performance, runtime, implementation details and
robustness to noise, blur and affine transformations.
In particular, we demonstrate the applicability of superpixel algorithms to indoor,
outdoor and person images. To ensure a fair comparison, parameters have been optimized
on separate training sets; as the number of generated superpixels heavily influences
parameter optimization, we additionally enforced connectivity. Furthermore, to evaluate
superpixel algorithms independent of the number of superpixels, we propose to integrate
over commonly used metrics such as Boundary Recall \cite{MartinFowlkesMalik:2004},
Undersegmentation Error \cite{LevinshteinStereKutulakosFleetDickinsonSiddiqi:2009, AchantaShajiSmithLucchiFuaSuesstrunk:2012, NeubertProtzel:2012}
and Explained Variation \cite{MoorePrinceWarrellMohammedJones:2008}. Finally,
we present a ranking of the superpixel algorithms considering multiple metrics and
independent of the number of generated superpixels.

\textbf{Outline.} In Section \ref{sec:related-work} we discuss important related
work regarding the comparison of superpixel algorithms and subsequently, in Section \ref{sec:algorithms},
we present the evaluated superpixel algorithms. In Section \ref{sec:datasets} we discuss
relevant datasets and introduce the used metrics in Section \ref{sec:benchmark}.
Then, Section \ref{sec:parameter-optimization} briefly discusses problems related
to parameter optimization before we present experimental results in Section \ref{sec:experiments}.
We conclude with a short summary in Section~\ref{sec:conclusion}.

\section{Related Work}
\label{sec:related-work}

Our efforts towards a comprehensive comparison of available superpixel algorithms
is motivated by the lack thereof within the literature.
Notable publications in this regard are \cite{SchickFischerStiefelhagen:2012},
\cite{AchantaShajiSmithLucchiFuaSuesstrunk:2012}, \cite{NeubertProtzel:2012}, and \cite{NeubertProtzel:2013}.
Schick \etal \cite{SchickFischerStiefelhagen:2012} introduce a metric for evaluating
the compactness of superpixels, while Achanta \etal \cite{AchantaShajiSmithLucchiFuaSuesstrunk:2012}
as well as Neubert and Protzel \cite{NeubertProtzel:2012} concentrate on using
known metrics. Furthermore, Neubert and Protzel evaluate the robustness of superpixel
algorithms with respect to affine transformations such as scaling, rotation, shear
and translation. However, they do not consider ground truth for evaluating robustness.
More recently, Neubert and Protzel \cite{NeubertProtzel:2013} used the Sintel dataset
\cite{ButlerWulffStanleyBlack:2012} to evaluate superpixel algorithms based on optical
flow in order to assess the stability of superpixel algorithms in video sequences.

Instead of relying on an application independent evaluation of superpixel algorithms, some authors
compared the use of superpixel algorithms for specific computer vision tasks.
Achanta \etal \cite{AchantaShajiSmithLucchiFuaSuesstrunk:2012} use
the approaches of \cite{GouldRodgersCohenElidanKoller:2008} and \cite{GonfausBoschVanDeWeijerBagdanovSerratGonzalez:2010}
to assess superpixel algorithms as pre-processing step for semantic segmentation.
Similarly, Strassburg \etal \cite{StrassburgGrzeszickRothackerFink:2015} evaluate
superpixel algorithms based on the semantic segmentation approach described in \cite{TigheLazebnik:2010}.
Weikersdorfer \etal \cite{WeikersdorferGossowBeetz:2012} use superpixels as basis for the normalized cuts
algorithm \cite{ShiMalik:2000} applied to classical segmentation and compare the
results with the well-known segmentation algorithm by Arbel{\'a}ez \etal \cite{ArbelaezMaireFowlkesMalik:2011}.
Koniusz and Mikolajczyk \cite{KoniuszMikolajczyk:2009}, in contrast, evaluate superpixel algorithms
for interest point extraction.

In addition to the above publications, authors of superpixel algorithms usually
compare their proposed approaches to existing superpixel algorithms. Usually, the
goal is to demonstrate superiority with regard to specific aspects. However, used
parameter settings are usually not reported, or default parameters are used, and
implementations of metrics differ. Therefore, these experiments are
not comparable across publications.

Complementing the discussion of superpixel algorithms in the literature so far, and
similar to \cite{SchickFischerStiefelhagen:2012}, \cite{AchantaShajiSmithLucchiFuaSuesstrunk:2012}
and \cite{NeubertProtzel:2012}, we concentrate on known metrics
to give a general, application independent evaluation of superpixel algorithms. However, we
consider minimum/maximum as well as standard deviation in addition to metric
averages in order assess the stability of superpixel algorithms as also considered
by Neubert and Protzel \cite{NeubertProtzel:2012,NeubertProtzel:2013}.
Furthermore, we explicitly document parameter optimization
and strictly enforce connectivity to ensure fair comparison. In contrast to \cite{NeubertProtzel:2012},
our robustness experiments additionally consider noise and blur and make use of ground truth for evaluation.
Finally, we render three well-known metrics independent of the number of generated
superpixels allowing us to present a final ranking of superpixel algorithms.

\section{Algorithms}
\label{sec:algorithms}

\newcommand{\algobox}[9]{%
    \begin{framed}
        \vskip -0.15cm
        \hskip -0.7cm
        \begin{minipage}[t]{0.25\textwidth}
            \vspace{-0.2cm}
            \def\temp{#9}\ifx\temp\empty
                \includegraphics[scale=0.425]{#1}
            \else
                \includegraphics[scale=#9]{#1}
            \fi
        \end{minipage}
        \hskip 0.15cm
        \begin{minipage}[t]{0.75\textwidth}
            {\scriptsize
                {\tiny Name}\\[-4px]
                #2\\[-2px]
                {\tiny Reference (Google Scholar Citations)\hfill Color}\\[-4px]
                #3\\[-2px]
            }
        \end{minipage}
        \vskip -0.15cm
        \hskip -0.7cm
        \begin{minipage}[t]{1.025\textwidth}
            {\tiny Implementation \hfill Superpixels Compactness Iterations}\\[-4px]
            {\scriptsize #4 \hfill #5\hskip 1.3cm#7\hskip 1.2cm#6}\\[-2px]
            \def\temp{#8}\ifx\temp\empty

            \else
                {\tiny Description}\\[-4px]
                {\scriptsize #8}
            \fi
        \end{minipage}
        \def\temp{#8}\ifx\temp\empty
            \vskip -0.5cm
        \else
            \vskip -0.2cm
        \fi
    \end{framed}
    \vspace{-0.65cm}
}

In our comparison, we aim to discuss popular algorithms with publicly available implementations
alongside less-popular and more recent algorithms for which implementations were partly provided by the authors.
To address the large number of superpixel algorithms,
we find a rough categorization of the discussed algorithms helpful.
Based on the categorization by Achanta \etal \cite{AchantaShajiSmithLucchiFuaSuesstrunk:2012}
 -- who presented (to the best of our knowledge) the first and only
categorization of superpixel algorithms -- we categorized algorithms according to their high-level approach.
We found that this categorization provides an adequate abstraction of algorithm details, allowing
to give the reader a rough understanding of the different approaches,
while being specific enough to relate categories to experimental results, as done in Section \ref{sec:experiments}.
For each algorithm, we present the used acronym, the reference and its number of citations\footnote{
    Google Scholar citations as of October 13, 2016.
}.
In addition, we provide implementation details such as the programming language, the used color space,
the number of parameters as well as whether the number of superpixels,
the compactness and the number of iterations (if applicable) are controllable.


\textbf{Watershed-based.} These algorithms are based on the waterhed algorithm (\W)
and usually differ in how the image is pre-processed and how markers are set.
The number of superpixels is determined by the number of markers, and some watershed-based
superpixel algorithms offer control over the compactness, for example \WP or~\CW.

\vspace{-0.25cm}
\algobox{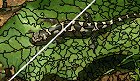}{\W\xspace-- \underline{W}atershed}{Meyer \cite{Meyer:1992}, 1992 (234)\hfill\ref{plot:w}}{C/C++; RGB; 1 Parameter}{\checkmark}{--}{--}{}{}
\algobox{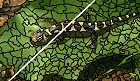}{\CW\xspace-- \underline{C}ompact \underline{W}atershed}{Neubert and Protzel \cite{NeubertProtzel:2014}, 2014 (11)\hfill\ref{plot:cw}}{C/C++; RGB; 2 Parameters}{\checkmark}{--}{\checkmark}{}{}
\algobox{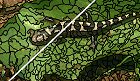}{\MSS\xspace-- \underline{M}orphological \underline{S}uperpixel \underline{S}egmentation}{Benesova and Kottman \cite{BenesovaKottman:2014}, 2014 (4)\hfill\ref{plot:mss}}{C/C++; RGB; 5 Parameters}{\checkmark}{--}{--}{}{}
\algobox{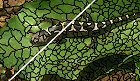}{\WP\xspace-- \underline{W}ater \underline{P}ixels}{Machairas \etal \cite{MachairasDecenciereWalter:2014,MachairesFaesselCardenasPenaChabardesWalterDecenciere:2015}, 2014 (5 + 8)\hfill\ref{plot:wp}}{Python; RGB; 2 Parameters}{\checkmark}{--}{\checkmark}{}{}
\vspace{0.5cm}

\textbf{Density-based.} Popular density-based algorithms are Edge-Augmented Mean Shift (\EAMS) and Quick Shift (\QS).
Both perform mode-seeking in a computed density image; each pixel is assigned to the corresponding
mode it falls into. Density-based algorithms usually cannot offer control over the
number of superpixels or their compactness and are, therefore, also categorized as oversegmentation algorithms.

\vspace{-0.25cm}
\algobox{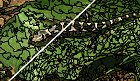}{\EAMS\xspace-- \underline{E}dge-\underline{A}ugmented \underline{M}ean \underline{S}hift}{Comaniciu and Meer \cite{ComaniciuMeer:2002}, 2002 (9631)\hfill\ref{plot:eams}}{MatLab/C; RGB; 2 Parameters}{--}{--}{--}{}{}
\algobox{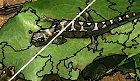}{\QS\xspace-- \underline{Q}uick \underline{S}hift}{Vedaldi and Soatto \cite{VedaldiSoatto:2008}, 2002 (376)\hfill\ref{plot:qs}}{MatLab/C; Lab; 3 Parameters}{--}{--}{--}{}{}
\vspace{0.5cm}

\textbf{Graph-based.} Graph-based algorithms treat the image as undirected graph
and partition this graph based on edge-weights which are often computed as color differences or similarities.
The algorithms differ in the partitioning algorithm, for example \FH, \ERS and \POISE
exhibit a bottom-up merging of pixels into superpixels, while \NC and \CIS use cuts
and \PB uses elimination \cite{CarrHartley:2009}.

\vspace{-0.25cm}
\algobox{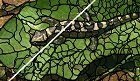}{\NC\xspace-- \underline{N}ormalized \underline{C}uts}{Ren and Malik \cite{RenMalik:2003}, 2002 (996)\hfill\ref{plot:nc}}{MatLab/C; RGB; 3 Parameters}{\checkmark}{--}{--}{}{}
\algobox{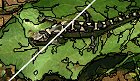}{\FH\xspace-- \underline{F}elzenswalb and \underline{H}uttenlocher}{Felzenswalb \etal \cite{FelzenswalbHuttenlocher:2004}, 2004 (4144)\hfill\ref{plot:fh}}{C/C++; RGB; 3 Parameters}{--}{--}{--}{}{}
\algobox{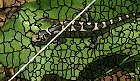}{\RW\xspace-- \underline{R}andom \underline{W}alks}{Grady \etal \cite{GradyFunkaLea:2004,Grady:2006}, 2004 (189 + 1587)\hfill\ref{plot:rw}}{MatLab/C; RGB; 2 Parameters}{\checkmark}{--}{--}{}{}
\algobox{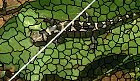}{\CIS\xspace-- \underline{C}onstant \underline{I}ntensity \underline{S}uperpixels}{Veksler \etal \cite{VekslerBoykovMehrani:2010}, 2010 (223)\hfill\ref{plot:cis}}{C/C++; Gray; 4 Parameters}{\checkmark}{\checkmark}{--}{}{}
\algobox{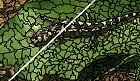}{\ERS -- \underline{E}ntropy \underline{R}ate \underline{S}uperpixels}{Liu \etal \etal \cite{LiuTuzelRamalingamChellappa:2011}, 2011 (216)\hfill\ref{plot:ers}}{C/C++; RGB; 3 Parameters}{\checkmark}{--}{--}{}{}
\algobox{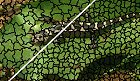}{\PB\xspace-- \underline{B}oolean Optimization Superpixels}{Zhang \etal \cite{ZhangHartleyMashfordBurn:2011}, 2011 (36)\hfill\ref{plot:pb}}{C/C++; RGB; 3 Parameters}{\checkmark}{--}{--}{}{}
\algobox{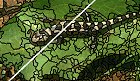}{{\renewcommand{\baselinestretch}{0.8}\POISE\xspace-- \underline{P}roposals for \underline{O}bjects from \underline{I}mproved\\\hphantom{\POISE\xspace-- }\underline{S}eeds and \underline{E}nergies}}{Humayun \etal \cite{HumayunLiRehg:2015}, 2015 (3)\hfill\ref{plot:poise}}{MatLab/C; RGB; 5 Parameters}{\checkmark}{--}{--}{}{}
\vspace{0.5cm}

\textbf{Contour evolution.} These algorithms represent superpixels as evolving
contours starting from inital seed pixels.

\vspace{-0.25cm}
\algobox{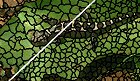}{\TP\xspace-- \underline{T}urbo \underline{P}ixels}{Levinshtein \etal \cite{LevinshteinStereKutulakosFleetDickinsonSiddiqi:2009}, 2009 (559)\hfill\ref{plot:tp}}{MatLab/C; RGB; 4 Parameters}{\checkmark}{--}{--}{}{}
\algobox{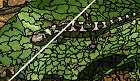}{\ERGC\xspace-- \underline{E}ikonal \underline{R}egion \underline{G}rowing \underline{C}lustering}{Buyssens \etal \cite{BuyssensGardinRuan:2014,BuyssensToutainElmoatazLezoray:2014}, 2014 (2 + 1)\hfill\ref{plot:ergc}}{C/C++; Lab; 3 Parameters}{\checkmark}{--}{\checkmark}{}{}
\vspace{0.5cm}

\textbf{Path-based.} Path-based approaches partition an image into superpixels by
connecting seed points through pixel paths following specific criteria. The number of superpixels
is easily controllable, however, compactness usually is not.
Often, these algorithms use edge information: \PF uses discrete
image gradients and \TPS uses edge detection as proposed in~\cite{DollarZitnick:2013}.

\vspace{-0.25cm}
\algobox{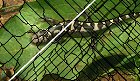}{\PF\xspace-- \underline{P}ath \underline{F}inder}{Drucker \etal \cite{DruckerMacCormick:2009}, 2009 (18)\hfill\ref{plot:pf}}{Java; RGB; 2 Parameters}{\checkmark}{--}{--}{}{}
\algobox{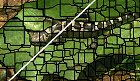}{\TPS\xspace-- \underline{T}opology \underline{P}reserving \underline{S}uperpixels}{Dai \etal \cite{DaiTangHuazhaFuXiaochunCao:2012,HuazhuFuXiaochunCaoDaiTangYahongHanDongXu:2014}, 2012 (8 + 1)\hfill\ref{plot:tps}}{MatLab/C; RGB; 4 Parameters}{\checkmark}{--}{--}{}{}
\vspace{0.5cm}

\textbf{Clustering-based.} These superpixel algorithms are inspired by clustering algorithms
such as $k$-means initialized by seed pixels and using color information, spatial
information and additional information such as depth (as for example done by \DASP).
Intuitively, the number of generated superpixels and their compactness is controllable.
Although these algorithms are iterative, post-processing is required in order to enforce connectivity.

\vspace{-0.25cm}
\algobox{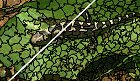}{\SLIC\xspace-- \underline{S}imple \underline{L}inear \underline{I}terative \underline{C}lustering}{Achanta \etal \cite{AchantaShajiSmithLucchiFuaSuesstrunk:2010,AchantaShajiSmithLucchiFuaSuesstrunk:2012}, 2010 (438 + 1843)\hfill\ref{plot:slic}}{C/C++; Lab; 4 Parameters}{\checkmark}{\checkmark}{\checkmark}{}{}
\algobox{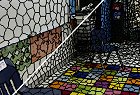}{\DASP\xspace-- \underline{D}epth-\underline{A}daptive \underline{S}uperpixels}{Weikersdorfer \etal \cite{WeikersdorferGossowBeetz:2012}, 2012 (22)\hfill\ref{plot:dasp}}{C/C++; RGB\textbf{D}; 5 Parameters}{\checkmark}{\checkmark}{\checkmark}{}{0.37}
\algobox{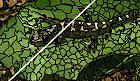}{\VC\xspace-- \underline{VC}ells}{Wang and Wang \cite{WangWang:2012}, 2012 (36)\hfill\ref{plot:vc}}{C/C++; Lab; 6 Parameters}{\checkmark}{--}{\checkmark}{}{}
\algobox{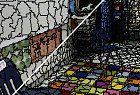}{\VCCS\xspace-- \underline{V}oxel-\underline{C}loud \underline{C}onnectivity \underline{S}egmentation}{Papon \etal \cite{PaponAbramovSchoelerWoergoetter:2013}, 2013 (87)\hfill\ref{plot:vccs}}{C/C++; RGB\textbf{D}; 4 Parameters}{--}{--}{\checkmark}{}{0.37}
\algobox{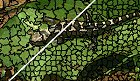}{\preSLIC\xspace-- \underline{Pre}emptive \underline{SLIC}}{Neubert and Protzel \cite{NeubertProtzel:2014}, 2014 (11)\hfill\ref{plot:preslic}}{C/C++; Lab; 4 Parameters}{\checkmark}{\checkmark}{\checkmark}{}{}
\algobox{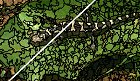}{\LSC\xspace-- \underline{L}inear \underline{S}pectral \underline{C}lustering}{Li and Chen \cite{LiChen:2015}, 2015 (2)\hfill\ref{plot:lsc}}{C/C++; Lab; 4 Parameters}{\checkmark}{\checkmark}{\checkmark}{}{}
\vspace{0.5cm}

We note that \VCCS directly operates within a point cloud and we, therefore, backproject the
generated supervoxels onto the image plane. Thus, the number of generated superpixels
is harder to control.

\textbf{Energy optimization.} These algorithms iteratively optimize a formulated
energy. The image is partitioned into a regular grid as initial superpixel segmentation,
and pixels are exchanged between neighboring superpixels with regard to the energy.
The number of superpixels is controllable, compactness can be controlled and the
iterations can usually be aborted at any point.

\vspace{-0.25cm}
\algobox{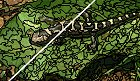}{\CRS\xspace-- \underline{C}ontour \underline{R}elaxed \underline{S}uperpixels}{Conrad \etal \cite{MesterConradGuevara:2011,ConradMertzMester:2013}, 2011 (14 + 4)\hfill\ref{plot:crs}}{C/C++; YCrCb; 4 Parameters}{\checkmark}{\checkmark}{\checkmark}{}{}
\algobox{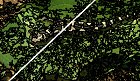}{{\renewcommand{\baselinestretch}{0.8}\SEEDS\xspace-- \underline{S}uperpixels \underline{E}xtracted via \underline{E}nergy-\\\hphantom{\SEEDS\xspace-- }\underline{D}riven \underline{S}ampling}}{Van den Bergh \etal \cite{VanDenBerghBoixRoigCapitaniVanGool:2012}, 2012 (98)\hfill\ref{plot:seeds}}{C/C++; Lab; 6 Parameters}{\checkmark}{\checkmark}{--}{}{}
\footnotetext[2]{We note that, as in \cite{VanDenBerghBoixRoigVanGool:2013}, \SEEDS actually provides a compactness parameter. But the compactness parameter is not implemented in the publicly available implementation.}
\algobox{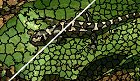}{\CCS\xspace-- \underline{C}onvexity \underline{C}onstrained \underline{S}uperpixels}{Tasli \etal \cite{TasliCiglaGeversAlatan:2013,TasliCiglaAlatan:2015}, 2013 (6 + 4)\hfill\ref{plot:ccs}}{C/C++; Lab; 3 Parameters}{\checkmark}{\checkmark}{\checkmark}{}{}
\algobox{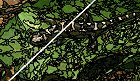}{{\renewcommand{\baselinestretch}{0.8}\ETPS\xspace-- \underline{E}xtended \underline{T}opology \underline{P}reserving\\\hphantom{\ETPS\xspace-- }\underline{S}egmentation}}{Yao \etal \cite{YaoBobenFidlerUrtasun:2015}, 2015 (6)\hfill\ref{plot:etps}}{C/C++; RGB; 5 Parameters}{\checkmark}{\checkmark}{\checkmark}{}{}
\vspace{0.65cm}

\textbf{Wavelet-based.} We found that Superpixels from Edge-Avoiding Wavelets (\SEAW) \cite{StrassburgGrzeszickRothackerFink:2015} is not yet captured in the discussed categories. In particular, it is not comparable to the algorithms discussed so far.
\algobox{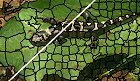}{\SEAW\xspace-- \underline{S}uperpixels from \underline{E}dge-\underline{A}voiding \underline{W}avelets}{Strassburg \etal \cite{StrassburgGrzeszickRothackerFink:2015}, 2015 (0)\hfill\ref{plot:seaw}}{MatLab/C; RGB; 3 Parameters}{\checkmark}{--}{--}{}{}

\subsection{Further Algorithms}

While the above algorithms represent a large part of the proposed superpixel algorithms, 
some algorithms are not included due to missing, unnoticed or only recently published implementations\footnote{
	Visit \url{davidstutz.de/projects/superpixel-benchmark/} to integrate your algorithm into the comparison.
}. These include \cite{RohkohlEngel:2007,
EngelSpinelloTriebelSiegwartBulthoffCurio:2009, DucournauRitalBrettoLaget:2010, ZengWangWangGanZha:2011,
PerbetStengerMaki:2012, DuShenYuWang:2012, YangGanGuiLiHou:2013, ZhangKanSchwingUrtasun:2013,
RenShakhnarovich:2013, KimZhangKangKo:2013, ShenDuWangLi:2014, MorerioMarcenaroRegazzoni:2014,
SivaWong:2014, MalladiRamRodriguez:2014, ShenDuWangLi:2014, FreifeldLiFisher:2015, Lv:2015}.

\section{Datasets}
\label{sec:datasets}

\begin{figure}[t]
	\centering
	\begin{subfigure}[b]{0.29\textwidth}
		\begin{subfigure}[t]{0.5\textwidth}%
			\includegraphics[height=1.75cm]{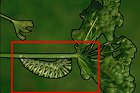}\phantomsubcaption\label{subfig:datasets-bsds500}%
		\end{subfigure}
		\begin{subfigure}[t]{0.425\textwidth}%
			\includegraphics[height=1.75cm]{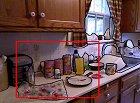}\phantomsubcaption\label{subfig:datasets-nyuv2}%
		\end{subfigure}
		\\[4px]
		\begin{subfigure}[t]{0.5\textwidth}%
			\includegraphics[height=1.75cm]{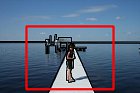}\phantomsubcaption\label{subfig:datasets-sbd}%
		\end{subfigure}
		\begin{subfigure}[t]{0.425\textwidth}%
			\includegraphics[height=1.75cm]{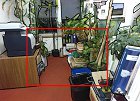}\phantomsubcaption\label{subfig:datasets-sunrgbd}%
		\end{subfigure}
	\end{subfigure}
	\hskip -3px
	\begin{subfigure}[t]{0.12\textwidth}%
		\includegraphics[height=3.68cm]{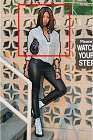}\phantomsubcaption\label{subfig:datasets-fash}%
	\end{subfigure}
	\caption{Example images from the used datasets. From left to right: \BSDS, \SBD, \NYU, \SUNRGBD,
	and \Fash. Black contours represent ground truth and red
	rectangles indicate excerpts used for qualitative comparison in Figures \ref{fig:experiments-qualitative-bsds500-sbd-fash} and \ref{fig:experiments-qualitative-nyuv2-sunrgbd}.
	\textbf{Best viewed in color.}}
	\label{fig:datasets}
\end{figure}

We chose five different datasets to evaluate superpixel algorithms: two indoor datasets,
two outdoor datasets and one person dataset. We found that these datasets
realisticly reflect the setting of common applications (\cite{YamaguchiKiapourOrtizBerg:2012,LinFidlerUrtasun:2013,
LermaKosecka:2014,LiuSalzmannHe:2014,ArbelaezPontTusetBarronMarquesMalik:2014,GeigerWang:2015,
HeLauLiuHuangYang:2015,GuptaArbelaezGirshickMalik:2015} to mention just a few 
applications on the used datasets), while leveraging the
availability of large, pixel-level annotated datasets. 
However, we also note that by focusing on natural images some application domains might not be represented well
 -- these include for example specialized research areas such as medical imaging where superpixels are also commonly used
\cite{AndresKotheHelmstaedterDenkHamprecht:2008,LucchiSmithAchantaLepetitFua:2010,HaasDonnerBurnerHolzerLangs:2011,LucchiSmithAchantaKnottFua:2012}.
Still, we believe that the experiments conducted on the
chosen datasets will aid algorithm selection in these cases, as well. Furthermore, we expect the experiments
to be useful for similar but larger datasets (such as PASCAL VOC \cite{EveringhamVanGoolWilliamsWinnZisserman:2007}, 
ImageNet \cite{DengDongSocherLiLiFeiFei:2009} or
MS COCO \cite{LinMairebelongieBourdevGirshickHaysPeronaRamananDollarZitnick:2014} to name a few prominent ones).
In addition, the selected datasets enable
us to draw a more complete picture of algorithm performance going beyond the
datasets commonly used within the literature.
Furthermore, both indoor datasets provide depth information, allowing us to evaluate superpixel algorithms
requiring depth information as additional cue. In the following we briefly discuss the
main aspects of these datasets; Figure \ref{fig:datasets} shows example
images and Table~\ref{table:datasets} summarizes key statistics.

{\BSDS \cite{ArbelaezMaireFowlkesMalik:2011}.} The Berkeley Segmentation
Dataset $500$ (\BSDS) was the first to be used for superpixel algorithm evaluation
(\eg \cite{RenMalik:2003,LevinshteinStereKutulakosFleetDickinsonSiddiqi:2009}).
It contains $500$ images and provides at least $5$ high-quality ground truth segmentations
per image. Therefore, we evaluate algorithms on all ground truth segmentations and,
for each image and a given metric, choose the ground truth segmentation resulting 
in the worst score for averaging. The images
represent simple outdoor scenes, showing landscape, buildings, animals and humans,
where foreground and background are usually easily identified. Nevertheless, natural
scenes where segment boundaries are not clearly identifiable contribute to the difficulty of the dataset.

{\SBD \cite{GouldFultonKoller:2009}.} The Stanford Background Dataset (\SBD)
combines $715$ images from several datasets \cite{RussellTorralbaMurphyFreeman:2008, Criminisi:2004, EveringhamVanGoolWilliamsWinnZisserman:2007, HoiemEfrosHebert:2007}.
As result, the dataset contains images of varying size, quality and scenes.
The images show outdoor scenes such as landscape, animals or street scenes.
In contrast to the \BSDS dataset the scenes tend to be more complex, often containing
multiple foreground objects or scenes without clearly identifiable foreground.
The semantic ground truth has been pre-processed to ensure connected segments.

{\NYU \cite{SilbermanHoiemKohliFergus:2012}.} The NYU Depth Dataset V2 (\NYU)
contains $1449$ images including pre-processed depth. Silberman \etal provide instance
labels which are used to ensure connected segments. Furthermore, following
Ren and Bo \cite{RenBo:2012}, we pre-processed the ground truth to remove small
unlabeled regions. The provided ground truth is of lower quality compared to
the \BSDS dataset. The images show varying indoor scenes of private apartments and
commercial accomodations which are often cluttered and badly lit.
The images were taken using Microsoft's Kinect.

{\SUNRGBD \cite{SongLichtenbergXiao:2015}.} The Sun RGB-D dataset (\SUNRGBD)
contains $10335$ images including pre-processed depth. The dataset combines images
from the \NYU dataset and other datasets \cite{JanochKarayevJiaBarronFritzSaenkoDarrell:2011, XiaoOwensTorralba:2013}
with newly acquired images. In contrast to the \NYU dataset, the \SUNRGBD dataset combines images
from the following devices: Intel RealSense, Asus Xtion and Microsoft Kinect v1 and v2
-- we refer to \cite{SongLichtenbergXiao:2015} for details. We removed the images
taken from the \NYU dataset. The images show cluttered indoor scenes with bad lighting taken
from private apartments as well as commercial accomodations. The provided semantic
ground truth has been pre-processed similarly to the \NYU dataset.

\begin{table}[t]
    \centering
    {\scriptsize
        \begin{tabular}{r | r | c c c c c}
            && \BSDS & \SBD & \NYU & \SUNRGBD & \Fash\\\hline
            \multirow{2}{*}{\rot{Images}} & Train & 100 & 238 & 199 & 200 & 222\\
            & Test & 200 & 477 & 399 & 400 & 463\\\hline
            \multirow{2}{*}{\rot{Size}} & Train & $481 \times 321$ & $316 \times 240$ & $608 \times 448$ & $658 \times 486$ & $400 \times 600$\\
            & Test & $481 \times 321$ & $314 \times 242$ & $608 \times 448$ & $660 \times 488$ & $400 \times 600$\\\hline
        \end{tabular}
    }
    \caption{Basic statistics of the used datasets: the total number of images,
	the number of training and test images and the size of the images (averaged per dimension).
	The number of images for the \SUNRGBD dataset excludes the images from the \NYU dataset.
	For the \NYU and \SUNRGBD datasets, training and test images have been chosen
	uniformly at random if necessary. Note that the odd numbers used for the \NYU dataset are for no
	special reason.}
    \label{table:datasets}
\end{table}

{\Fash \cite{YamaguchiKiapourOrtizBerg:2012}.} The Fashionista dataset (\Fash)
contains $685$ images which have previously been used for clothes parsing. The
images show the full body of fashion bloggers in front of various backgrounds.
Yamaguchi et al. leveraged Amazon Mechanical Turk to acquire semantic ground truth based on
pre-computed segments (\cite{YamaguchiKiapourOrtizBerg:2012} suggests that the
algorithm in \cite{ArbelaezMaireFowlkesMalik:2011} has been used). The ground truth has been
pre-processed to ensure connected segments.

\section{Benchmark}
\label{sec:benchmark}

Our benchmark aims to score the requirements for superpixels discussed in
Section \ref{sec:introduction}; in particular boundary adherence and compactness
(note that connectivity is enforced during parameter optimization, see Section \ref{subsec:parameter-optimization-connectivity}).
As these metrics inherently depend on the number of generated superpixels, we
further extend these metrics to allow the assessment of superpixel algorithms independent
of the number of generated superpixels. Therefore, let $S = \{S_j\}_{j = 1}^\K$ and 
$G = \{G_i\}$ be partitions of the same image $I: x_n \mapsto I(x_n)$, $ 1\leq n \leq N$, where 
$S$ represents a superpixel segmentation and $G$ a ground truth segmentation.

Boundary \underline{Rec}all (\Rec) \cite{MartinFowlkesMalik:2004} is the most commonly used
metric to asses boundary adherence given ground truth. Let $\text{FN}(G, S)$ and $\text{TP}(G,S)$
be the number of false negative and true positive boundary pixels in $S$
with respect to $G$. Then \Rec is defined as
\vskip -6px
\begin{align}
    \Rec(G, S) = \frac{\text{TP}(G, S)}{\text{TP}(G, S) + \text{FN}(G, S)}.
\end{align}
\vskip -4px
Overall, high \Rec represents better boundary adherence with respect to to the ground truth boundaries, \ie higher is better.
In practice, a boundary pixel in $S$ is matched to an arbitrary boundary pixel
in $G$ within a local neighborhood of size $(2r + 1) \times (2r + 1)$, with $r$ being $0.0025$ times the image diagonal
rounded to the next integer (e.g. $r = 1$ for the \BSDS dataset).

\underline{U}ndersegmentation \underline{E}rror (\UE) \cite{LevinshteinStereKutulakosFleetDickinsonSiddiqi:2009, AchantaShajiSmithLucchiFuaSuesstrunk:2012, NeubertProtzel:2012} measures the ``leakage'' of superpixels with respect to $G$ and, therefore, implicitly also measures boundary adherence.
Here, ``leakage'' refers to the overlap of superpixels with multiple, nearby ground truth segments. The original formulation by Levinshtein \etal \cite{LevinshteinStereKutulakosFleetDickinsonSiddiqi:2009} can be written as
\vskip -6px
\begin{align}
    \UEL(G, S) = \frac{1}{|G|} \sum_{G_i} \frac{\left(\sum_{S_j \cap G_i \neq \emptyset} |S_j|\right) - |G_i|}{|G_i|}\label{eq:benchmark-ue-levin}
\end{align}
\vskip -4px
where the inner term represents the ``leakage'' of superpixel $S_j$ with respect to $G$. However, some authors \cite{AchantaShajiSmithLucchiFuaSuesstrunk:2012, NeubertProtzel:2012} argue that Equation \eqref{eq:benchmark-ue-levin} penalizes superpixels overlapping only slightly with neighboring ground truth segments and is not constrained to lie in $[0, 1]$. Achanta \etal \cite{AchantaShajiSmithLucchiFuaSuesstrunk:2012} suggest to threshold the ``leakage'' term of Equation \eqref{eq:benchmark-ue-levin} and only consider those superpixels $S_j$ with a minimum overlap of $\frac{5}{100}\cdot|S_j|$. Both van den Bergh et al. \cite{VanDenBerghBoixRoigVanGool:2013} and Neubert and Protzel \cite{NeubertProtzel:2012} propose formulations not suffering from the above drawbacks. In the former,
\vskip -6px
\begin{align}
	\UEB(G, S) = \frac{1}{N} \sum_{S_j} |S_j - \arg \max_{G_i} |S_j \cap G_i||,
\end{align}
\vskip -4px
each superpixel is assigned to the ground truth segment with the largest overlap, and only the ``leakage'' with respect to other
ground truth segments is considered. Therefore, \UEB corresponds to $(1 - \ASA)$ -- with \ASA being Achievable Segmentation Accuracy as described below.
The latter,
\vskip -6px
\begin{align}
    \hspace{-0.25cm}\UENP(G, S) = \frac{1}{N} \sum_{G_i} \sum_{S_j \cap G_i \neq \emptyset} \hspace{-0.225cm} \min\{|S_j \cap G_i|, |S_j - G_i|\},
\label{eq:benchmark-ue-np}
\end{align}
\vskip -4px
is not directly equivalent to $(1 - \ASA)$, however, \UENP and \ASA are still strongly correlated as we will see later.
All formulations have in common that lower \UE refers to less ``leakage'' with respect to the ground truth, \ie lower is better.
In the following we use $\UE \equiv \UENP$.

\underline{E}xplained \underline{V}ariation (\EV) \cite{MoorePrinceWarrellMohammedJones:2008} quantifies the
quality of a superpixel segmentation without relying on ground truth.
As image boundaries tend to exhibit strong change in color and structure, \EV
assesses boundary adherence independent of human annotions. \EV is defined as
\vskip -6px
\begin{align}
    \EV(S) = \frac{\sum_{S_j} |S_j| (\mu(S_j) - \mu(I))^2}{\sum_{x_n} (I(x_n) - \mu(I))^2}
\end{align}
\vskip -4px
where $\mu(S_j)$ and $\mu(I)$ are the mean color of superpixel $S_j$ and the image $I$,
respectively. As result, \EV quantifies the variation of the image explained by the superpixels,
\ie higher is better.

\underline{Co}mpactness (\CO) \cite{SchickFischerStiefelhagen:2012} has been introduced
by Schick \etal \cite{SchickFischerStiefelhagen:2012} to evaluate the compactness of superpixels:
\vskip -6px
\begin{align}
    \CO(G, S) = \frac{1}{N} \sum_{S_j} |S_j| \frac{4\pi A(S_j)}{P(S_j)}.\label{eq:co}
\end{align}
\vskip -4px
\CO compares the area $A(S_j)$ of each superpixel $S_j$ with the
area of a circle (the most compact 2-dimensional shape) with same perimeter $P(S_j)$,
\ie higher is better.

While we will focus on \Rec, \UE, \EV and \CO, further notable metrics are briefly discussed in the following.
\underline{A}chievable \underline{S}egmentation \underline{A}ccuracy (\ASA) \cite{LiuTuzelRamalingamChellappa:2011}
quantifies the achievable accuracy for segmentation using superpixels as pre-processing step:
\vskip -6px
\begin{align}
    \ASA(G, S) = \frac{1}{N} \sum_{S_j} \max_{G_i}\{|S_j \cap G_i|\}\label{eq:benchmark-asa};
\end{align}
\vskip -4px
\underline{I}ntra-\underline{C}luster \underline{V}ariation (\ICV) \cite{BenesovaKottman:2014} computes the average variation within each superpixel:
\vskip -6px
\begin{align}
    \ICV(S) = \frac{1}{|S|} \sum_{S_j} \frac{\sqrt{\sum_{x_n \in S_j} (I(x_n) - \mu(S_j))^2}}{|S_j|};\label{eq:icv}
\end{align}
\vskip -4px
\underline{M}ean \underline{D}istance to \underline{E}dge (\MDE) \cite{BenesovaKottman:2014} refines \Rec by also considering
the distance to the nearest boundary pixel within the ground truth segmentation:
\vskip -6px
\begin{align}
    \MDE(G, S) = \frac{1}{N} \sum_{x_n \in B(G)} \text{dist}_{S}(x_n)
\end{align}
\vskip -4px
where $B(G)$ is the set of boundary pixels in $G$, and $\text{dist}_S$ is a distance transform of $S$.

\subsection{Expressiveness and Chosen Metrics}
\label{subsec:benchmark-correlation}

Due to the large number of available metrics, we examined their expressiveness
in order to systematically concentrate
on few relevant metrics. We found that \UE tends to correlate
strongly with \ASA which can be explained by Equations \eqref{eq:benchmark-ue-np}
and \eqref{eq:benchmark-asa}, respectively. In particular, simple calculation shows that
\ASA strongly resembles $(1 - \UE)$. Surprisingly, \UENP does not correlate
with \UEL suggesting that either both metrics reflect different aspects of superpixels,
or \UEL unfairly penalizes some superpixels as suggested in
\cite{AchantaShajiSmithLucchiFuaSuesstrunk:2012} and \cite{NeubertProtzel:2012}.
Unsurprisingly, \MDE correlates strongly with \Rec which can also be explained by
their respective definitions. In this sense, \MDE does not provide additional information.
Finally, \ICV does not correlate with \EV which may be attributed to the missing
normalization in Equation \eqref{eq:icv} when compared to \EV. This also results in
\ICV not begin comparable across images as the intra-cluster variation is not related to the overall
variation within the image. As of these considerations,
we concentrate on \Rec, \UE and \EV for the presented experiments. 
Details can be found in \ref{subsec:appendix-benchmark-expressiveness}.

\subsection{\AvgRec, \AvgUE and \AvgEV}
\label{subsec:benchmark-average}

As the chosen metrics inherently depend on the number of superpixels,
we seek a way of quantifying the performance with respect to \Rec, \UE and \EV 
independent of~\K and in a single plot per dataset.
In order to summarize performance over a given interval $[\K_{\min}, \K_{\max}]$,
we consider $\MR = (1 - \Rec)$, \UE and $\UEV = (1 - \EV)$. Here, the first corresponds to the
Boundary \underline{M}iss \underline{R}ate
(\MR) and the last, \underline{U}nexplained \underline{V}ariation (\UEV), 
quantifies the variation in the image not explained by the superpixels.
We use the area below these curves in $[\K_{\min}, \K_{\max}] = [200, 5200]$ 
to quantify performance independent of \K. In Section \ref{subsec:experiments-quantitative}, 
we will see that these metrics appropriately summarize the performance of superpixel algorithms.
We denote these metrics by \uAvgRec (\ARec), \uAvgUE (\AUE) and
\uAvgEV (\AEV) -- note that this refers to an average over \K.
By construction (and in contrast to \Rec and \EV), lower \ARec, \AUE and \AEV is better,
making side-by-side comparison across datasets easy.

\section{Parameter Optimization}
\label{sec:parameter-optimization}

For the sake of fair comparison, we optimized parameters on the training sets depicted in Table \ref{table:datasets}.
Unfortunately, parameter optimization is not explicitly discussed in related work (\eg \cite{SchickFischerStiefelhagen:2012, AchantaShajiSmithLucchiFuaSuesstrunk:2012, NeubertProtzel:2012, SchickFischerStiefelhagen:2012})
and used parameters are not reported in most publications. In addition, varying runtimes as well as
categorical and integer parameters render parameter optimization difficult such that
we had to rely on discrete grid search, jointly optimizing \Rec and \UE, \ie minimizing $(1 - \Rec) + \UE$.
As \Rec and \UE operate on different representations (boundary pixels and superpixel segmentations, respectively),
the additive formulation ensures that algorithms balance both metrics. For example, we observed 
that using a multiplicative formulation allows superpixel algorithms to drive $(1 - \Rec)$ towards zero while disregarding \UE.
We optimized parameters for $\K \in \{400, 1200, 3600\}$ and interpolated linearly in between (however,
we found that for many algorithms, parameters are consistent across different values of $\K$).
Optimized parameters also include compactness parameters and the number of iterations
as well as the color space. We made sure that all algorithms at least support RGB color space for fairness.
In the following, we briefly discuss the main difficulties encountered during parameter optimization, namely controlling the number
of generated superpixels and ensuring connectivity.

\subsection{Controlling the Number of Generated Superpixels}
\label{subsec:parameter-optimization-superpixels}

As discussed in Section \ref{sec:introduction}, superpixel algorithms are expected
to offer control over the number of generated superpixels. We further expect the
algorithms to meet the desired number of superpixels within acceptable bounds. For
several algorithms, however, the number of generated superpixels is strongly dependent
on other parameters. Figure \ref{fig:parameter-optimization-superpixels} demonstrates the
influence of specific parameters on the number of generated superpixels
(before ensuring connectivity as in Section \ref{subsec:parameter-optimization-connectivity})
for \LSCr, \CISr, \VCr, \CRSr and \PBr. For some of the algorithms, such parameters needed to be constrained
to an appropriate value range even after enforcing connectivity.

For oversegmentation algorithms such as \FH, \EAMS and \QS not providing control
over the number of generated superpixels, we attempted to exploit simple relationships
between the provided parameters and the number of generated superpixels. For \EAMS
and \QS this allows to control the number of generated superpixels at least roughly.
\FH, in contrast, does not allow to control the number of generated superpixels
as easily. Therefore, we evaluated \FH for a large set of parameter combinations
and chose the parameters resulting in approximately the desired number of superpixels.

\FloatBarrier
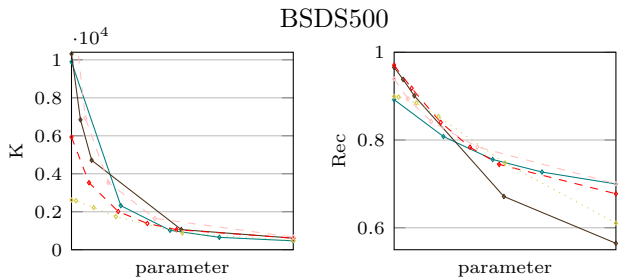
\begin{figure}
	\centering
\begin{subfigure}[b]{\halftwoone\textwidth}
    \begin{tikzpicture}
        \begin{axis}[
                POsuperpixelsSP,
                xmin=1,xmax=10,
                xtick=\empty,
                ylabel=\K,
                xlabel=parameter,
                title=\BSDS,
                title style={xshift=2cm}]

            \addplot[CIS,transparent] coordinates{
                (1,9887.95)
                (3,2321)
                (5,1022.8)
                (7,652.94)
                (10,468.2)
            };
        \end{axis}
        \begin{axis}[
                POsuperpixelsSP,
                xmin=1,xmax=10,
                axis y line=none,
                axis x line=none,
                xlabel=none,
                ylabel=none]

            \addplot[CIS] coordinates{
                (1,9887.95)
                (3,2321)
                (5,1022.8)
                (7,652.94)
                (10,468.2)
            };
            \label{plot:parameter-optimization-superpixels-cis}
        \end{axis}

        \begin{axis}[
                POsuperpixelsSP,
                xmin=0.001,xmax=0.1,
                axis y line=none,
                axis x line=none,
                xlabel=none,
                ylabel=none,]

            \addplot[CRS] coordinates{
                (0.001,10314.1)
                (0.005,6844.61)
                (0.01,4712.74)
                (0.05,1060.46)
                (0.1,609.12)
            };
            \label{plot:parameter-optimization-superpixels-crs}
        \end{axis}

        \begin{axis}[
                POsuperpixelsSP,
                xmin=1,xmax=20,
                axis y line=none,
                axis x line=none,
                xlabel=none,
                ylabel=none]

            \addplot[PB] coordinates{
                (1,5933.48)
                (2.5,3532.01)
                (5,2012.83)
                (7.5,1382.84)
                (10,1057.3)
                (20,608.12)
            };
            \label{plot:parameter-optimization-superpixels-pb}
        \end{axis}

        \begin{axis}[
                POsuperpixelsSP,
                xmin=0,xmax=0.5,
                axis y line=none,
                axis x line=none,
                xlabel=none,
                ylabel=none]

            \addplot[LSC] coordinates{
                (0,2616.8)
                (0.01,2581.57)
                (0.05,2218.48)
                (0.1,1743.55)
                (0.25,880.76)
                (0.5,488.51)
            };
            \label{plot:parameter-optimization-superpixels-lsc}
        \end{axis}

        \begin{axis}[
                POsuperpixelsSP,
                xmin=10,xmax=250,
                axis y line=none,
                axis x line=none,
                xlabel=none,
                ylabel=none]

            \addplot[VC] coordinates{
                (10,13548.4)
                (25,6936.92)
                (50,3539.36)
                (100,1651.79)
                (250,662.58)
            };
            \label{plot:parameter-optimization-superpixels-vc}
        \end{axis}
    \end{tikzpicture}
\end{subfigure}
\begin{subfigure}[b]{\halftwoone\textwidth}
    \begin{tikzpicture}
        \begin{axis}[
                POsuperpixelsRec,
                xmin=1,xmax=10,
                xtick=\empty,
                ylabel=\Rec,
                xlabel=parameter]

            \addplot[CIS,transparent] coordinates{
                (1,0.891961)
                (3,0.807977)
                (5,0.755691)
                (7,0.726976)
                (10,0.69987)
            };
        \end{axis}
        \begin{axis}[
                POsuperpixelsRec,
                xmin=1,xmax=10,
                axis y line=none,
                axis x line=none,
                xlabel=none,
                ylabel=none]

            \addplot[CIS] coordinates{
                (1,0.891961)
                (3,0.807977)
                (5,0.755691)
                (7,0.726976)
                (10,0.69987)
            };
            \label{plot:parameter-optimization-superpixels-cis}
        \end{axis}

        \begin{axis}[
                POsuperpixelsRec,
                xmin=0.001,xmax=0.1,
                axis y line=none,
                axis x line=none,
                xlabel=none,
                ylabel=none,]

            \addplot[CRS] coordinates{
                (0.001,0.965601)
                (0.005,0.938007)
                (0.01,0.900849)
                (0.05,0.671317)
                (0.1,0.564656)
            };
            \label{plot:parameter-optimization-superpixels-crs}
        \end{axis}

        \begin{axis}[
                POsuperpixelsRec,
                xmin=1,xmax=20,
                axis y line=none,
                axis x line=none,
                xlabel=none,
                ylabel=none]

            \addplot[PB] coordinates{
                (1,0.970505)
                (2.5,0.917979)
                (5,0.840081)
                (7.5,0.783602)
                (10,0.744347)
                (20,0.677509)
            };
            \label{plot:parameter-optimization-superpixels-pb}
        \end{axis}

        \begin{axis}[
                POsuperpixelsRec,
                xmin=0,xmax=0.5,
                axis y line=none,
                axis x line=none,
                xlabel=none,
                ylabel=none]

            \addplot[LSC] coordinates{
                (0,0.89937)
                (0.01,0.898078)
                (0.05,0.884433)
                (0.1,0.853266)
                (0.25,0.747625)
                (0.5,0.61054)
            };
            \label{plot:parameter-optimization-superpixels-lsc}
        \end{axis}

        \begin{axis}[
                POsuperpixelsRec,
                xmin=10,xmax=250,
                axis y line=none,
                axis x line=none,
                xlabel=none,
                ylabel=none]

            \addplot[VC] coordinates{
                (10,0.939249)
                (25,0.893879)
                (50,0.842431)
                (100,0.784252)
                (250,0.700503)
            };
            \label{plot:parameter-optimization-superpixels-vc}
        \end{axis}
    \end{tikzpicture}
\end{subfigure}
	\caption{\K and \Rec on the training set of the \BSDS dataset when varying
	parameters strongly influencing the number of generated superpixels
	of: \LSCr; \CISr; \VCr; \CRSr; and \PBr.
	The parameters have been omitted and scaled for clarity.
	A higher number of superpixels results in increased \Rec. Therefore, unnoticed
	superpixels inherently complicate fair comparison.
	\textbf{Best viewed in color.}
	}
	\label{fig:parameter-optimization-superpixels}
\end{figure}
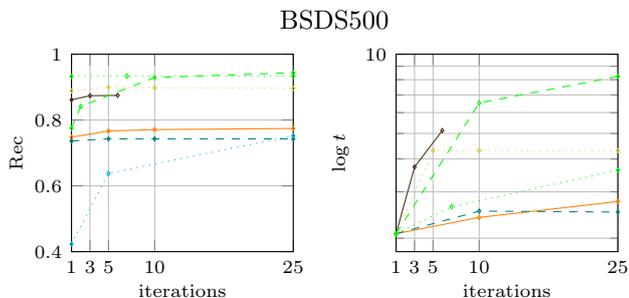
\begin{figure}
	\centering
\begin{subfigure}[b]{\halftwoone\textwidth}
    \begin{tikzpicture}
        \begin{axis}[
                POiterationsRec,
                xmin=1,xmax=25,
                yshift=-0cm,
                xtick={1,3,5,10,25},
                ylabel=\Rec,
                xlabel=iterations,
                title=\BSDS,
                title style={xshift=2cm}]

            \addplot[CIS,transparent] coordinates{
                (1,0.729313)
                (3,0.69987)
            };
        \end{axis}
        %

        \begin{axis}[
                POiterationsRec,
                xmin=1,xmax=25,
                axis y line=none,
                axis x line=none,
                xlabel=none,
                ylabel=none]

            \addplot[SLIC] coordinates{
                (1,0.747771)
                (5,0.766599)
                (10,0.770892)
                (25,0.774302)
                (50,0.773565)
            };
            \label{plot:parameter-optimization-iterations-preslic}
        \end{axis}

        \begin{axis}[
                POiterationsRec,
                xmin=1,xmax=25,
                axis y line=none,
                axis x line=none,
                xlabel=none,
                ylabel=none]

            \addplot[preSLIC] coordinates{
                (1,0.736228)
                (5,0.742702)
                (10,0.742702)
                (25,0.742702)
                (50,0.742702)
            };
            \label{plot:parameter-optimization-iterations-preslic}
        \end{axis}

        \begin{axis}[
                POiterationsRec,
                xmin=1,xmax=25,
                axis y line=none,
                axis x line=none,
                xlabel=none,
                ylabel=none]

            \addplot[LSC] coordinates{
                (1,0.888483)
                (5,0.89937)
                (10,0.898183)
                (25,0.896311)
                (50,0.895474)
            };
            \label{plot:parameter-optimization-iterations-lsc}
        \end{axis}

        \begin{axis}[
                POiterationsRec,
                xmin=1,xmax=25,
                axis y line=none,
                axis x line=none,
                xlabel=none,
                ylabel=none]

            \addplot[CRS] coordinates{
                (1,0.861408)
                (3,0.873871)
                (6,0.874965)
            };
            \label{plot:parameter-optimization-iterations-crs}
        \end{axis}

        \begin{axis}[
                POiterationsRec,
                xmin=1,xmax=25,
                axis y line=none,
                axis x line=none,
                xlabel=none,
                ylabel=none]

            \addplot[SEEDS] coordinates{
                (1,0.775507)
                (2,0.841535)
                (10,0.929071)
                (25,0.94386)
            };
            \label{plot:parameter-optimization-iterations-seeds}
        \end{axis}

        \begin{axis}[
                POiterationsRec,
                xmin=5,xmax=25,
                axis y line=none,
                axis x line=none,
                xlabel=none,
                ylabel=none]

            \addplot[ETPS] coordinates{
                (5,0.932699)
                (10,0.932985)
                (25,0.933014)
            };
            \label{plot:parameter-optimization-iterations-etps}
        \end{axis}

        \begin{axis}[
                POiterationsRec,
                xmin=1,xmax=25,
                axis y line=none,
                axis x line=none,
                xlabel=none,
                ylabel=none]

            \addplot[CCS] coordinates{
                (1,0.423352)
                (5,0.637979)
                (25,0.752122)
            };
            \label{plot:parameter-optimization-iterations-ccs}
        \end{axis}
    \end{tikzpicture}
\end{subfigure}
\begin{subfigure}[b]{\halftwoone\textwidth}
    \begin{tikzpicture}
        \begin{axis}[
                POiterationst,
                xmin=1,xmax=25,
                yshift=-0cm,
                xtick={1,3,5,10,25},
                ytick={0,0.1,1,10},
                ylabel=$\log t$,
                xlabel=iterations,
                ymode=log]

            \addplot[CIS,transparent] coordinates{
                (1,2.08)
                (3,5.26)
            };
        \end{axis}
        %

        \begin{axis}[
                POiterationst,
                xmin=1,xmax=25,
                axis y line=none,
                axis x line=none,
                xlabel=none,
                ylabel=none,
                ymode=log]

            \addplot[SLIC] coordinates{
                (1,0.059)
                (10,0.094)
                (25,0.148)
                (50,0.242)
            };
            \label{plot:parameter-optimization-iterations-preslic}
        \end{axis}

        \begin{axis}[
                POiterationst,
                xmin=1,xmax=25,
                axis y line=none,
                axis x line=none,
                xlabel=none,
                ylabel=none,
                ymode=log]

            \addplot[preSLIC] coordinates{
                (1,0.013)
                (10,0.03)
                (25,0.029)
                (50,0.03)
            };
            \label{plot:parameter-optimization-iterations-preslic}
        \end{axis}

        \begin{axis}[
                POiterationst,
                xmin=1,xmax=25,
                axis y line=none,
                axis x line=none,
                xlabel=none,
                ylabel=none,
                ymode=log]

            \addplot[LSC] coordinates{
                (1,0.112)
                (5,0.89937)
                (10,0.898183)
                (25,0.896311)
                (50,0.895474)
            };
            \label{plot:parameter-optimization-iterations-lsc}
        \end{axis}

        \begin{axis}[
                POiterationst,
                xmin=1,xmax=25,
                axis y line=none,
                axis x line=none,
                xlabel=none,
                ylabel=none,
                ymode=log]

            \addplot[CRS] coordinates{
                (1,0.329)
                (3,1.169)
                (6,2.342)
            };
            \label{plot:parameter-optimization-iterations-crs}
        \end{axis}

        \begin{axis}[
                POiterationst,
                xmin=1,xmax=25,
                axis y line=none,
                axis x line=none,
                xlabel=none,
                ylabel=none,
                ymode=log]

            \addplot[SEEDS] coordinates{
                (1,0.026)
                (10,1.986)
                (25,4.775)
            };
            \label{plot:parameter-optimization-iterations-seeds}
        \end{axis}

        \begin{axis}[
                POiterationst,
                xmin=5,xmax=25,
                axis y line=none,
                axis x line=none,
                xlabel=none,
                ylabel=none,
                ymode=log]

            \addplot[ETPS] coordinates{
                (5,0.167)
                (10,0.308)
                (25,0.711)
            };
            \label{plot:parameter-optimization-iterations-etps}
        \end{axis}

        \begin{axis}[
                POiterationsRec,
                xmin=1,xmax=25,
                axis y line=none,
                axis x line=none,
                xlabel=none,
                ylabel=none]

            \addplot[CCS] coordinates{
                (1,0.0665)
                (5,0.0985002)
                (25,0.14135)
            };
            \label{plot:parameter-optimization-iterations-ccs}
        \end{axis}
    \end{tikzpicture}
\end{subfigure}
    \caption{\Rec and runtime in seconds $t$ on the training set of the \BSDS dataset
	when varying the number of iterations of: \SLICr; \CRSr; \SEEDSr; \preSLICr; \LSCr; and \ETPSr.
	Most algorithms achieve reasonable \Rec with about $3 - 10$ iterations. Still,
	parameter optimization with respect to \Rec and \UE favors more iterations.
	\textbf{Best viewed in color.}}
    \label{fig:parameter-optimization-iterations}
\end{figure}
\begin{figure}
	\centering
    \input{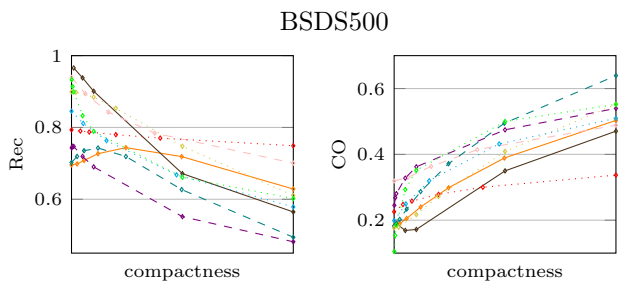}
    \caption{\Rec and \CO on the training set of the \BSDS dataset when varying the compactness parameter of:
	\SLICr; \CRSr; \VCr; \preSLICr; \CWr; \ERGCr; \LSCr; and \ETPSr.
	The parameters have been omitted and scaled for clarity.
	High \CO comes at the cost of reduced \Rec and parameter optimization
	with respect to \Rec and \UE results in less compact superpixels.
	\textbf{Best viewed in color.}}
    \label{fig:parameter-optimization-compactness}
	\vskip 12px
	{\scriptsize
		\def\arraystretch{0.8}
		\begin{tabularx}{0.475\textwidth}{X X X l}
			\ref{plot:cis} \CIS &
			\ref{plot:slic} \SLIC &
			\ref{plot:crs} \CRS &
			\ref{plot:ers} \ERS \\
			\ref{plot:pb} \PB &
			\ref{plot:seeds} \SEEDS &
			\ref{plot:vc} \VC &
			\ref{plot:ccs} \CCS \\
			\ref{plot:cw} \CW &
			\ref{plot:ergc} \ERGC &
			\ref{plot:preslic} \preSLIC &
			\ref{plot:wp} \WP \\
			\ref{plot:etps} \ETPS &
			\ref{plot:lsc} \LSC & &
		\end{tabularx}
	}

\end{figure}
\FloatBarrier

\subsection{Ensuring Connectivity}
\label{subsec:parameter-optimization-connectivity}

Unfortunately, many implementations (note the difference between implementation and algorithm)
cannot ensure the connectivity of the generated superpixels as required in Section \ref{sec:introduction}.
Therefore, we decided to strictly enforce connectivity using a connected components algorithm,
\ie  after computing superpixels, each connected component is relabeled as separate superpixel. For some implementations,
this results in many unintended superpixels comprising few pixels. In these cases
we additionally merge the newly generated superpixels into larger neighboring ones.
However, even with these post-processing steps, the evaluated implementations of \CIS, \CRS, \PB, \DASP, \VC, \VCCS or \LSC
generate highly varying numbers of superpixels across different images.

\subsection{Common Trade-Offs: Runtime and Compactness}
\label{subsec:parameter-optimization-trade-offs}

Two other types of parameters deserve detailed discussion: the number of iterations
and the compactness parameter. The former controls the trade-off between runtime and performance,
exemplarily demonstrated in Figure \ref{fig:parameter-optimization-iterations}
showing that more iterations usually result in higher \Rec and higher runtime in
seconds $t$. The latter controls the trade-off between compactness and performance and
Figure \ref{fig:parameter-optimization-compactness} shows that higher \CO usually
results in lower \Rec. Overall, parameter optimization with respect to \Rec and \UE
results in higher runtime and lower compactness.

\section{Experiments}
\label{sec:experiments}

Our experiments include visual quality, performance with respect to \Rec, \UE and
\EV as well as runtime. In contrast to existing work \cite{SchickFischerStiefelhagen:2012, AchantaShajiSmithLucchiFuaSuesstrunk:2012, NeubertProtzel:2012, SchickFischerStiefelhagen:2012}, we consider
minimum/maximum and standard deviation of \Rec, \UE and \EV (in relation to the number of generated superpixels \K) and present results
for the introduced metrics \ARec, \AUE and \AEV.
Furthermore, we present experiments regarding implementation details as well as robustness
against noise, blur and affine transformations. Finally, we give an overall ranking based on \ARec and \AUE.

\subsection{Qualitative}
\label{subsec:experiments-qualitative}

\def\BSDSCroppedScale{0.25}
\def\SBDCroppedScale{0.3}
\def\FashCroppedScale{0.195}
\begin{figure*}
	\centering
	\vspace{-0.5cm}
	\begin{subfigure}[b]{0.02\textwidth}
		\rotatebox{90}{\small\hphantom{aaai}\W}
	\end{subfigure}
	\begin{subfigure}[b]{0.16\textwidth}
        \begin{center}
            \BSDS
        \end{center}
        \vskip -6px
		\includegraphics[height=1.65cm]{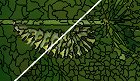}
	\end{subfigure}
	\begin{subfigure}[b]{0.129\textwidth}
        \begin{center}
            \SBD
        \end{center}
        \vskip -6px
		\includegraphics[height=1.65cm]{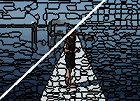}
	\end{subfigure}
	\begin{subfigure}[b]{0.10\textwidth}
        \begin{center}
            \Fash
        \end{center}
        \vskip -6px
		\includegraphics[height=1.65cm]{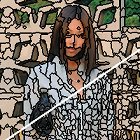}
	\end{subfigure}
	\begin{subfigure}[b]{0.02\textwidth}
		\rotatebox{90}{\small\hphantom{ai}\EAMS}
	\end{subfigure}
	\begin{subfigure}[b]{0.16\textwidth}
        \begin{center}
            \BSDS
        \end{center}
        \vskip -6px
		\includegraphics[height=1.65cm]{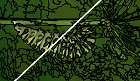}
	\end{subfigure}
	\begin{subfigure}[b]{0.129\textwidth}
        \begin{center}
            \SBD
        \end{center}
        \vskip -6px
		\includegraphics[height=1.65cm]{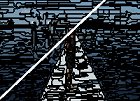}
	\end{subfigure}
	\begin{subfigure}[b]{0.10\textwidth}
        \begin{center}
            \Fash
        \end{center}
        \vskip -6px
		\includegraphics[height=1.65cm]{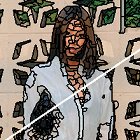}
	\end{subfigure}\\
	\begin{subfigure}[b]{0.02\textwidth}
		\rotatebox{90}{\small\hphantom{aaa}\NC}
	\end{subfigure}
	\begin{subfigure}[b]{0.16\textwidth}
		\includegraphics[height=1.65cm]{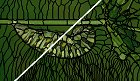}
	\end{subfigure}
	\begin{subfigure}[b]{0.129\textwidth}
		\includegraphics[height=1.65cm]{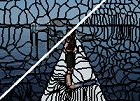}
	\end{subfigure}
	\begin{subfigure}[b]{0.10\textwidth}
		\includegraphics[height=1.65cm]{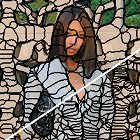}
	\end{subfigure}
	\begin{subfigure}[b]{0.02\textwidth}
		\rotatebox{90}{\small\hphantom{aaa}\FH}
	\end{subfigure}
	\begin{subfigure}[b]{0.16\textwidth}
		\includegraphics[height=1.65cm]{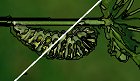}
	\end{subfigure}
	\begin{subfigure}[b]{0.129\textwidth}
		\includegraphics[height=1.65cm]{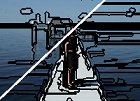}
	\end{subfigure}
	\begin{subfigure}[b]{0.10\textwidth}
		\includegraphics[height=1.65cm]{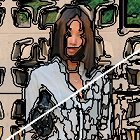}
	\end{subfigure}\\
	\begin{subfigure}[b]{0.02\textwidth}
		\rotatebox{90}{\small\hphantom{aaa}\RW}
	\end{subfigure}
	\begin{subfigure}[b]{0.16\textwidth}
		\includegraphics[height=1.65cm]{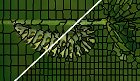}
	\end{subfigure}
	\begin{subfigure}[b]{0.129\textwidth}
		\includegraphics[height=1.65cm]{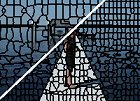}
	\end{subfigure}
	\begin{subfigure}[b]{0.10\textwidth}
		\includegraphics[height=1.65cm]{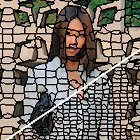}
	\end{subfigure}
	\begin{subfigure}[b]{0.02\textwidth}
		\rotatebox{90}{\small\hphantom{aaa}\QS}
	\end{subfigure}
	\begin{subfigure}[b]{0.16\textwidth}
		\includegraphics[height=1.65cm]{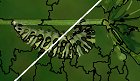}
	\end{subfigure}
	\begin{subfigure}[b]{0.129\textwidth}
		\includegraphics[height=1.65cm]{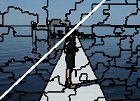}
	\end{subfigure}
	\begin{subfigure}[b]{0.10\textwidth}
		\includegraphics[height=1.65cm]{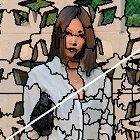}
	\end{subfigure}\\
	\begin{subfigure}[b]{0.02\textwidth}
		\rotatebox{90}{\small\hphantom{aaa}\PF}
	\end{subfigure}
	\begin{subfigure}[b]{0.16\textwidth}
		\includegraphics[height=1.65cm]{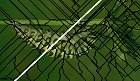}
	\end{subfigure}
	\begin{subfigure}[b]{0.129\textwidth}
		\includegraphics[height=1.65cm]{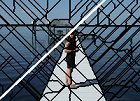}
	\end{subfigure}
	\begin{subfigure}[b]{0.10\textwidth}
		\includegraphics[height=1.65cm]{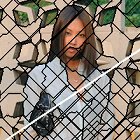}
	\end{subfigure}
	\begin{subfigure}[b]{0.02\textwidth}
		\rotatebox{90}{\small\hphantom{aaa}\TP}
	\end{subfigure}
	\begin{subfigure}[b]{0.16\textwidth}
		\includegraphics[height=1.65cm]{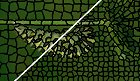}
	\end{subfigure}
	\begin{subfigure}[b]{0.129\textwidth}
		\includegraphics[height=1.65cm]{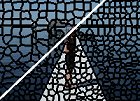}
	\end{subfigure}
	\begin{subfigure}[b]{0.10\textwidth}
		\includegraphics[height=1.65cm]{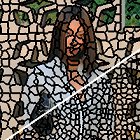}
	\end{subfigure}\\
	\begin{subfigure}[b]{0.02\textwidth}
		\rotatebox{90}{\small\hphantom{aai}\CIS}
	\end{subfigure}
	\begin{subfigure}[b]{0.16\textwidth}
		\includegraphics[height=1.65cm]{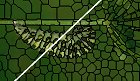}
	\end{subfigure}
	\begin{subfigure}[b]{0.129\textwidth}
		\includegraphics[height=1.65cm]{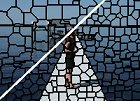}
	\end{subfigure}
	\begin{subfigure}[b]{0.10\textwidth}
		\includegraphics[height=1.65cm]{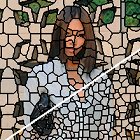}
	\end{subfigure}
	\begin{subfigure}[b]{0.02\textwidth}
		\rotatebox{90}{\small\hphantom{aa}\SLIC}
	\end{subfigure}
	\begin{subfigure}[b]{0.16\textwidth}
		\includegraphics[height=1.65cm]{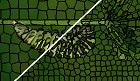}
	\end{subfigure}
	\begin{subfigure}[b]{0.129\textwidth}
		\includegraphics[height=1.65cm]{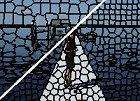}
	\end{subfigure}
	\begin{subfigure}[b]{0.10\textwidth}
		\includegraphics[height=1.65cm]{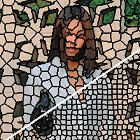}
	\end{subfigure}\\
	\begin{subfigure}[b]{0.02\textwidth}
		\rotatebox{90}{\small\hphantom{aai}\CRS}
	\end{subfigure}
	\begin{subfigure}[b]{0.16\textwidth}
		\includegraphics[height=1.65cm]{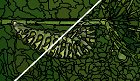}
	\end{subfigure}
	\begin{subfigure}[b]{0.129\textwidth}
		\includegraphics[height=1.65cm]{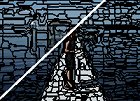}
	\end{subfigure}
	\begin{subfigure}[b]{0.10\textwidth}
		\includegraphics[height=1.65cm]{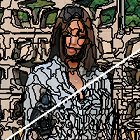}
	\end{subfigure}
	\begin{subfigure}[b]{0.02\textwidth}
		\rotatebox{90}{\small\hphantom{aai}\ERS}
	\end{subfigure}
	\begin{subfigure}[b]{0.16\textwidth}
		\includegraphics[height=1.65cm]{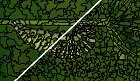}
	\end{subfigure}
	\begin{subfigure}[b]{0.129\textwidth}
		\includegraphics[height=1.65cm]{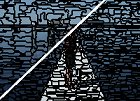}
	\end{subfigure}
	\begin{subfigure}[b]{0.10\textwidth}
		\includegraphics[height=1.65cm]{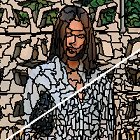}
	\end{subfigure}\\
	\begin{subfigure}[b]{0.02\textwidth}
		\rotatebox{90}{\small\hphantom{aaa}\PB}
	\end{subfigure}
	\begin{subfigure}[b]{0.16\textwidth}
		\includegraphics[height=1.65cm]{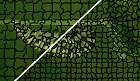}
	\end{subfigure}
	\begin{subfigure}[b]{0.129\textwidth}
		\includegraphics[height=1.65cm]{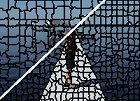}
	\end{subfigure}
	\begin{subfigure}[b]{0.10\textwidth}
		\includegraphics[height=1.65cm]{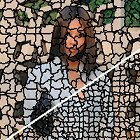}
	\end{subfigure}
	\begin{subfigure}[b]{0.02\textwidth}
		\rotatebox{90}{\small\hphantom{a}\SEEDS}
	\end{subfigure}
	\begin{subfigure}[b]{0.16\textwidth}
		\includegraphics[height=1.65cm]{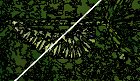}
	\end{subfigure}
	\begin{subfigure}[b]{0.129\textwidth}
		\includegraphics[height=1.65cm]{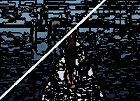}
	\end{subfigure}
	\begin{subfigure}[b]{0.10\textwidth}
		\includegraphics[height=1.65cm]{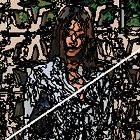}
	\end{subfigure}\\
	\begin{subfigure}[b]{0.02\textwidth}
		\rotatebox{90}{\small\hphantom{aai}\TPS}
	\end{subfigure}
	\begin{subfigure}[b]{0.16\textwidth}
		\includegraphics[height=1.65cm]{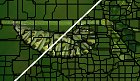}
	\end{subfigure}
	\begin{subfigure}[b]{0.129\textwidth}
		\includegraphics[height=1.65cm]{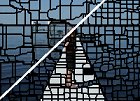}
	\end{subfigure}
	\begin{subfigure}[b]{0.10\textwidth}
		\includegraphics[height=1.65cm]{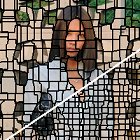}
	\end{subfigure}
	\begin{subfigure}[b]{0.02\textwidth}
		\rotatebox{90}{\small\hphantom{aaa}\VC}
	\end{subfigure}
	\begin{subfigure}[b]{0.16\textwidth}
		\includegraphics[height=1.65cm]{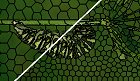}
	\end{subfigure}
	\begin{subfigure}[b]{0.129\textwidth}
		\includegraphics[height=1.65cm]{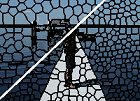}
	\end{subfigure}
	\begin{subfigure}[b]{0.10\textwidth}
		\includegraphics[height=1.65cm]{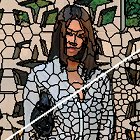}
	\end{subfigure}\\
	\begin{subfigure}[b]{0.02\textwidth}
		\rotatebox{90}{\small\hphantom{aai}\CCS}
	\end{subfigure}
	\begin{subfigure}[b]{0.16\textwidth}
		\includegraphics[height=1.65cm]{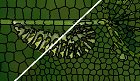}
	\end{subfigure}
	\begin{subfigure}[b]{0.129\textwidth}
		\includegraphics[height=1.65cm]{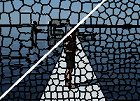}
	\end{subfigure}
	\begin{subfigure}[b]{0.10\textwidth}
		\includegraphics[height=1.65cm]{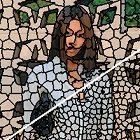}
	\end{subfigure}
	\begin{subfigure}[b]{0.02\textwidth}
		\rotatebox{90}{\small\hphantom{aaa}\CW}
	\end{subfigure}
	\begin{subfigure}[b]{0.16\textwidth}
		\includegraphics[height=1.65cm]{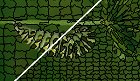}
	\end{subfigure}
	\begin{subfigure}[b]{0.129\textwidth}
		\includegraphics[height=1.65cm]{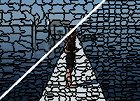}
	\end{subfigure}
	\begin{subfigure}[b]{0.10\textwidth}
		\includegraphics[height=1.65cm]{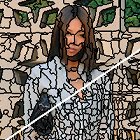}
	\end{subfigure}\\
	\begin{subfigure}[b]{0.02\textwidth}
		\rotatebox{90}{\small\hphantom{aa}\ERGC}
	\end{subfigure}
	\begin{subfigure}[b]{0.16\textwidth}
		\includegraphics[height=1.65cm]{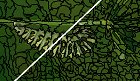}
	\end{subfigure}
	\begin{subfigure}[b]{0.129\textwidth}
		\includegraphics[height=1.65cm]{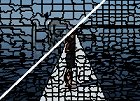}
	\end{subfigure}
	\begin{subfigure}[b]{0.10\textwidth}
		\includegraphics[height=1.65cm]{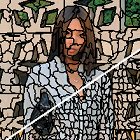}
	\end{subfigure}
	\begin{subfigure}[b]{0.02\textwidth}
		\rotatebox{90}{\small\hphantom{aai}\MSS}
	\end{subfigure}
	\begin{subfigure}[b]{0.16\textwidth}
		\includegraphics[height=1.65cm]{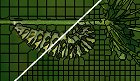}
	\end{subfigure}
	\begin{subfigure}[b]{0.129\textwidth}
		\includegraphics[height=1.65cm]{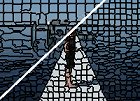}
	\end{subfigure}
	\begin{subfigure}[b]{0.10\textwidth}
		\includegraphics[height=1.65cm]{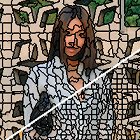}
	\end{subfigure}\\
	\begin{subfigure}[b]{0.02\textwidth}
		\rotatebox{90}{\small\hphantom{a}\preSLIC}
	\end{subfigure}
	\begin{subfigure}[b]{0.16\textwidth}
		\includegraphics[height=1.65cm]{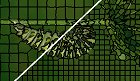}
	\end{subfigure}
	\begin{subfigure}[b]{0.129\textwidth}
		\includegraphics[height=1.65cm]{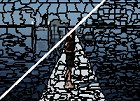}
	\end{subfigure}
	\begin{subfigure}[b]{0.10\textwidth}
		\includegraphics[height=1.65cm]{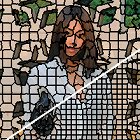}
	\end{subfigure}
	\begin{subfigure}[b]{0.02\textwidth}
		\rotatebox{90}{\small\hphantom{aaa}\WP}
	\end{subfigure}
	\begin{subfigure}[b]{0.16\textwidth}
		\includegraphics[height=1.65cm]{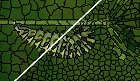}
	\end{subfigure}
	\begin{subfigure}[b]{0.129\textwidth}
		\includegraphics[height=1.65cm]{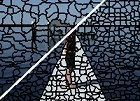}
	\end{subfigure}
	\begin{subfigure}[b]{0.10\textwidth}
		\includegraphics[height=1.65cm]{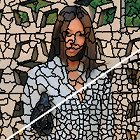}
	\end{subfigure}\\
	\begin{subfigure}[b]{0.02\textwidth}
		\rotatebox{90}{\small\hphantom{aa}\ETPS}
	\end{subfigure}
	\begin{subfigure}[b]{0.16\textwidth}
		\includegraphics[height=1.65cm]{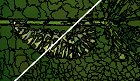}
	\end{subfigure}
	\begin{subfigure}[b]{0.129\textwidth}
		\includegraphics[height=1.65cm]{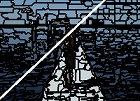}
	\end{subfigure}
	\begin{subfigure}[b]{0.10\textwidth}
		\includegraphics[height=1.65cm]{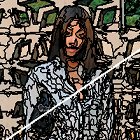}
	\end{subfigure}
	\begin{subfigure}[b]{0.02\textwidth}
		\rotatebox{90}{\small\hphantom{aai}\LSC}
	\end{subfigure}
	\begin{subfigure}[b]{0.16\textwidth}
		\includegraphics[height=1.65cm]{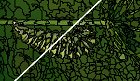}
	\end{subfigure}
	\begin{subfigure}[b]{0.129\textwidth}
		\includegraphics[height=1.65cm]{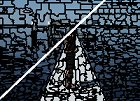}
	\end{subfigure}
	\begin{subfigure}[b]{0.10\textwidth}
		\includegraphics[height=1.65cm]{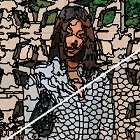}
	\end{subfigure}\\
	\begin{subfigure}[b]{0.02\textwidth}
		\rotatebox{90}{\small\hphantom{a}\POISE}
	\end{subfigure}
	\begin{subfigure}[b]{0.16\textwidth}
		\includegraphics[height=1.65cm]{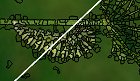}
	\end{subfigure}
	\begin{subfigure}[b]{0.129\textwidth}
		\includegraphics[height=1.65cm]{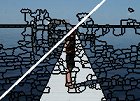}
	\end{subfigure}
	\begin{subfigure}[b]{0.10\textwidth}
		\includegraphics[height=1.65cm]{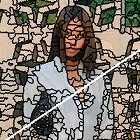}
	\end{subfigure}
	\begin{subfigure}[b]{0.02\textwidth}
		\rotatebox{90}{\small\hphantom{ai}\SEAW}
	\end{subfigure}
	\begin{subfigure}[b]{0.16\textwidth}
		\includegraphics[height=1.65cm]{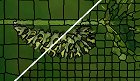}
	\end{subfigure}
	\begin{subfigure}[b]{0.129\textwidth}
		\includegraphics[height=1.65cm]{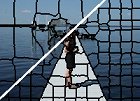}
	\end{subfigure}
	\begin{subfigure}[b]{0.10\textwidth}
		\includegraphics[height=1.65cm]{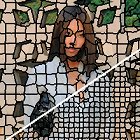}
	\end{subfigure}
	\caption{Qualitative results on the \BSDS, \SBD and \Fash datasets. Excerpts from the
	images in Figure \ref{fig:datasets} are shown for $K \approx 400$ in the upper left corner
	and $K \approx 1200$ in the lower right corner. Superpixel boundaries are depicted
	in black; best viewed in color. We judge visual quality on the basis of
	boundary adherence, compactness, smoothness and regularity.
	Boundary adherence can be judged both on the caterpillar image as well 
	as on the woman image -- the caterpillar's boundaries are hard to detect and 
	the woman's face exhibits small details. In contrast, compactness, regularity and smoothness 
	can be evaluated considering the background in the caterpillar and see images.
	\textbf{Best viewed in color.}}
	\label{fig:experiments-qualitative-bsds500-sbd-fash}
\end{figure*}

\def\NYUCroppedScale{0.18}
\def\SUNRGBDCroppedScale{0.14}
\begin{figure*}
	\centering
	\begin{subfigure}[b]{0.02\textwidth}
		\rotatebox{90}{\small\hphantom{aaa}\NC}
	\end{subfigure}
	\begin{subfigure}[b]{0.1375\textwidth}
		\begin{center}
			\NYU
		\end{center}
		\vskip -6px
		\includegraphics[height=1.65cm]{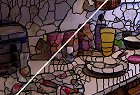}
	\end{subfigure}
	\begin{subfigure}[b]{0.129\textwidth}
		\begin{center}
			\NYU
		\end{center}
		\vskip -6px
		\includegraphics[height=1.65cm]{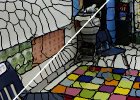}
	\end{subfigure}
	\begin{subfigure}[b]{0.02\textwidth}
		\rotatebox{90}{\small\hphantom{aaa}\RW}
	\end{subfigure}
	\begin{subfigure}[b]{0.1375\textwidth}
		\begin{center}
			\NYU
		\end{center}
		\vskip -6px
		\includegraphics[height=1.65cm]{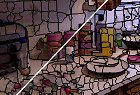}
	\end{subfigure}
	\begin{subfigure}[b]{0.129\textwidth}
		\begin{center}
			\NYU
		\end{center}
		\vskip -6px
		\includegraphics[height=1.65cm]{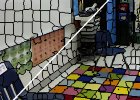}
	\end{subfigure}
	\begin{subfigure}[b]{0.02\textwidth}
		\rotatebox{90}{\small\hphantom{ai}\SEAW}
	\end{subfigure}
	\begin{subfigure}[b]{0.1375\textwidth}
		\begin{center}
			\NYU
		\end{center}
		\vskip -6px
		\includegraphics[height=1.65cm]{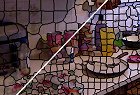}
	\end{subfigure}
	\begin{subfigure}[b]{0.129\textwidth}
		\begin{center}
			\NYU
		\end{center}
		\vskip -6px
		\includegraphics[height=1.65cm]{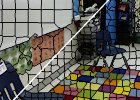}
	\end{subfigure}\\[4px]
	\begin{subfigure}[b]{0.02\textwidth}
		\rotatebox{90}{\small\hphantom{aaai}\W}
	\end{subfigure}
	\begin{subfigure}[b]{0.1375\textwidth}
		\begin{center}
			\NYU
		\end{center}
		\vskip -6px
		\includegraphics[height=1.65cm]{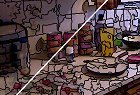}
	\end{subfigure}
	\begin{subfigure}[b]{0.129\textwidth}
		\begin{center}
			\SUNRGBD
		\end{center}
		\vskip -6px
		\includegraphics[height=1.65cm]{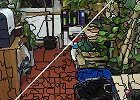}
	\end{subfigure}
	\begin{subfigure}[b]{0.02\textwidth}
		\rotatebox{90}{\small\hphantom{ai}\EAMS}
	\end{subfigure}
	\begin{subfigure}[b]{0.1375\textwidth}
		\begin{center}
			\NYU
		\end{center}
		\vskip -6px
		\includegraphics[height=1.65cm]{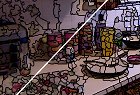}
	\end{subfigure}
	\begin{subfigure}[b]{0.129\textwidth}
		\begin{center}
			\SUNRGBD
		\end{center}
		\vskip -6px
		\includegraphics[height=1.65cm]{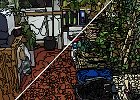}
	\end{subfigure}
	\begin{subfigure}[b]{0.02\textwidth}
		\rotatebox{90}{\small\hphantom{aaa}\FH}
	\end{subfigure}
	\begin{subfigure}[b]{0.1375\textwidth}
		\begin{center}
			\NYU
		\end{center}
		\vskip -6px
		\includegraphics[height=1.65cm]{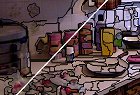}
	\end{subfigure}
	\begin{subfigure}[b]{0.129\textwidth}
		\begin{center}
			\SUNRGBD
		\end{center}
		\vskip -6px
		\includegraphics[height=1.65cm]{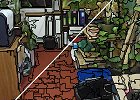}
	\end{subfigure}\\
	\begin{subfigure}[b]{0.02\textwidth}
		\rotatebox{90}{\small\hphantom{aaa}\QS}
	\end{subfigure}
	\begin{subfigure}[b]{0.1375\textwidth}
		\includegraphics[height=1.65cm]{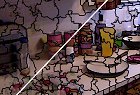}
	\end{subfigure}
	\begin{subfigure}[b]{0.129\textwidth}
		\includegraphics[height=1.65cm]{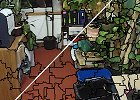}
	\end{subfigure}
	\begin{subfigure}[b]{0.02\textwidth}
		\rotatebox{90}{\small\hphantom{aaa}\PF}
	\end{subfigure}
	\begin{subfigure}[b]{0.1375\textwidth}
		\includegraphics[height=1.65cm]{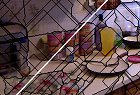}
	\end{subfigure}
	\begin{subfigure}[b]{0.129\textwidth}
		\includegraphics[height=1.65cm]{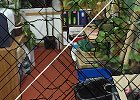}
	\end{subfigure}
	\begin{subfigure}[b]{0.02\textwidth}
		\rotatebox{90}{\small\hphantom{aaa}\TP}
	\end{subfigure}
	\begin{subfigure}[b]{0.1375\textwidth}
		\includegraphics[height=1.65cm]{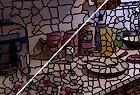}
	\end{subfigure}
	\begin{subfigure}[b]{0.129\textwidth}
		\includegraphics[height=1.65cm]{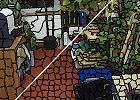}
	\end{subfigure}\\
	\begin{subfigure}[b]{0.02\textwidth}
		\rotatebox{90}{\small\hphantom{aai}\CIS}
	\end{subfigure}
	\begin{subfigure}[b]{0.1375\textwidth}
		\includegraphics[height=1.65cm]{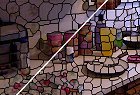}
	\end{subfigure}
	\begin{subfigure}[b]{0.129\textwidth}
		\includegraphics[height=1.65cm]{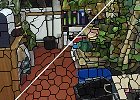}
	\end{subfigure}
	\begin{subfigure}[b]{0.02\textwidth}
		\rotatebox{90}{\small\hphantom{aa}\SLIC}
	\end{subfigure}
	\begin{subfigure}[b]{0.1375\textwidth}
		\includegraphics[height=1.65cm]{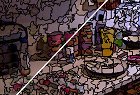}
	\end{subfigure}
	\begin{subfigure}[b]{0.129\textwidth}
		\includegraphics[height=1.65cm]{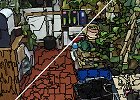}
	\end{subfigure}
	\begin{subfigure}[b]{0.02\textwidth}
		\rotatebox{90}{\small\hphantom{aai}\CRS}
	\end{subfigure}
	\begin{subfigure}[b]{0.1375\textwidth}
		\includegraphics[height=1.65cm]{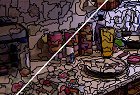}
	\end{subfigure}
	\begin{subfigure}[b]{0.129\textwidth}
		\includegraphics[height=1.65cm]{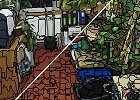}
	\end{subfigure}\\
	\begin{subfigure}[b]{0.02\textwidth}
		\rotatebox{90}{\small\hphantom{aai}\ERS}
	\end{subfigure}
	\begin{subfigure}[b]{0.1375\textwidth}
		\includegraphics[height=1.65cm]{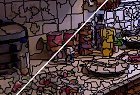}
	\end{subfigure}
	\begin{subfigure}[b]{0.129\textwidth}
		\includegraphics[height=1.65cm]{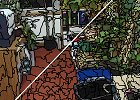}
	\end{subfigure}
	\begin{subfigure}[b]{0.02\textwidth}
		\rotatebox{90}{\small\hphantom{aaa}\PB}
	\end{subfigure}
	\begin{subfigure}[b]{0.1375\textwidth}
		\includegraphics[height=1.65cm]{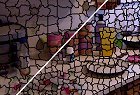}
	\end{subfigure}
	\begin{subfigure}[b]{0.129\textwidth}
		\includegraphics[height=1.65cm]{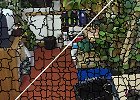}
	\end{subfigure}
	\begin{subfigure}[b]{0.02\textwidth}
		\rotatebox{90}{\small\hphantom{aa}\DASP}
	\end{subfigure}
	\begin{subfigure}[b]{0.1375\textwidth}
		\includegraphics[height=1.65cm]{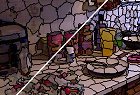}
	\end{subfigure}
	\begin{subfigure}[b]{0.129\textwidth}
		\includegraphics[height=1.65cm]{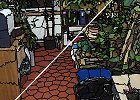}
	\end{subfigure}\\
	\begin{subfigure}[b]{0.02\textwidth}
		\rotatebox{90}{\small\hphantom{a}\SEEDS}
	\end{subfigure}
	\begin{subfigure}[b]{0.1375\textwidth}
		\includegraphics[height=1.65cm]{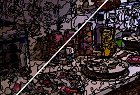}
	\end{subfigure}
	\begin{subfigure}[b]{0.129\textwidth}
		\includegraphics[height=1.65cm]{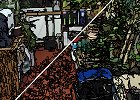}
	\end{subfigure}
	\begin{subfigure}[b]{0.02\textwidth}
		\rotatebox{90}{\small\hphantom{aai}\TPS}
	\end{subfigure}
	\begin{subfigure}[b]{0.1375\textwidth}
		\includegraphics[height=1.65cm]{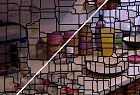}
	\end{subfigure}
	\begin{subfigure}[b]{0.129\textwidth}
		\includegraphics[height=1.65cm]{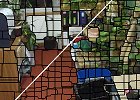}
	\end{subfigure}
	\begin{subfigure}[b]{0.02\textwidth}
		\rotatebox{90}{\small\hphantom{aaai}\VC}
	\end{subfigure}
	\begin{subfigure}[b]{0.1375\textwidth}
		\includegraphics[height=1.65cm]{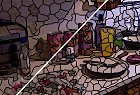}
	\end{subfigure}
	\begin{subfigure}[b]{0.129\textwidth}
		\includegraphics[height=1.65cm]{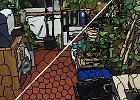}
	\end{subfigure}\\
	\begin{subfigure}[b]{0.02\textwidth}
		\rotatebox{90}{\small\hphantom{aai}\CCS}
	\end{subfigure}
	\begin{subfigure}[b]{0.1375\textwidth}
		\includegraphics[height=1.65cm]{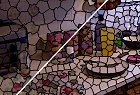}
	\end{subfigure}
	\begin{subfigure}[b]{0.129\textwidth}
		\includegraphics[height=1.65cm]{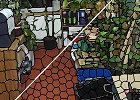}
	\end{subfigure}
	\begin{subfigure}[b]{0.02\textwidth}
		\rotatebox{90}{\small\hphantom{aa}\VCCS}
	\end{subfigure}
	\begin{subfigure}[b]{0.1375\textwidth}
		\includegraphics[height=1.65cm]{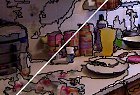}
	\end{subfigure}
	\begin{subfigure}[b]{0.129\textwidth}
		\includegraphics[height=1.65cm]{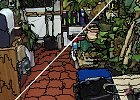}
	\end{subfigure}
	\begin{subfigure}[b]{0.02\textwidth}
		\rotatebox{90}{\small\hphantom{aaa}\CW}
	\end{subfigure}
	\begin{subfigure}[b]{0.1375\textwidth}
		\includegraphics[height=1.65cm]{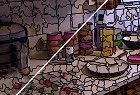}
	\end{subfigure}
	\begin{subfigure}[b]{0.129\textwidth}
		\includegraphics[height=1.65cm]{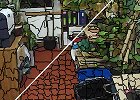}
	\end{subfigure}\\
	\begin{subfigure}[b]{0.02\textwidth}
		\rotatebox{90}{\small\hphantom{aa}\ERGC}
	\end{subfigure}
	\begin{subfigure}[b]{0.1375\textwidth}
		\includegraphics[height=1.65cm]{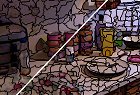}
	\end{subfigure}
	\begin{subfigure}[b]{0.129\textwidth}
		\includegraphics[height=1.65cm]{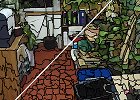}
	\end{subfigure}
	\begin{subfigure}[b]{0.02\textwidth}
		\rotatebox{90}{\small\hphantom{aai}\MSS}
	\end{subfigure}
	\begin{subfigure}[b]{0.1375\textwidth}
		\includegraphics[height=1.65cm]{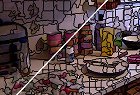}
	\end{subfigure}
	\begin{subfigure}[b]{0.129\textwidth}
		\includegraphics[height=1.65cm]{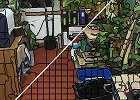}
	\end{subfigure}
	\begin{subfigure}[b]{0.02\textwidth}
		\rotatebox{90}{\small\hphantom{a}\preSLIC}
	\end{subfigure}
	\begin{subfigure}[b]{0.1375\textwidth}
		\includegraphics[height=1.65cm]{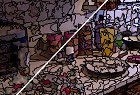}
	\end{subfigure}
	\begin{subfigure}[b]{0.129\textwidth}
		\includegraphics[height=1.65cm]{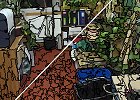}
	\end{subfigure}\\
	\begin{subfigure}[b]{0.02\textwidth}
		\rotatebox{90}{\small\hphantom{aaa}\WP}
	\end{subfigure}
	\begin{subfigure}[b]{0.1375\textwidth}
		\includegraphics[height=1.65cm]{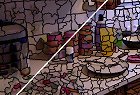}
	\end{subfigure}
	\begin{subfigure}[b]{0.129\textwidth}
		\includegraphics[height=1.65cm]{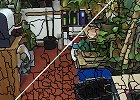}
	\end{subfigure}
	\begin{subfigure}[b]{0.02\textwidth}
		\rotatebox{90}{\small\hphantom{aa}\ETPS}
	\end{subfigure}
	\begin{subfigure}[b]{0.1375\textwidth}
		\includegraphics[height=1.65cm]{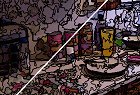}
	\end{subfigure}
	\begin{subfigure}[b]{0.129\textwidth}
		\includegraphics[height=1.65cm]{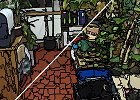}
	\end{subfigure}
	\begin{subfigure}[b]{0.02\textwidth}
		\rotatebox{90}{\small\hphantom{aaa}\LSC}
	\end{subfigure}
	\begin{subfigure}[b]{0.1375\textwidth}
		\includegraphics[height=1.65cm]{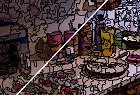}
	\end{subfigure}
	\begin{subfigure}[b]{0.129\textwidth}
		\includegraphics[height=1.65cm]{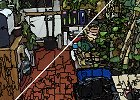}
	\end{subfigure}\\
	\begin{subfigure}[b]{0.02\textwidth}
		\rotatebox{90}{\small\hphantom{a}\POISE}
	\end{subfigure}
	\begin{subfigure}[b]{0.1375\textwidth}
		\includegraphics[height=1.65cm]{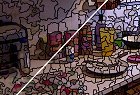}
	\end{subfigure}
	\begin{subfigure}[b]{0.129\textwidth}
		\includegraphics[height=1.65cm]{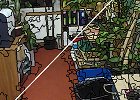}
	\end{subfigure}
	\begin{subfigure}[b]{0.02\textwidth}
		\hphantom{aa}
	\end{subfigure}
	\begin{subfigure}[b]{0.1375\textwidth}
		\hphantom{aaaaaaaaaaaaaaaaaaaaaaaaaaaaaii}
	\end{subfigure}
	\begin{subfigure}[b]{0.129\textwidth}
		\hphantom{aaaaaaaaaaaaaaaaaaaaaaaaaaa}
	\end{subfigure}
	\begin{subfigure}[b]{0.02\textwidth}
		\hphantom{aa}
	\end{subfigure}
	\begin{subfigure}[b]{0.1375\textwidth}
		\hphantom{aaaaaaaaaaaaaaaaaaaaaaaaaaaaii}
	\end{subfigure}
	\begin{subfigure}[b]{0.129\textwidth}
		\hphantom{aaaaaaaaaaaaaaaaaaaaaaaaaaa}
	\end{subfigure}
	\caption{Qualitative results on the \NYU and \SUNRGBD datasets. Excerpts from the images
	in Figure \ref{fig:datasets} are shown for $K \approx 400$ in the upper left corner
	and $K \approx 1200$ in the lower right corner. Superpixel boundaries are depicted in black; best viewed in color.
	\NC, \RW and \SEAW could not be evaluated on the \SUNRGBD dataset due
	to exhaustive memory usage of the corresponding MatLab implementations.
	Therefore, results for the \NYU dataset are shown.
    Visual quality is judged regarding boundary adherence, compactness,
	smoothness and regularity. We also find
	that depth information, as used in \DASP and \VCCS, may help resemble the underlying 3D-structure.
	\textbf{Best viewed in color.}}
	\label{fig:experiments-qualitative-nyuv2-sunrgbd}
\end{figure*}
\begin{figure*}
	\centering
	\begin{subfigure}[b]{0.02\textwidth}
		\rotatebox{90}{\small\hphantom{aa}\SLIC}
	\end{subfigure}
	\begin{subfigure}[b]{0.141\textwidth}
		\includegraphics[height=1.525cm]{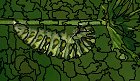}
	\end{subfigure}
	\begin{subfigure}[b]{0.141\textwidth}
		\includegraphics[height=1.525cm]{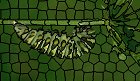}
	\end{subfigure}
	\begin{subfigure}[b]{0.141\textwidth}
		\includegraphics[height=1.525cm]{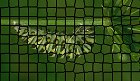}
	\end{subfigure}
	\begin{subfigure}[b]{0.02\textwidth}
		\rotatebox{90}{\small\hphantom{aai}\CRS}
	\end{subfigure}
	\begin{subfigure}[b]{0.141\textwidth}
		\includegraphics[height=1.525cm]{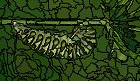}
	\end{subfigure}
	\begin{subfigure}[b]{0.141\textwidth}
		\includegraphics[height=1.525cm]{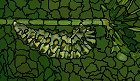}
	\end{subfigure}
	\begin{subfigure}[b]{0.141\textwidth}
		\includegraphics[height=1.525cm]{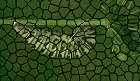}
	\end{subfigure}
	\caption{The influence of a low, on the left, and high, on the right, compactness parameter
	demonstrated on the caterpillar image from the \BSDS datasets using \SLIC and \CRS
	for $\K \approx 400$. Superpixel boundaries are depicted in black; best viewed in color.
	Superpixel algorithms providing a compactness parameter allow to trade boundary
	adherence for compactness.
	\textbf{Best viewed in color.}}
	\label{fig:experiments-qualitative-bsds500-compactness}
\end{figure*}

Visual quality is best determined by considering compactness, regularity and smoothness
on the one hand and boundary adherence on the other. Here, compactness refers to
the area covered by individual superpixels (as captured in Equation \eqref{eq:co});
regularity corresponds to both the superpixels' sizes and their arrangement;
and smootness refers to the superpixels' boundaries.
Figures \ref{fig:experiments-qualitative-bsds500-sbd-fash}
and \ref{fig:experiments-qualitative-nyuv2-sunrgbd} show results on all datasets.
We begin by discussing boundary adherence, in particular with regard to the difference
between superpixel and oversegmentation algorithms, before considering compactness,
smoothness and regularity.

The majority of algorithms provides solid adherence to important image boundaries, especially for large \K.
We consider the woman image -- in particular, the background --
and the caterpillar image in Figure~\ref{fig:experiments-qualitative-bsds500-sbd-fash}.
Algorithms with inferior boundary adherence are easily identified as those not capturing
the pattern in the background or the silhouette of the caterpillar:
\FH, \QS, \CIS, \PF, \PB, \TPS, \TP and \SEAW.
The remaining algorithms do not necessarily
capture all image details, as for example the woman's face, but important image boundaries
are consistently captured. 
We note that of the three evaluated oversegmentation algorithms, \ie \EAMS, \FH and \QS,
only \EAMS demonstrates adequate boundary adherence.
Furthermore, we observe that increasing \K results in more details
being captured by all algorithms. Notable algorithms regarding boundary adherence include \CRS, \ERS, \SEEDS, \ERGC and \ETPS.
These algorithms are able to capture even smaller details such as the coloring of the caterpillar or
elements of the woman's face.


Compactness strongly varies across algorithms and a compactness parameter
is beneficial to control the degree of compactness as it allows to gradually trade boundary adherence for compactness.
We consider the caterpillar image in Figure~\ref{fig:experiments-qualitative-bsds500-sbd-fash}.
\TP, \RW, \W, and \PF are examples for algorithms not providing a compactness parameter.
While \TP generates very compact superpixels and \RW tends to resemble grid-like superpixels, \W and \PF generate highly
non-compact superpixels. In this regard, compactness depends on algorithm and
implementation details (\eg grid-like initialization) and varies across algorithms.
For algorithms providing control over the compactness of the generated superpixels,
we find that parameter optimization has strong impact on compactness.
Examples are \CRS, \LSC, \ETPS and \ERGC showing highly irregular superpixels,
while \SLIC, \CCS, \VC and \WP generate more compact superpixels.
For \DASP and \VCCS, requiring depth information, similar observations can be made
on the kitchen image in Figure \ref{fig:experiments-qualitative-nyuv2-sunrgbd}.
Inspite of the influence of parameter optimization, we find that a compactness parameter
is beneficial. This can best be observed in Figure~\ref{fig:experiments-qualitative-bsds500-compactness}, showing superpixels
generated by \SLIC and \CRS for different degrees of compactness. We observe that compactness
can be increased while only gradually sacrificing boundary adherence.

We find that compactness does not necessarily induce regularity and smoothness;
some algorithms, however, are able to unite compactness, regularity and smoothness.
Considering the sea image in Figure~\ref{fig:experiments-qualitative-bsds500-sbd-fash}
for \CIS and \TP, we observe that compact superpixels are not necessarily arranged regularly.
Similarly, compact superpixels do not need to exhibit smooth boundaries, as can be seen for \PB.
On the other hand, compact superpixels are often generated in a regular fashion,
as can be seen for many algorithms providing a compactness parameter such as \SLIC, \VC and \CCS.
In such cases, compactness also induces smoother and more regular superpixels.
We also observe that many algorithms exhibiting excellent boundary adherence such
as \CRS, \SEEDS or \ETPS generate highly irregular and non-smooth superpixels.
These observations also justify the separate consideration of compactness,
regularity and smoothness to judge visual quality.
While the importance of compactness, regularity and smoothness may depend on the
application at hand, these properties represent the trade-off between
abstraction from and sensitivity to low-level image content which
is inherent to all superpixel algorithms.

In conclusion, we find that the evaluated path-based and density-based algorithms as well as oversegmentation algorithms show
inferior visual quality. On the other hand, clustering-based, contour evolution
and iterative energy optimization algorithms mostly demonstrate good boundary adherence
and some provide a compactness parameter, \eg \SLIC, \ERGC and \ETPS. Graph-based algorithms
show mixed results -- algorithms such as \FH, \CIS and \PB show inferior
boundary adherence, while \ERS, \RW, \NC and \POISE exhibit better boundary adherence.
However, good boundary adherence, especially regarding details in the image, often
comes at the price of lower compactness, regularity and/or smoothness as can be seen for \ETPS and \SEEDS.
Furthermore, compactness, smoothness and regularity are not necessarily linked and should be discussed separately.

\subsubsection{Compactness}
\label{subsubsec:experiments-qualitative-compactness}

\begin{figure}[t]
	\centering
	\input{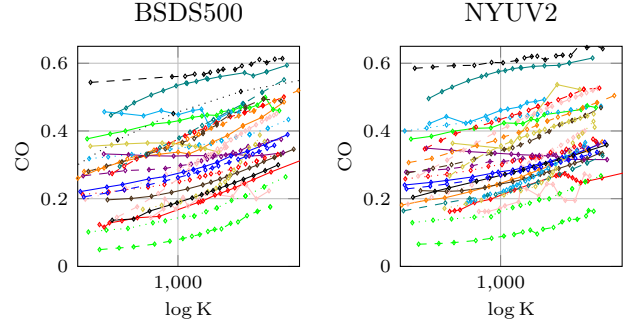}
	\caption{\CO on the \BSDS and \NYU datasets. Considering Figures \ref{fig:experiments-qualitative-bsds500-sbd-fash}
	and \ref{fig:experiments-qualitative-nyuv2-sunrgbd}, \CO appropriately reflects compactness.
	However, it does not take into account other aspects of visual quality such as regularity and smoothness.
	Therefore, we find that \CO is of limited use in a quantitative assessment of visual quality.
	\textbf{Best viewed in color.}}
	\label{fig:experiments-qualitative-compactness}
    \vskip 12px
	{\scriptsize
		\begin{tabularx}{0.475\textwidth}{X X X l}
			\ref{plot:w} \W &
			\ref{plot:eams} \EAMS &
			\ref{plot:nc} \NC &
			\ref{plot:fh} \FH\\
			\ref{plot:rw} \RW &
			\ref{plot:qs} \QS &
			\ref{plot:pf} \PF &
			\ref{plot:tp} \TP\\
			\ref{plot:cis} \CIS &
			\ref{plot:slic} \SLIC &
			\ref{plot:crs} \CRS &
			\ref{plot:ers} \ERS\\
			\ref{plot:pb} \PB &
			\ref{plot:seeds} \SEEDS &
			\ref{plot:tps} \TPS &
			\ref{plot:vc} \VC\\
			\ref{plot:ccs} \CCS &
			\ref{plot:cw} \CW &
			\ref{plot:ergc} \ERGC &
			\ref{plot:mss} \MSS\\
			\ref{plot:preslic} \preSLIC &
			\ref{plot:wp} \WP &
			\ref{plot:etps} \ETPS &
			\ref{plot:lsc} \LSC\\
			\ref{plot:poise} \POISE &
			\ref{plot:seaw} \SEAW &
		\end{tabularx}
	}

\end{figure}

\CO measures compactness, however, does not reflect regularity or smoothness;
therefore, \CO is not sufficient to objectively judge visual quality. We
consider Figure \ref{fig:experiments-qualitative-compactness}, showing \CO on
the \BSDS and \NYU datasets, and we observe that \CO correctly measures compactness. For example,
\WPr, \TPr and \CISr, exhibiting high \CO, also present very compact superpixels
in Figures \ref{fig:experiments-qualitative-bsds500-sbd-fash} and \ref{fig:experiments-qualitative-nyuv2-sunrgbd}.
However, these superpixels do not necessarily have to be visually appealing,
\ie may lack regularity and/or smoothness.
This can exemplarily be seen for \TPS, exhibiting high compactness bur poor regularity,
or \PB showing high compactness but inferior smoothness.
Overall, we find that \CO should not be considered isolated from a
qualitative evaluation.

\subsubsection{Depth}
\label{subsubsec:experiments-qualitative-depth}

Depth information helps superpixels resemble the 3D-structure within the image.
Considering Figure \ref{fig:experiments-qualitative-nyuv2-sunrgbd}, in particular
both images for \DASP and \VCCS, we deduce that depth information may be beneficial
for superpixels to resemble the 3D-structure of a scene. For example, when considering
planar surfaces (e.g. the table) in both images from Figure \ref{fig:experiments-qualitative-nyuv2-sunrgbd} for
\DASP, we clearly see that the superpixels easily align with the surface in a way perceived as 3-dimensional.
For \VCCS, this effect is less observable which may be due to the compactness parameter.

\subsection{Quantitative}
\label{subsec:experiments-quantitative}

\begin{figure*}
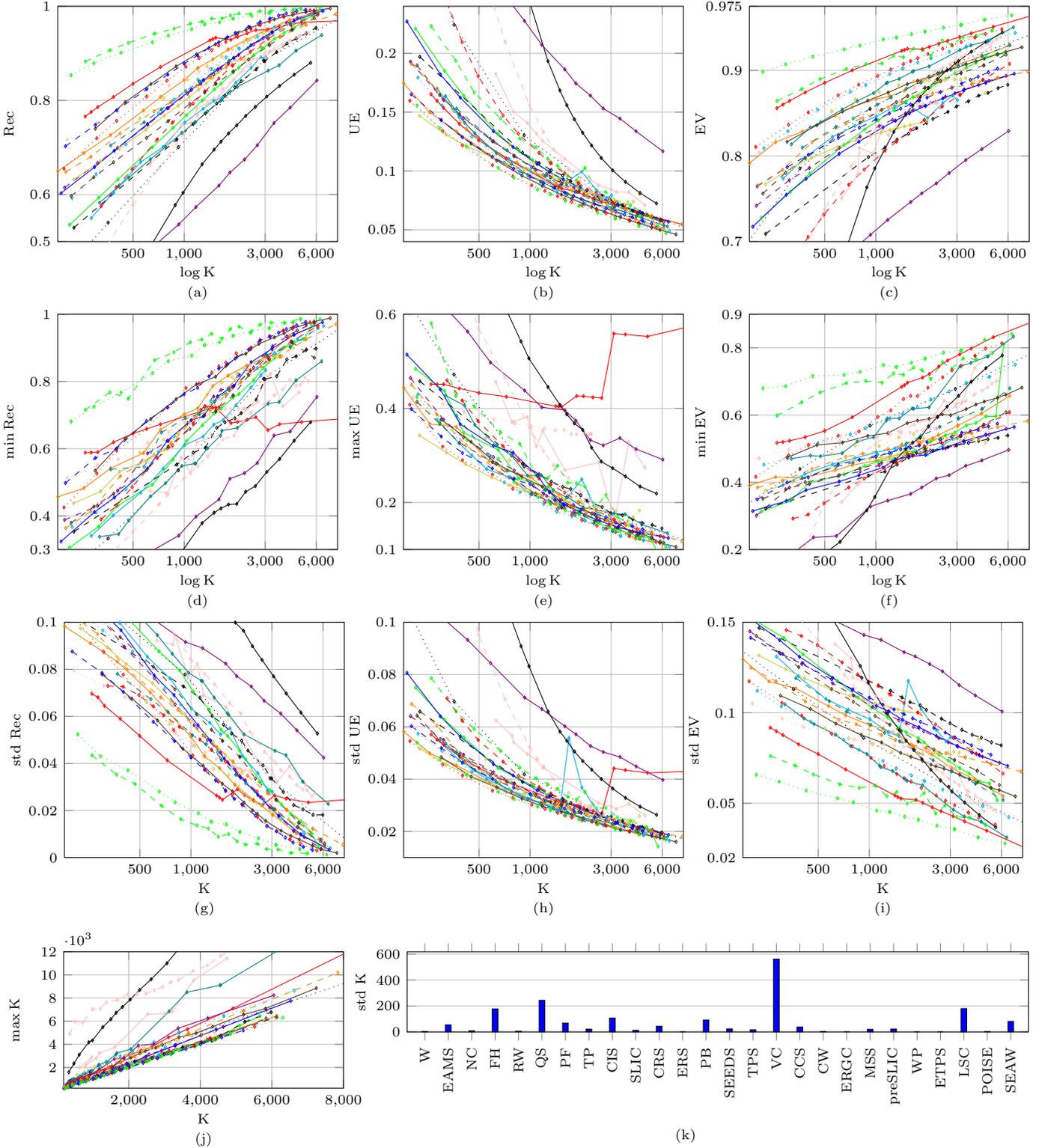

	\centering
	\vspace{-12px}
	\input{plots/quantitative-bsds500-avg-min-max}\\[-4px]
   	\input{plots/quantitative-bsds500-std}\\[-8px]
   	\input{plots/quantitative-bsds500-k}\\[-4px]
    \caption{Quantitative experiments on the \BSDS dataset; remember that \K denotes the number of generated superpixels.
    \Rec (higher is better) and \UE (lower is better) give a concise overview of the performance with respect to ground
    truth. In contrast, \EV (higher is better) gives a ground truth independent view on performance.
    While top-performers as well as poorly performing algorithms are easily
    identified, we provide more find-grained experimental results by considering
    $\min\Rec$, $\max\UE$ and $\min\EV$. These statistics additionally can be used
    to quantity the stability of superpixel algorithms. In particular, stable
	algorithms are expected to exhibit monotonically improving $\min\Rec$, $\max\UE$ and $\min\EV$.
	The corresponding $\text{std }\Rec$, $\text{std }\UE$ and $\text{std }\EV$ as
	well as $\max\K$ and $\text{std }\K$ help to identify stable algorithms.
	\textbf{Best viewed in color.}}
    \label{fig:experiments-quantitative-bsds500}
	\vskip 12px
	{\scriptsize
		\begin{tabularx}{\textwidth}{X X X X X X X X l}
			\ref{plot:w} \W &
			\ref{plot:eams} \EAMS &
			\ref{plot:nc} \NC &
			\ref{plot:fh} \FH &
			\ref{plot:rw} \RW &
			\ref{plot:qs} \QS &
			\ref{plot:pf} \PF &
			\ref{plot:tp} \TP &
			\ref{plot:cis} \CIS \\
			\ref{plot:slic} \SLIC &
			\ref{plot:crs} \CRS &
			\ref{plot:ers} \ERS &
			\ref{plot:pb} \PB &
			\ref{plot:seeds} \SEEDS &
			\ref{plot:tps} \TPS &
			\ref{plot:vc} \VC &
			\ref{plot:ccs} \CCS &
			\ref{plot:cw} \CW \\
			\ref{plot:ergc} \ERGC &
			\ref{plot:mss} \MSS &
			\ref{plot:preslic} \preSLIC &
			\ref{plot:wp} \WP &
			\ref{plot:etps} \ETPS &
			\ref{plot:lsc} \LSC &
			\ref{plot:poise} \POISE &
			\ref{plot:seaw} \SEAW &
		\end{tabularx}
	}

\end{figure*}
\begin{figure*}
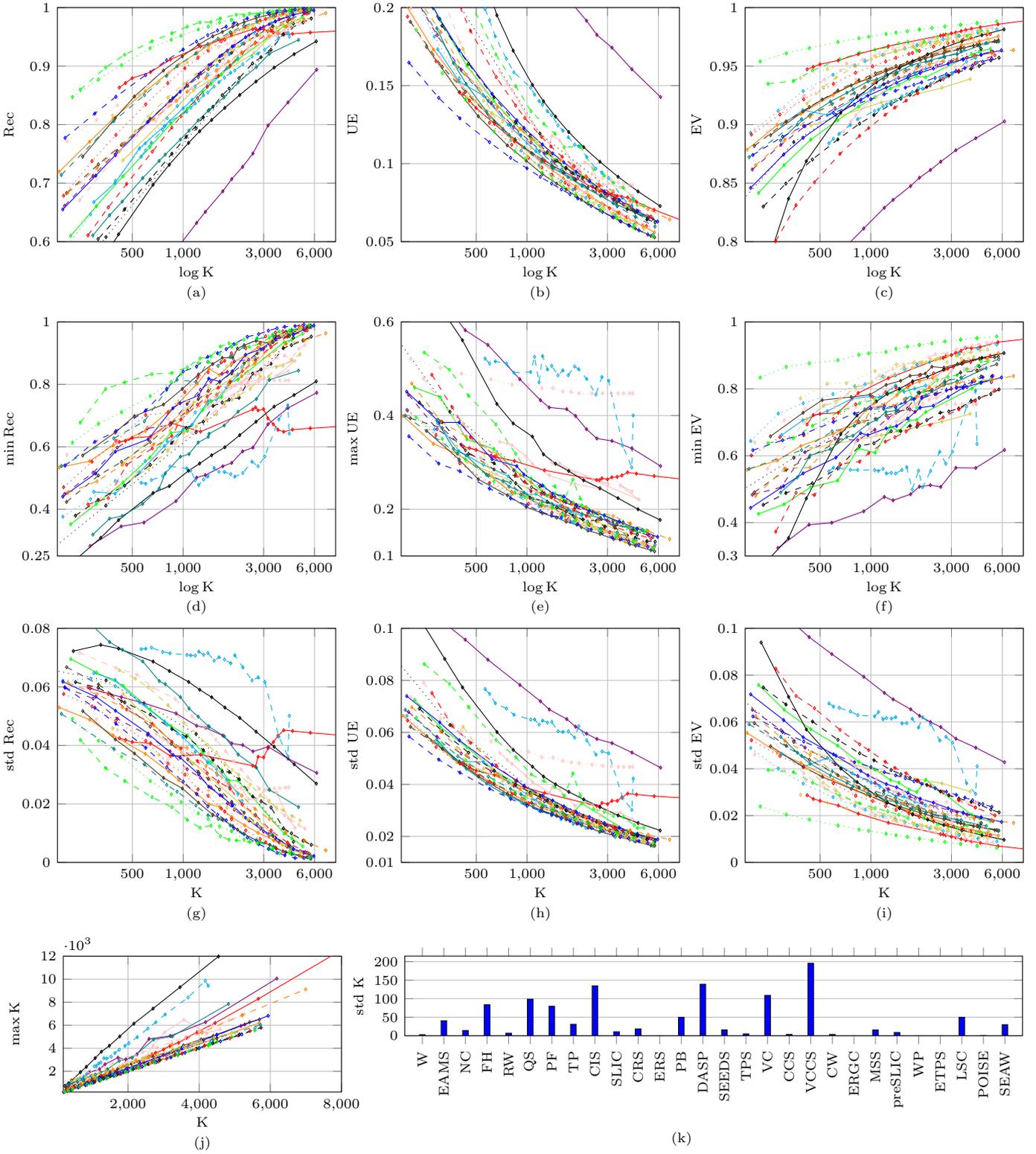

	\centering
	\vspace{-12px}
	\input{plots/quantitative-nyuv2-avg-min-max}\\[-4px]
	\input{plots/quantitative-nyuv2-std}\\[-8px]
	\input{plots/quantitative-nyuv2-k}\\[-4px]
	\caption{Quantitative results on the \NYU dataset; remember that \K denotes the number of generated superpixels.
    The presented experimental results complement the discussion in Figure \ref{fig:experiments-quantitative-bsds500}
    and show that most observations can be confirmed across datasets. Furthermore,
    \DASP and \VCCS show inferior performance suggesting that depth information does
    not necessarily improve performance.
	\textbf{Best viewed in color.}}
	\label{fig:experiments-quantitative-nyuv2}
	\vskip 12px
	{\scriptsize
		\begin{tabularx}{\textwidth}{X X X X X X X X X l}
			\ref{plot:w} \W &
			\ref{plot:eams} \EAMS &
			\ref{plot:nc} \NC &
			\ref{plot:fh} \FH &
			\ref{plot:rw} \RW &
			\ref{plot:qs} \QS &
			\ref{plot:pf} \PF &
			\ref{plot:tp} \TP &
			\ref{plot:cis} \CIS \\
			\ref{plot:slic} \SLIC &
			\ref{plot:crs} \CRS &
			\ref{plot:ers} \ERS &
			\ref{plot:pb} \PB &
			\ref{plot:dasp} \DASP &
			\ref{plot:seeds} \SEEDS &
			\ref{plot:tps} \TPS &
			\ref{plot:vc} \VC &
			\ref{plot:ccs} \CCS \\
			\ref{plot:vccs} \VCCS &
			\ref{plot:cw} \CW &
			\ref{plot:ergc} \ERGC &
			\ref{plot:mss} \MSS &
			\ref{plot:preslic} \preSLIC &
			\ref{plot:wp} \WP &
			\ref{plot:etps} \ETPS &
			\ref{plot:lsc} \LSC &
			\ref{plot:poise} \POISE \\
			\ref{plot:seaw} \SEAW & & & & & & & &
		\end{tabularx}
	}

\end{figure*}

Performance is determined by \Rec, \UE and \EV. In contrast to most authors,
we will look beyond metric averages. In particular, we consider the
minimum/maximum as well as the standard deviation to get an impression of the behavior of superpixel algorithms.
Furthermore, this allows us to quantify the stability of superpixel algorithms as
also considered by Neubert and Protzel in~\cite{NeubertProtzel:2013}.

\Rec and \UE offer a ground truth dependent overview to assess the performance of
superpixel algorithms. We consider
Figures \ref{subfig:experiments-quantitative-bsds500-rec.mean_min} and \ref{subfig:experiments-quantitative-bsds500-ue_np.mean_max},
showing \Rec and \UE on the \BSDS dataset. With respect to \Rec, we can easily identify top performing
algorithms, such as \ETPSr and \SEEDSr, as well as low performing algorithms,
such as \FHr, \QSr and \PFr. However, the remaining algorithms lie closely together
in between these two extremes, showing (apart from some exceptions) similar performance
especially for large~\K. Still, some algorithms perform consistently better than others,
as for example \ERGCr, \SLICr, \ERSr and \CRSr. For \UE, low performing algorithms,
such as \PFr or \QSr, are still easily identified while the remaining algorithms
tend to lie more closely together. Nevertheless, we can identify algorithms consistently
demonstrating good performance, such as \ERGCr, \ETPSr, \CRSr, \SLICr and \ERSr.
On the \NYU dataset, considering Figures \ref{subfig:experiments-quantitative-nyuv2-rec.mean[0]} and \ref{subfig:experiments-quantitative-nyuv2-ue_np.mean[0]},
these observations can be confirmed except for minor differences as for example the
excellent performance of \ERS regarding \UE or the better performance of \QS regarding \UE.
Overall, \Rec and \UE provide a quick overview of superpixel algorithm performance
but might not be sufficient to reliably discriminate superpixel algorithms.

In contrast to \Rec and \UE, \EV offers a ground truth independent assessment of superpixel algorithms.
Considering Figure \ref{subfig:experiments-quantitative-bsds500-ev.mean_min}, showing \EV on the \BSDS dataset,
we observe that algorithms are dragged apart and even for large \K
significantly different \EV values are attained. This suggests, that considering
ground truth independent metrics may be beneficial for comparison. However, \EV
cannot replace \Rec or \UE, as we can observe when comparing to
Figures \ref{subfig:experiments-quantitative-bsds500-rec.mean_min} and \ref{subfig:experiments-quantitative-bsds500-ue_np.mean_max},
showing \Rec and \UE on the \BSDS dataset; in particular
\QSr, \FHr and \CISr are performing significantly better with respect to \EV than regarding \Rec and \UE.
This suggests that \EV may be used to identify poorly performing algorithms, such as \TPSr, \PFr, \PBr or \NCr.
On the other hand, \EV is not necessarily suited to identify well-performing algorithms
due to the lack of underlying ground truth.
Overall, \EV is suitable to complement the view provided by \Rec and \UE,
however, should not be considered in isolation.

The stability of superpixel algorithms can be quantified by $\min\Rec$, $\max\UE$ and $\min\EV$ considering the behavior for increasing \K.
We consider Figures \ref{subfig:experiments-quantitative-bsds500-rec.min_min}, \ref{subfig:experiments-quantitative-bsds500-ue_np.max_max}
and \ref{subfig:experiments-quantitative-bsds500-ev.min_min},
showing $\min\Rec$, $\max\UE$ and $\min\EV$ on the \BSDS dataset. We define the stability of superpixel algorithms
as follows: an algorithm is considered stable if performance monotonically increases with \K
(\ie monotonically increasing \Rec and \EV and monotonically decreasing \UE).
Furthermore, these experiments can be interpreted as empirical bounds on the performance.
For example algorithms such as \ETPSr, \ERGCr, \ERSr, \CRSr and \SLICr can be considered stable and provide good bounds.
In contrast, algorithms such as \EAMSr, \FHr, \VCr or \POISEr are punished by
considering $\min\Rec$, $\max\UE$ and $\min\EV$ and cannot be described as stable.
Especially oversegmentation algorithms show poor stability. Most strikingly,
\EAMS seems to perform especially poorly on at least one image from the \BSDS dataset.
Overall, we find that $\min\Rec$, $\max\UE$ and $\min\EV$ appropriately reflect
the stability of superpixel algorithms.

The minimum/maximum of \Rec, \UE and \EV captures lower/upper bounds on performance.
In contrast, the corresponding standard deviation can be thought of as the expected
deviation from the average performance.
We consider Figures \ref{subfig:appendix-experiments-bsds500-rec.std[0]},
\ref{subfig:appendix-experiments-bsds500-ue_np.std[0]} and \ref{subfig:appendix-experiments-bsds500-ev.std[0]}
showing the standard deviation of \Rec, \UE and \EV on the \BSDS dataset.
We can observe that in many cases good performing algorithms such as \ETPS, \CRS, \SLIC or \ERS
also demonstrate low standard deviation. Oversegmentation algorithms, on the other hand, show higher standard deviation
-- together with algorithms such as \PF, \TPS, \VC, \CIS and \SEAW.
In this sense, stable algorithms can also be identified by low and monotonically decreasing standard deviation.

The variation in the number of generated superpixels is an important aspect
for many superpixel algorithms. In particular, high standard deviation in the number of generated superpixels can be related to
poor performance regarding \Rec, \UE and \EV. We find that superpixel algorithms ensuring that
the desired number of superpixels is met within appropriate bounds are preferrable. We consider
Figures \ref{subfig:experiments-quantitative-bsds500-sp.max[0]} and \ref{subfig:experiments-quantitative-bsds500-sp.std[0]},
showing $\max\K$ and $\text{std }\K$ for $\K \approx 400$ on the \BSDS dataset. Even after enforcing connectivity
as described in Section \ref{subsec:parameter-optimization-connectivity}, we observe
that several implementations are not always able to meet the desired number of superpixels
within acceptable bounds. Among these algorithms are \QSr, \VCr, \FHr, \CISr and \LSCr.
Except for the latter case, this can be related to poor performance with respect
to \Rec, \UE and~\EV. Conversely, considering algorithms such as \ETPSr, \ERGCr or \ERSr
which guarantee that the desired number of superpixels is met exactly, this can be
related to good performance regarding these metrics. To draw similar conclusions
for algorithms utilizing depth information, \ie \DASP and \VCCS,
the reader is encouraged to consider
Figures \ref{subfig:experiments-quantitative-nyuv2-sp.max[0]} and \ref{subfig:experiments-quantitative-nyuv2-sp.std[0]},
showing $\max\K$ and $\text{std }\K$ for $\K\approx 400$ on the \NYU dataset.
We can conclude that superpixel algorithms with low standard deviation in the number
of generated superpixels are showing better performance in many cases.

Finally, we discuss the proposed metrics \ARec, \AUE and \AEV (computed as the area
below the $\MR = (1 - \Rec)$, \UE and $\UEV = (1 - \EV)$ curves within the interval $[\K_{\min}, \K_{\max}] = [200,5200]$, \ie lower is better).
We find that these metrics appropriately reflect and summarize the performance of superpixel
algorithms independent of \K. As can be seen in Figure \ref{subfig:experiments-quantitative-bsds500-average},
showing \ref{plot:experiments-quantitative-bsds500-average-rec} \ARec, \ref{plot:experiments-quantitative-bsds500-average-ue_np}
\AUE and \ref{plot:experiments-quantitative-bsds500-average-ev} \AEV on the \BSDS dataset, most of the
previous observations can be confirmed. For example, we exemplarily consider \SEEDSr
and observe low \ARec and \AEV which is confirmed by
Figures \ref{subfig:experiments-quantitative-bsds500-rec.mean_min} and \ref{subfig:experiments-quantitative-bsds500-ev.mean_min},
showing \Rec and \EV on the \BSDS dataset, where \SEEDSr consistently outperforms all algorithms except for \ETPSr.
However, we can also observe higher \AUE compared to algorithms such as
\ETPSr, \ERSr or \CRSr wich is also consistent with Figure \ref{subfig:experiments-quantitative-bsds500-ue_np.mean_max},
showing \UE on the \BSDS dataset. We conclude, that \ARec, \AUE and \AEV give an easy-to-understand summary of algorithm performance.
Furthermore, \ARec, \AUE and \AEV can be used to rank the different
algorithms according to the corresponding metrics; we will follow up on this idea in Section \ref{subsec:experiments-ranking}.

The observed \ARec, \AUE and \AEV also properly reflect the difficulty of the different datasets. We
consider Figure \ref{fig:experiments-quantitative-avg} showing \ref{plot:experiments-quantitative-bsds500-average-rec} \ARec,
\ref{plot:experiments-quantitative-bsds500-average-ue_np} \AUE and \ref{plot:experiments-quantitative-bsds500-average-ev}  \AEV for all five datasets.
Concentrating on \SEEDS and \ETPS, we see that the relative
performance (\ie the performance of \SEEDS compared to \ETPS) is consistent across
datasets; \SEEDS usually showing higher \AUE while \ARec and \AEV are usually similar.
Therefore, we observe that these metrics can be used to characterize
the difficulty and ground truth of the datasets. For example, considering
the \Fash dataset, we observe very high \AEV compared
to the other datasets, while \ARec and \AUE are usually very low. This can be
explained by the ground truth shown in Figure~\ref{subfig:datasets-fash},
\ie the ground truth is limited to the foreground (in the case of Figure \ref{subfig:datasets-fash}, the woman),
leaving even complicated background unannotated. Similar arguments can be developed
for the consistently lower \ARec, \AUE and \AEV for the \NYU and \SUNRGBD datasets compared to the \BSDS dataset.
For the \SBD dataset, lower \ARec, \AUE and \AEV can also be explained by the smaller average image size.

In conclusion, \ARec, \AUE and \AEV accurately reflect the performance of superpixel
algorithms and can be used to judge datasets. Across the different datasets, path-based
and density-based algorithms perform poorly, while the remaining classes show mixed
performance. However, some iterative energy optimization, clustering-based and
graph-based algorithms such as \ETPS, \SEEDS, \CRS, \ERS and \SLIC show favorable performance.

\begin{figure*}
	\centering
	\begin{subfigure}[b]{\fullthreethree\textwidth}\phantomsubcaption\label{subfig:experiments-quantitative-bsds500-average}
	\begin{tikzpicture}
		\begin{axis}[EQBSDS500Avg,%
				symbolic x coords = {
					W,
					EAMS,
					NC,
					FH,
					RW,
					QS,
					PF,
					TP,
					CIS,
					SLIC,
					CRS,
					ERS,
					PB,
					(DASP),
					SEEDS,
					TPS,
					VC,
					CCS,
					(VCCS),
					CW,
					ERGC,
					MSS,
					preSLIC,
					WP,
					ETPS,
					LSC,
					POISE,
					SEAW,
				},
			]

			\addplot[fill=blue] coordinates {
				(CCS,9.9318965816336)
				(SEEDS,2.28232472)
				(SLIC,8.6849153136699)
				(RW,13.347694664807)
				(CW,10.508734794086)
				(TP,16.91966442)
				(POISE,13.09785248)
				(FH,15.02527176)
				(EAMS,6.9713463075023)
				(CRS,6.4499133686543)
				(SEAW,18.718694626414)
				(ERGC,7.0299239643847)
				(PF,32.339046914181)
				(TPS,16.608716929614)
				(NC,14.5729878)
				(VC,16.099002920902)
				(PB,12.475172916609)
				(preSLIC,10.364926165195)
				(W,10.512659928738)
				(LSC,9.11646619)
				(WP,12.859419948232)
				(QS,26.357822257892)
				(CIS,17.742002940095)
				(ERS,6.227106)
				(MSS,11.451229616209)
				(ETPS,2.9666871716831)
				((DASP),0)
				((VCCS),0)
			};
			\label{plot:experiments-quantitative-bsds500-average-rec}

			\addplot[fill=red] coordinates {
				(CCS,7.5289953858794)
				(SEEDS,9.446739759)
				(SLIC,7.6497140156322)
				(RW,8.6756965619526)
				(CW,8.4402624332611)
				(TP,9.271960248)
				(POISE,8.924878757)
				(FH,10.243686489)
				(EAMS,8.2105337749209)
				(CRS,7.3448022077077)
				(SEAW,10.305148679693)
				(ERGC,7.1872745931238)
				(PF,17.356643633834)
				(TPS,8.9472800577507)
				(NC,9.334483981)
				(VC,10.953437795205)
				(PB,9.3881740126911)
				(preSLIC,8.1143337244377)
				(W,8.7587242552172)
				(LSC,8.969104772)
				(WP,8.3393118766649)
				(QS,16.426937818043)
				(CIS,7.883995044819)
				(ERS,7.6528118)
				(MSS,8.2658572868153)
				(ETPS,7.0390794600518)
				((DASP),0)
				((VCCS),0)
			};
			\label{plot:experiments-quantitative-bsds500-average-ue_np}

			\addplot[fill=green] coordinates {
				(CCS,8.810595522855)
				(SEEDS,7.32645749)
				(SLIC,10.695688557583)
				(RW,12.150049603053)
				(CW,13.965614906343)
				(TP,14.73410334)
				(POISE,14.2154631)
				(FH,12.70897273)
				(EAMS,7.0888362851786)
				(CRS,10.759921739142)
				(SEAW,11.431681615602)
				(ERGC,8.1405080708485)
				(PF,23.576251534528)
				(TPS,16.153750090356)
				(NC,16.62427233)
				(VC,11.799903143852)
				(PB,15.551225371015)
				(preSLIC,10.942140607989)
				(W,14.204407365309)
				(LSC,11.99522686)
				(WP,12.032242915515)
				(QS,14.490709318973)
				(CIS,9.332430822)
				(ERS,13.001537)
				(MSS,13.02655533315)
				(ETPS,5.3336737754583)
				((DASP),0)
				((VCCS),0)
			};
			\label{plot:experiments-quantitative-bsds500-average-ev}

		\end{axis}
	\end{tikzpicture}
\end{subfigure}
\begin{tikzpicture}[overlay]
	\node at (-9.25, 2.15){\rotatebox{90}{\small no depth}};
	\node at (-6.2, 2.15){\rotatebox{90}{\small no depth}};
\end{tikzpicture}\\
	\begin{subfigure}[b]{0.985\textwidth}\phantomsubcaption\label{subfig:experiments-quantitative-nyuv2-average}
	\begin{tikzpicture}
		\begin{axis}[EQNYUV2Avg,%
				symbolic x coords = {
					W,
					EAMS,
					NC,
					FH,
					RW,
					QS,
					PF,
					TP,
					CIS,
					SLIC,
					CRS,
					ERS,
					PB,
					DASP,
					SEEDS,
					TPS,
					VC,
					CCS,
					VCCS,
					CW,
					ERGC,
					MSS,
					preSLIC,
					WP,
					ETPS,
					LSC,
					POISE,
					SEAW,
				},
			]

			\addplot[fill=blue] coordinates {
				(CCS,7.9704040194755)
				(SEEDS,1.99918493)
				(SLIC,4.8869580065139)
				(RW,9.7872709)
				(CW,7.0241612257619)
				(TP,12.01724336)
				(POISE,6.71664896)
				(FH,7.84434987)
				(EAMS,5.7893706432839)
				(CRS,4.2017993563281)
				(SEAW,14.491755414489)
				(ERGC,5.3342177711828)
				(PF,26.952844206793)
				(TPS,12.578696618482)
				(NC,9.39717382)
				(VC,8.892891812)
				(PB,9.5242039277316)
				(VCCS,10.5670046)
				(preSLIC,4.7838102442251)
				(W,7.0637844541329)
				(LSC,4.90401351)
				(WP,7.5084420846341)
				(QS,16.046153933985)
				(CIS,13.38194051)
				(ERS,3.477366)
				(DASP,7.43460309)
				(MSS,8.5582389358142)
				(ETPS,2.4733715019512)
			};
			\label{plot:experiments-quantitative-nyuv2-average-rec}
			
			\addplot[fill=red] coordinates {
				(CCS,8.2978239111225)
				(SEEDS,11.001924982)
				(SLIC,8.4907755437663)
				(RW,9.441367153)
				(CW,9.2316959318342)
				(TP,9.73918887)
				(POISE,10.01700179)
				(FH,10.094505885)
				(EAMS,9.1389145184757)
				(CRS,7.9479839954883)
				(SEAW,10.778732686267)
				(ERGC,8.3641164685591)
				(PF,20.064094800719)
				(TPS,9.372713628934)
				(NC,9.32182317)
				(VC,8.8199022232361)
				(PB,10.00583010179)
				(VCCS,10.94676269)
				(preSLIC,9.0053832940137)
				(W,9.5520796412727)
				(LSC,8.527777268)
				(WP,8.9339077710976)
				(QS,12.273312979277)
				(CIS,8.90993327)
				(ERS,7.7164086)
				(DASP,8.315646668)
				(MSS,9.6461625872257)
				(ETPS,8.0038928736829)
			};
			\label{plot:experiments-quantitative-nyuv2-average-ue_np}
			
			\addplot[fill=green] coordinates {
				(CCS,3.2656276782785)
				(SEEDS,2.9067482)
				(SLIC,4.3347220783406)
				(RW,5.90116557)
				(CW,5.4825582152504)
				(TP,5.94252827)
				(POISE,5.46585384)
				(FH,5.15784339)
				(EAMS,2.6588718157037)
				(CRS,4.5231009601953)
				(SEAW,5.7570158156)
				(ERGC,3.7341660154839)
				(PF,14.434696904625)
				(TPS,7.0409476476568)
				(NC,7.22154213)
				(VC,3.1121424192048)
				(PB,7.3257554894976)
				(VCCS,6.61132094)
				(preSLIC,4.8991294197023)
				(W,5.7317004976434)
				(LSC,3.39076539)
				(WP,4.4520422653659)
				(QS,4.7715113993379)
				(CIS,4.81193817)
				(ERS,6.250889)
				(DASP,3.3488285)
				(MSS,5.2941333502832)
				(ETPS,1.8714741002439)
			};
			\label{plot:experiments-quantitative-nyuv2-average-ev}
			
		\end{axis}
	\end{tikzpicture}
\end{subfigure}\\
	\begin{subfigure}[b]{0.985\textwidth}\phantomsubcaption\label{subfig:experiments-quantitative-sbd-average}
	\begin{tikzpicture}
		\begin{axis}[EQSBDAvg,%
				symbolic x coords = {
					W,
					EAMS,
					(NC),
					FH,
					RW,
					QS,
					PF,
					TP,
					CIS,
					SLIC,
					CRS,
					ERS,
					PB,
					(DASP),
					SEEDS,
					TPS,
					VC,
					CCS,
					(VCCS),
					CW,
					ERGC,
					MSS,
					preSLIC,
					WP,
					ETPS,
					LSC,
					POISE,
					SEAW,
				},
			]

			\addplot[fill=blue] coordinates {
				(CCS,3.99207976)
				(SEEDS,0.98564913)
				(SLIC,3.8467011057143)
				(RW,6.5381300390722)
				(CW,4.6874497903046)
				(TP,12.66695784)
				(POISE,10.6843381)
				(FH,6.46660202)
				(EAMS,1.7266924017788)
				(CRS,3.3346620325286)
				(SEAW,9.5443596898462)
				(ERGC,3.2488854819788)
				(PF,18.64936024)
				(TPS,9.0273009484173)
				((NC),0)
				(VC,6.39442433)
				(PB,5.51861166)
				(preSLIC,4.3352076513208)
				(W,4.7278366937943)
				(LSC,5.4246496)
				(WP,5.5110872544954)
				(QS,12.26025829483)
				(CIS,8.60568384)
				(ERS,2.074466)
				(MSS,6.3959980483071)
				(ETPS,1.1757854151376)
				((DASP),0)
				((VCCS),0)
			};
			\label{plot:experiments-quantitative-sbd-average-rec}

			\addplot[fill=red] coordinates {
				(CCS,6.500144161)
				(SEEDS,7.255817146)
				(SLIC,6.6241608762857)
				(RW,7.4583036919485)
				(CW,7.2332045751798)
				(TP,9.126335872)
				(POISE,8.133637934)
				(FH,7.801094687)
				(EAMS,6.2440697203207)
				(CRS,6.3964248642735)
				(SEAW,8.146456334)
				(ERGC,6.5602665651458)
				(PF,12.47642349)
				(TPS,7.165171797705)
				((NC),0)
				(VC,7.076657975)
				(PB,7.431007376)
				(preSLIC,7.2020326533585)
				(W,7.4067354291667)
				(LSC,8.788028426)
				(WP,6.9608341891835)
				(QS,9.3828209431597)
				(CIS,6.567490704)
				(ERS,6.3167828)
				(MSS,7.6111817829679)
				(ETPS,6.2243865069174)
				((DASP),0)
				((VCCS),0)
			};
			\label{plot:experiments-quantitative-sbd-average-ue_np}

			\addplot[fill=green] coordinates {
				(CCS,5.1215566)
				(SEEDS,3.65680745)
				(SLIC,6.6596438680952)
				(RW,6.9632355873196)
				(CW,9.4167775171791)
				(TP,11.13864972)
				(POISE,10.16088613)
				(FH,6.05722211)
				(EAMS,3.5112406391239)
				(CRS,7.1839177763178)
				(SEAW,6.9888148935385)
				(ERGC,4.5890374901253)
				(PF,15.33460288)
				(TPS,9.9312681434532)
				((NC),)
				(VC,5.53762571)
				(PB,9.06619777)
				(preSLIC,7.4900418458491)
				(W,9.5496883188652)
				(LSC,9.69159198)
				(WP,7.2626703088073)
				(QS,5.6803576142068)
				(CIS,5.08452376)
				(ERS,8.269597)
				(MSS,9.1936434007485)
				(ETPS,2.9591285079816)
				((DASP),0)
				((VCCS),0)
			};
			\label{plot:experiments-quantitative-sbd-average-ev}

		\end{axis}
	\end{tikzpicture}
\end{subfigure}
\begin{tikzpicture}[overlay]
	\node at (-15.95,  1.925){\rotatebox{90}{\small failed}};
	\node at (-9.25, 2.15){\rotatebox{90}{\small no depth}};
	\node at (-6.2, 2.15){\rotatebox{90}{\small no depth}};
\end{tikzpicture}\\
	\begin{subfigure}[b]{0.985\textwidth}\phantomsubcaption\label{subfig:experiments-quantitative-sunrgbd-average}
	\begin{tikzpicture}
		\begin{axis}[EQSUNRGBDAvg,%
				symbolic x coords = {
					W,
					EAMS,
					(NC),
					FH,
					(RW),
					QS,
					PF,
					TP,
					CIS,
					SLIC,
					CRS,
					ERS,
					PB,
					DASP,
					SEEDS,
					TPS,
					VC,
					CCS,
					VCCS,
					CW,
					ERGC,
					MSS,
					preSLIC,
					WP,
					ETPS,
					LSC,
					POISE,
					(SEAW),
				},
			]

			\addplot[fill=blue] coordinates {
				(CCS,9.7793567967647)
				(SEEDS,2.3172278871707)
				(SLIC,5.9045951381443)
				(CW,8.301950156873)
				(TP,10.73901102)
				(POISE,8.10561845)
				(FH,10.229925891852)
				(EAMS,5.5189034036842)
				(CRS,5.8588099082847)
				(ERGC,6.1777726336)
				(PF,29.342998077107)
				(TPS,14.937158274754)
				(VC,10.412830284681)
				(PB,10.85350273)
				(VCCS,17.86737119)
				(preSLIC,6.1443742248148)
				(W,8.2592052550445)
				(LSC,6.7356841054902)
				(WP,8.93473851)
				(QS,17.416250791579)
				(CIS,14.52232126)
				(ERS,4.387237)
				(DASP,9.0420388)
				(MSS,10.080962459341)
				(ETPS,3.44468441)
				((SEAW),0)
				((NC),0)
				((RW),0)
			};
			\label{plot:experiments-quantitative-sunrgbd-average-rec}

			\addplot[fill=red] coordinates {
				(CCS,6.9436340856471)
				(SEEDS,8.5258507456325)
				(SLIC,7.0139590942577)
				(CW,7.4875824507232)
				(TP,7.512286646)
				(POISE,8.228683284)
				(FH,8.6631790703704)
				(EAMS,7.6068779476316)
				(CRS,6.7287454884014)
				(ERGC,6.7503710224)
				(PF,16.519507852642)
				(TPS,7.8340534858443)
				(VC,7.4476611429255)
				(PB,8.360526273)
				(VCCS,16.14082359)
				(preSLIC,7.5937574069206)
				(W,7.7706754118519)
				(LSC,7.0259231029804)
				(WP,7.331545901)
				(QS,10.196305483789)
				(CIS,7.28551162)
				(ERS,6.6387882)
				(DASP,7.339403359)
				(MSS,7.9004804341018)
				(ETPS,6.617327532)
				((SEAW),0)
				((NC),0)
				((RW),0)
			};
			\label{plot:experiments-quantitative-sunrgbd-average-ue_np}

			\addplot[fill=green] coordinates {
				(CCS,2.9527510860784)
				(SEEDS,2.3956850803252)
				(SLIC,4.2642592482474)
				(CW,5.0632259526441)
				(TP,5.22223075)
				(POISE,6.71430356)
				(FH,4.5491502197037)
				(EAMS,2.3456012263158)
				(CRS,4.3902226967581)
				(ERGC,2.9190980512)
				(PF,13.130088422932)
				(TPS,6.3351563696311)
				(VC,2.9665376725532)
				(PB,6.97417875)
				(VCCS,14.38189913)
				(preSLIC,5.3502755248677)
				(W,5.2309960849604)
				(LSC,3.1444705776471)
				(WP,3.85768184)
				(QS,4.2290703015789)
				(CIS,4.09318536)
				(ERS,6.138697)
				(DASP,3.31095176)
				(MSS,4.9316343176647)
				(ETPS,1.55775485)
				((SEAW),0)
				((NC),0)
				((RW),0)
			};
			\label{plot:experiments-quantitative-sunrgbd-average-ev}

			\end{axis}
	\end{tikzpicture}
\end{subfigure}
\begin{tikzpicture}[overlay]
	\node at (-16, 1.925){\rotatebox{90}{\small failed}};
	\node at (-14.75, 1.925){\rotatebox{90}{\small failed}};
	\node at (-0.725, 1.925){\rotatebox{90}{\small failed}};
\end{tikzpicture}\\
	\begin{subfigure}[b]{0.985\textwidth}\phantomsubcaption\label{subfig:experiments-quantitative-sbd-average}
	\begin{tikzpicture}
		\begin{axis}[EQFashAvg,%
				symbolic x coords = {
					W,
					EAMS,
					NC,
					FH,
					RW,
					QS,
					PF,
					TP,
					CIS,
					SLIC,
					CRS,
					ERS,
					PB,
					(DASP),
					SEEDS,
					TPS,
					VC,
					CCS,
					(VCCS),
					CW,
					ERGC,
					MSS,
					preSLIC,
					WP,
					ETPS,
					LSC,
					POISE,
					SEAW,
				},
			]

			\addplot[fill=blue] coordinates {
				(CCS,1.30381016)
				(SEEDS,0.27558563)
				(SLIC,1.2052600288462)
				(RW,3.7291897144478)
				(CW,1.9320466830252)
				(TP,2.90232317)
				(POISE,0.63749367)
				(FH,3.0623073551485)
				(EAMS,0.89636899)
				(CRS,0.96158964478261)
				(SEAW,5.1436382098319)
				(ERGC,0.95938017666667)
				(PF,15.903108370669)
				(TPS,4.0994467796099)
				(NC,2.25343566)
				(VC,2.69537224)
				(PB,3.16592995)
				(preSLIC,1.8012823817143)
				(W,2.0076433011981)
				(LSC,0.84386359)
				(WP,2.39111976)
				(QS,6.5004867285691)
				(CIS,2.99686849)
				(ERS,0.738335)
				(MSS,1.7895051124315)
				(ETPS,0.21114242)
				((DASP),0)
				((VCCS),0)
			};
			\label{plot:experiments-quantitative-fash-average-rec}

			\addplot[fill=red] coordinates {
				(CCS,2.764525675)
				(SEEDS,3.694062723)
				(SLIC,2.7143095554808)
				(RW,3.7020414663224)
				(CW,3.5731282445733)
				(TP,3.977292127)
				(POISE,2.579006278)
				(FH,3.9528368212529)
				(EAMS,2.6387901)
				(CRS,3.0353119784783)
				(SEAW,4.2277920126933)
				(ERGC,2.8178267316022)
				(PF,8.1170746999841)
				(TPS,2.637501723848)
				(NC,2.952651111)
				(VC,4.076593886)
				(PB,4.088234941)
				(preSLIC,3.1908578729905)
				(W,3.7307536424901)
				(LSC,2.876269579)
				(WP,3.277769837)
				(QS,5.4619018185004)
				(CIS,2.846133049)
				(ERS,3.068406)
				(MSS,3.7181266457414)
				(ETPS,2.486032333)
				((DASP),0)
				((VCCS),0)
			};
			\label{plot:experiments-quantitative-fash-average-ue_np}

			\addplot[fill=green] coordinates {
				(CCS,5.32973367)
				(SEEDS,4.18608671)
				(SLIC,5.7260763842308)
				(RW,8.1070688972537)
				(CW,8.7929110226793)
				(TP,9.261288)
				(POISE,8.75335052)
				(FH,7.7833847431379)
				(EAMS,6.895071)
				(CRS,7.1783888765217)
				(SEAW,7.9154503245378)
				(ERGC,5.3435673125806)
				(PF,15.863401454533)
				(TPS,10.028079727577)
				(NC,10.60275725)
				(VC,6.932652)
				(PB,10.76750559)
				(preSLIC,6.8505669954286)
				(W,8.978665027208)
				(LSC,5.86400029)
				(WP,7.23421236)
				(QS,9.3564948894804)
				(CIS,6.29582096)
				(ERS,8.031587)
				(MSS,8.1725618683075)
				(ETPS,3.05049199)
				((DASP),0)
				((VCCS),0)
			};
			\label{plot:experiments-quantitative-fash-average-ev}

			\end{axis}
	\end{tikzpicture}
\end{subfigure}
\begin{tikzpicture}[overlay]
	\node at (-9.25, 2.15){\rotatebox{90}{\small no depth}};
	\node at (-6.2, 2.15){\rotatebox{90}{\small no depth}};
\end{tikzpicture}
	\caption{\ARec, \AUE and \AEV (lower is better) on the used datasets.
    We find that \ARec, \AUE and \AEV appropriately
    summarize performance independent of the number of generated superpixels. Plausible
    examples to consider are top-performing algorithms such as \ETPS, \ERS, \SLIC or \CRS
    as well as poorly performing ones such as \QS and~\PF.
	\textbf{Best viewed in color.}}
	\label{fig:experiments-quantitative-avg}
\end{figure*}

\subsubsection{Depth}

Depth information does not necessarily improve performance regarding \Rec, \UE and \EV.
We consider Figures \ref{subfig:experiments-quantitative-nyuv2-rec.mean[0]},
\ref{subfig:experiments-quantitative-nyuv2-ue_np.mean[0]} and \ref{subfig:experiments-quantitative-nyuv2-ev.mean[0]}
presenting \Rec, \UE and \EV on the \NYU dataset. In particular, we consider \DASPr and \VCCSr.
We observe, that \DASPr consistently outperforms \VCCSr.
Therefore, we consider the performance of \DASPr and investigate whether depth information
improves performance. Note that \DASPr performs similar to \SLICr, exhibiting slightly
worse \Rec and slightly better \UE and \EV for large \K. However, \DASPr does not
clearly outperform \SLICr. As indicated in Section \ref{sec:algorithms}, \DASP and \SLIC are
both clustering-based algorithms. In particular, both algorithms are based on $k$-means
using color and spatial information and \DASPr additionally utilizes depth information.
This suggests that the clustering approach
does not benefit from depth information. We note that a similar line of thought can be
applied to \VCCS except that \VCCS directly operates within a point cloud, rendering the
comparison problematic. Still we conclude that depth information used in the form of
\DASP does not improve performance. This might be in contrast to
experiments with different superpixel algorithms, \eg a \SLIC variant using depth information
as in \cite{ZhangKanSchwingUrtasun:2013}. We suspect that regarding the used metrics, the number of superpixels ($\K = 200$)
and the used superpixel algorithm, the effect of depth information might be more pronounced 
in the experiments presented in \cite{ZhangKanSchwingUrtasun:2013} compared to ours.
Furthermore, it should be noted that our evaluation is carried out in the 2D image plane, 
which does not directly reflect the segmentation of point clouds.

\subsection{Runtime}
\label{subsec:experiments-runtime}

Considering runtime, we report CPU time\footnote{\label{foot:runtime}
    Runtimes have been taken on an Intel$^\text{\textregistered}$ Core\texttrademark\xspace i7-3770 @ $3.4GHz$, 64bit with 32GB RAM.
} excluding connected components but including color space conversions if applicable.
We made sure that no multi-threading or GPU computation were used. We begin by considering
runtime in general, with a glimpse on realtime applications, before considering iterative algorithms.

\begin{figure}[t]
    \centering
    \input{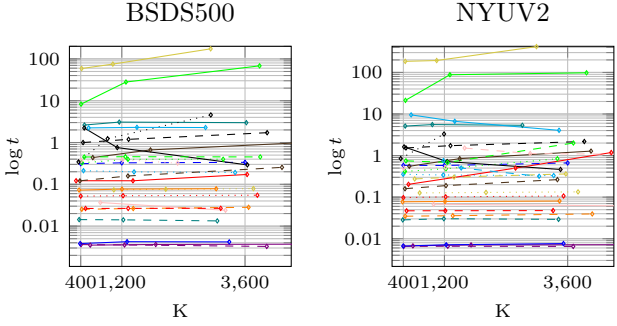}
    \caption{Runtime in seconds on the \BSDS and \NYU datasets.
    Watershed-based, some clustering-based algorithms as well as \PF offer
    runtimes below $100ms$. In the light of realtime applications, \CW, \W
    and \PF even provide runtimes below $10ms$. However, independent of the
    application at hand, we find runtimes below $1s$ beneficial for using
    superpixel algorithms as pre-processing tools.
    \textbf{Best viewed in color.}}
    \label{fig:experiments-runtime}
\end{figure}

We find that some well performing algorithms can be run at (near) realtime. We
consider Figure \ref{fig:experiments-runtime}
showing runtime in seconds $t$ on the \BSDS (image size $481 \times 321$) and \NYU (image size $608 \times 448$) datasets.
Concretely, considering the watershed-based algorithms \Wr and \CWr, we can report runtimes below $10\text{ms}$
on both datasets, corresponding to roughly~$100\text{fps}$. Similarly, \PFr
runs at below $10\text{ms}$. Furthermore, several algorithms, such as
\SLICr, \ERGCr, \FHr, \PBr, \MSSr and \preSLICr provide runtimes below $80\text{ms}$
and some of them are iterative, \ie reducing the number of iterations may further
reduce runtime. However, using the convention that realtime corresponds to
roughly $30\text{fps}$, this leaves \preSLIC and \MSS on the larger images
of the \NYU dataset. However, even without explicit runtime requirements, we find runtimes
below $1\text{s}$ per image to be beneficial for using superpixel algorithms as
pre-processing tool, ruling out \TPS, \CIS, \SEAW, \RW and \NC.
Overall, several superpixel algorithms provide runtimes appropriate for pre-processing tools;
realtime applications are still limited to a few fast algorithms.

\begin{figure}[t]
    \centering
    \begin{subfigure}[b]{\halfthreeone\textwidth}\phantomsubcaption\label{subfig:experiments-iterations-bsds500-rec}
	\begin{tikzpicture}
		\begin{axis}[EQBSDS500ItRec]
			\addplot[CCS] coordinates{
				(1,0.405664)
				(5,0.599782)
				(10,0.671463)
				(25,0.696412)
			};
			\addplot[CIS] coordinates{
				(1,0.668558)
				(3,0.636139)
				(6,0.635233)
			};
			\addplot[CRS] coordinates{
				(1,0.818608)
				(3,0.833132)
				(6,0.834531)
			};
			\addplot[ETPS] coordinates{
				(1,0.897415)
				(5,0.901659)
				(10,0.901555)
				(25,0.901663)
			};
			\addplot[LSC] coordinates{
				(1,0.833689)
				(5,0.843883)
				(10,0.843333)
				(25,0.841751)
			};
			\addplot[SLIC] coordinates{
				(1,0.703309)
				(5,0.722822)
				(10,0.725153)
				(25,0.726167)
			};
			\addplot[PRESLIC] coordinates{
				(1,0.694801)
				(5,0.704329)
				(10,0.704329)
				(25,0.704329)
			};
			\addplot[RESEEDS] coordinates{
				(1,0.862949)
				(5,0.90843)
				(10,0.916129)
				(25,0.918462)
			};
			\addplot[SEEDS] coordinates{
				(1,0.771216)
				(5,0.872416)
				(10,0.901845)
				(25,0.922132)
			};
			\addplot[VLSLIC] coordinates{
				(1,0.793577)
				(5,0.819088)
				(10,0.815093)
				(25,0.813072)
			};
		\end{axis}
	\end{tikzpicture}
\end{subfigure}
\begin{subfigure}[b]{\halfthreeone\textwidth}\phantomsubcaption\label{subfig:experiments-iterations-bsds500-ue}
	\begin{tikzpicture}
		\begin{axis}[EQBSDS500ItUE]
			\addplot[CCS] coordinates{
				(1,0.204253)
				(5,0.162655)
				(10,0.143563)
				(25,0.133202)
			};
			\addplot[CIS] coordinates{
				(1,0.123373)
				(3,0.132016)
				(6,0.132153)
			};
			\addplot[CRS] coordinates{
				(1,0.111657)
				(3,0.109511)
				(6,0.109207)
			};
			\addplot[ETPS] coordinates{
				(1,0.120458)
				(5,0.120686)
				(10,0.120595)
				(25,0.120575)
			};
			\addplot[LSC] coordinates{
				(1,0.110857)
				(5,0.107074)
				(10,0.106936)
				(25,0.107473)
			};
			\addplot[SLIC] coordinates{
				(1,0.14737)
				(5,0.136587)
				(10,0.134108)
				(25,0.133617)
			};
			\addplot[PRESLIC] coordinates{
				(1,0.156342)
				(5,0.15315)
				(10,0.15315)
				(25,0.15315)
			};
			\addplot[RESEEDS] coordinates{
				(1,0.135469)
				(5,0.132502)
				(10,0.138297)
				(25,0.138375)
			};
			\addplot[SEEDS] coordinates{
				(1,0.156297)
				(5,0.162523)
				(10,0.165167)
				(25,0.166744)
			};
			\addplot[VLSLIC] coordinates{
				(1,0.13491)
				(5,0.12431)
				(10,0.125229)
				(25,0.126088)
			};
		\end{axis}
	\end{tikzpicture}
\end{subfigure}
\begin{subfigure}[b]{\halfthreeone\textwidth}\phantomsubcaption\label{subfig:experiments-iterations-bsds500-t}
	\begin{tikzpicture}
		\begin{axis}[EQBSDS500Itt,ymode=log]
			\addplot[CCS] coordinates{
				(1,0.0665)
				(5,0.0985002)
				(10,0.14135)
				(25,0.24505)
			};
			\addplot[CIS] coordinates{
				(1,1.97275)
				(3,5.20055)
				(6,7.3538)
			};
			\addplot[CRS] coordinates{
				(1,0.3295)
				(3,0.8723)
				(6,1.68255)
			};
			\addplot[ETPS] coordinates{
				(1,0.0583001)
				(5,0.16035)
				(10,0.28495)
				(25,0.65425)
			};
			\addplot[LSC] coordinates{
				(1,0.1149)
				(5,0.1484)
				(10,0.19055)
				(25,0.31585)
			};
			\addplot[SLIC] coordinates{
				(1,0.0610501)
				(5,0.0748499)
				(10,0.0937001)
				(25,0.1465)
			};
			\addplot[PRESLIC] coordinates{
				(1,0.01265)
				(5,0.0294)
				(10,0.02955)
				(25,0.02925)
			};
			\addplot[RESEEDS] coordinates{
				(1,0.0414)
				(5,0.0555001)
				(10,0.0619001)
				(25,0.0773499)
			};
			\addplot[SEEDS] coordinates{
				(1,0.0870501)
				(5,0.22575)
				(10,0.41915)
				(25,0.92205)
			};
			\addplot[VLSLIC] coordinates{
				(1,0.03525)
				(5,0.11775)
				(10,0.22025)
				(25,0.5291)
			};
		\end{axis}
	\end{tikzpicture}
\end{subfigure}
    \caption{\Rec, \UE and runtime in seconds $t$ for iterative algorithms with 
    $\K \approx 400$ on the \BSDS dataset.
    Some algorithms allow to gradually trade performance for runtime, reducing runtime 
    by several $100ms$ in some cases.
    \textbf{Best viewed in color.}}
    \label{fig:experiments-iterations}
	\vskip 12px
	{\scriptsize
		\begin{tabularx}{0.475\textwidth}{X X X l}
			\ref{plot:w} \W &
			\ref{plot:eams} \EAMS &
			\ref{plot:nc} \NC &
			\ref{plot:fh} \FH \\
			\ref{plot:refh} \reFH &
			\ref{plot:rw} \RW &
			\ref{plot:qs} \QS &
			\ref{plot:pf} \PF \\
			\ref{plot:tp} \TP &
			\ref{plot:cis} \CIS &
			\ref{plot:slic} \SLIC &
			\ref{plot:vlslic} \vlSLIC \\
			\ref{plot:crs} \CRS &
			\ref{plot:ers} \ERS &
			\ref{plot:pb} \PB &
			\ref{plot:dasp} \DASP \\
			\ref{plot:seeds} \SEEDS &
			\ref{plot:reseeds} \reSEEDS &
			\ref{plot:tps} \TPS &
			\ref{plot:vc} \VC \\
			\ref{plot:ccs} \CCS &
			\ref{plot:vccs} \VCCS &
			\ref{plot:cw} \CW &
			\ref{plot:ergc} \ERGC \\
			\ref{plot:mss} \MSS &
			\ref{plot:preslic} \preSLIC &
			\ref{plot:wp} \WP &
			\ref{plot:etps} \ETPS \\
			\ref{plot:lsc} \LSC &
			\ref{plot:poise} \POISE &
			\ref{plot:seaw} \SEAW &
		\end{tabularx}
	}

\end{figure}
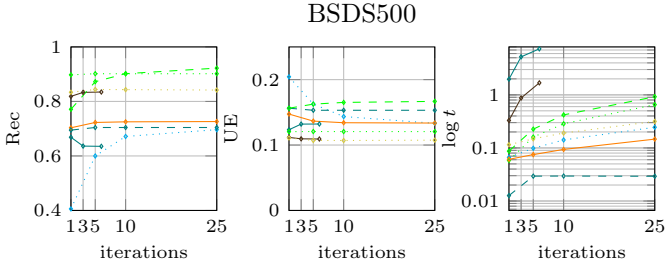
\FloatBarrier

Iterative algorithms offer to reduce runtime while gradually lowering performance.
Considering Figure \ref{fig:experiments-iterations},
showing \Rec, \UE and runtime in seconds $t$ for all iterative algorithms on the \BSDS dataset,
we observe that the number of iterations can safely be reduced to decrease
runtime while lowering \Rec and increasing \UE only slightly. In the best case,
for example considering \ETPSr, reducing the number of iterations from $25$ to $1$
reduces the runtime from $680\text{ms}$ to $58\text{ms}$, while keeping \Rec und \UE
nearly constant. For other cases, such as \SEEDSr\footnote{
	We note that we use the original \SEEDS implementation instead of the
	OpenCV \cite{Bradski:2000} implementation which is reported to be more efficient, see
	\url{http://docs.opencv.org/3.0-last-rst/modules/ximgproc/doc/superpixels.html}.
}, \Rec decreases abruptly when
using less than $5$ iterations. Still, runtime can be reduced from $920\text{ms}$
to $220\text{ms}$. For \CRSr and \CISr, runtime reduction is similarly significant,
but both algorithms still exhibit higher runtimes. If post-processing
is necessary, for example for \SLICr and \preSLICr, the number of iterations has
to be fixed in advance. However, for other iterative algorithms, the number of
iterations may be adapted at runtime depending on the available time.
Overall, iterative algorithms are beneficial as they are able to gradually
trade performance for runtime.

We conclude that watershed-based as well as some path-based and clustering-based
algorithms are candidates for realtime applications. Iterative algorithms,
in particular many clustering-based and iterative energy optimization algorithms, may
further be speeded up by reducing the number of iterations and trading performance for
runtime. On a final note, we want to remind the reader that the image sizes of all
used datasets may be relatively small compared to today's applications.
However, the relative runtime comparison is still valuable.

\subsection{Influence of Implementations}
\label{subsec:experiments-implementations}

We discuss the influence of implementation details on performance and runtime for
different implementations of \SEEDS, \SLIC and \FH. In particular, \reSEEDS and \reFH 
are our revised implementations of \SEEDS and \FH, respectively. 
Both implementations follow the algorithm as described in the corresponding publications,
provide exactly the same parameters and are also implemented in C/C++.
Still, we tried to optimize the implementations with respect to connectivity and
kept them as efficient as possible. 
\reFH additionally uses a slightly different graph data structure -- we refer to the
implementations for details; these will be made publicly available together with the benchmark.
Furthermore, we include \vlSLIC, an implementation of \SLIC as part of the VLFeat library
\cite{VedaldiFulkerson:2008}, and \preSLIC \cite{NeubertProtzel:2014}, an accelerated version of \SLIC based on the
original implementation. Both \vlSLIC and \preSLIC are also implemented in C/C++ as
their original counterparts.

We find that revisiting implementation details may be beneficial for both performance and runtime.
We consider Figure \ref{fig:experiments-implementations-bsds500} showing \Rec, \UE
and runtime in seconds $t$ for the introduced implementations of \SLIC, \SEEDS and \FH on the \BSDS dataset.
For \reSEEDSr and \reFHr, we observe improved performance which can be related to
the improved connectivity. However, even very similar implementations such as \SLICr
and \vlSLICr differ slightly in performance; note the lower \Rec and higher \UE
of \vlSLICr compared to \SLICr. Overall, the difference in runtime is most striking, for example \reSEEDSr
and \preSLICr show significantly lower runtime compared to \SEEDSr and \SLICr.
\reFHr, in contrast, shows higher runtime compared to \FHr due to a more complex data structure.

As expected, implementation details effect runtime, however, in the presented cases,
\ie for \SLIC, \SEEDS and \FH, performance is also affected. Nevertheless, it
still needs to be examined whether this holds true for the remaining algorithms, as well.
Furthermore, the experiments suggest that improving connectivity helps
performance.

\subsection{Robustness}
\label{subsec:experiments-robustness}

Similar to Neubert and Protzel \cite{NeubertProtzel:2012}, we investigate the influence
of noise, blur and affine transformations.
We evaluated all algorithms for $\K \approx 400$ on the \BSDS dataset.
In the following we exemplarily discuss salt and pepper noise and average blurring.

\begin{figure}[t]
	\centering
	\begin{subfigure}[b]{\halfthreeone\textwidth}\phantomsubcaption\label{subfig:experiments-implementations-bsds500-rec.mean_min}
	\begin{tikzpicture}
		\begin{axis}[EIBSDS500Rec,xmode=log]

			\addplot[SEEDS] coordinates{
				(261.62,0.882484)
				(365.675,0.906473)
				(468.81,0.922132)
				(670.57,0.941889)
				(870.75,0.952388)
				(1087.4,0.963175)
				(1270.11,0.967167)
				(1451.85,0.973778)
				(1669.18,0.974282)
				(1873.19,0.980114)
				(2104.62,0.984773)
				(2462.77,0.98341)
				(2793.43,0.989414)
				(3260.86,0.989969)
				(3895.78,0.993267)
				(3895.78,0.993267)
				(4846.12,0.995104)
				(4846.12,0.995104)
			};

			\addplot[RESEEDSI] coordinates{
				(200.795,0.876167)
				(301.525,0.902782)
				(401.465,0.918521)
				(602.33,0.937366)
				(800.99,0.949909)
				(1020.12,0.957079)
				(1201.34,0.965188)
				(1378.1,0.969268)
				(1601.55,0.97409)
				(1802.11,0.976241)
				(2040.11,0.980008)
				(2402.13,0.984131)
				(2720.13,0.987033)
				(3200.2,0.988506)
				(3840.22,0.992189)
				(3840.22,0.992189)
				(4800.37,0.994628)
				(4800.37,0.994628)
			};

			\addplot[FH] coordinates{
				(628.745,0.663508)
				(799.09,0.706555)
				(963.36,0.748153)
				(1090.39,0.75093)
				(1187.04,0.782457)
				(1605.71,0.808989)
				(2533.01,0.888881)
				(3000.74,0.889299)
				(3219.63,0.904142)
				(3814.42,0.913162)
				(4746.96,0.935408)
			};

			\addplot[REFHI] coordinates{
				(247.49,0.716919)
				(335.32,0.752939)
				(465.54,0.79338)
				(543.205,0.80981)
				(621.41,0.823826)
				(862.475,0.857431)
				(1059.11,0.876169)
				(1359.22,0.898679)
				(1871.67,0.923443)
				(2930.78,0.950073)
				(3532,0.962835)
				(4508.69,0.975612)
				(4998.21,0.978696)
			};

			\addplot[PRESLIC] coordinates{
				(369,0.704329)
				(581.54,0.753518)
				(734.575,0.781089)
				(1020.3,0.819649)
				(1229.22,0.839652)
				(1229.22,0.839652)
				(1511.3,0.862751)
				(1838.94,0.883831)
				(1838.94,0.883831)
				(2375.99,0.911038)
				(3059.43,0.934761)
				(3059.43,0.934761)
				(3059.43,0.934761)
				(4218.88,0.962025)
				(4218.88,0.962025)
				(6137.8,0.985124)
			};

			\addplot[VLSLICI] coordinates{
				(575.975,0.78762)
				(651.86,0.800183)
				(763.62,0.819088)
				(899,0.836123)
				(988.985,0.845203)
				(1193.67,0.860582)
				(1348.08,0.866749)
				(1349.01,0.865282)
				(1579.08,0.87797)
				(1849.65,0.891934)
				(1857.27,0.890967)
				(2307.73,0.909537)
				(2890.56,0.929862)
				(2924.52,0.929379)
				(2954.18,0.9279)
				(3858.98,0.952214)
				(3858.98,0.952214)
				(4809.57,0.969088)
			};

			\addplot[SLIC] coordinates{
				(180.32,0.649493)
				(256.725,0.687987)
				(368.57,0.726167)
				(575.335,0.778431)
				(726.36,0.807455)
				(1002.87,0.843839)
				(1203.61,0.868781)
				(1203.61,0.868781)
				(1475.69,0.887329)
				(1814.8,0.908264)
				(1814.8,0.908264)
				(2334.56,0.929101)
				(3038.13,0.949264)
				(3038.13,0.949264)
				(3038.13,0.949264)
				(4188.97,0.970933)
				(4188.97,0.970933)
				(6139.41,0.988423)
			};

		\end{axis}
	\end{tikzpicture}
\end{subfigure}
\begin{subfigure}[b]{\halfthreeone\textwidth}\phantomsubcaption\label{subfig:experiments-implementations-bsds500-ue_np.mean_max}
	\begin{tikzpicture}
		\begin{axis}[EIBSDS500UE,xmode=log]

			\addplot[SEEDS] coordinates{
				(261.62,0.223076)
				(365.675,0.19367)
				(468.81,0.166744)
				(670.57,0.144635)
				(870.75,0.127624)
				(1087.4,0.120216)
				(1270.11,0.110512)
				(1451.85,0.104601)
				(1669.18,0.0958062)
				(1873.19,0.091841)
				(2104.62,0.102696)
				(2462.77,0.0825942)
				(2793.43,0.0820788)
				(3260.86,0.0738429)
				(3895.78,0.0707603)
				(3895.78,0.0707603)
				(4846.12,0.0633063)
				(4846.12,0.0633063)
			};

			\addplot[SLIC] coordinates{
				(180.32,0.174507)
				(256.725,0.14981)
				(368.57,0.133617)
				(575.335,0.113918)
				(726.36,0.10487)
				(1002.87,0.0957898)
				(1203.61,0.0912283)
				(1203.61,0.0912283)
				(1475.69,0.0857781)
				(1814.8,0.0785351)
				(1814.8,0.0785351)
				(2334.56,0.0724657)
				(3038.13,0.0657068)
				(3038.13,0.0657068)
				(3038.13,0.0657068)
				(4188.97,0.0589157)
				(4188.97,0.0589157)
				(6139.41,0.0512524)
			};

			\addplot[FH] coordinates{
				(628.745,0.182601)
				(799.09,0.150389)
				(963.36,0.140866)
				(1090.39,0.131649)
				(1187.04,0.125432)
				(1605.71,0.110407)
				(2533.01,0.0877137)
				(3000.74,0.0845623)
				(3219.63,0.0821065)
				(3814.42,0.0819198)
				(4746.96,0.0709097)
			};

			\addplot[RESEEDSI] coordinates{
				(200.795,0.185835)
				(301.525,0.163531)
				(401.465,0.132874)
				(602.33,0.113933)
				(800.99,0.10611)
				(1020.12,0.0994214)
				(1201.34,0.0923218)
				(1378.1,0.087877)
				(1601.55,0.0800663)
				(1802.11,0.0781975)
				(2040.11,0.0832874)
				(2402.13,0.0699599)
				(2720.13,0.0694846)
				(3200.2,0.0644983)
				(3840.22,0.0617858)
				(3840.22,0.0617858)
				(4800.37,0.0561646)
				(4800.37,0.0561646)
			};

			\addplot[PRESLIC] coordinates{
				(369,0.15315)
				(581.54,0.127686)
				(734.575,0.117583)
				(1020.3,0.103212)
				(1229.22,0.0957013)
				(1229.22,0.0957013)
				(1511.3,0.0880409)
				(1838.94,0.0822743)
				(1838.94,0.0822743)
				(2375.99,0.0748201)
				(3059.43,0.068826)
				(3059.43,0.068826)
				(3059.43,0.068826)
				(4218.88,0.0614387)
				(4218.88,0.0614387)
				(6137.8,0.0537188)
			};

			\addplot[VLSLICI] coordinates{
				(575.975,0.146745)
				(651.86,0.133883)
				(763.62,0.12431)
				(899,0.112635)
				(988.985,0.106043)
				(1193.67,0.0981205)
				(1348.08,0.0935383)
				(1349.01,0.0929579)
				(1579.08,0.0889043)
				(1849.65,0.0848381)
				(1857.27,0.0842664)
				(2307.73,0.0796918)
				(2890.56,0.0748332)
				(2924.52,0.0736624)
				(2954.18,0.0727692)
				(3858.98,0.0707121)
				(3858.98,0.0707121)
				(4809.57,0.077104)
			};

			\addplot[REFHI] coordinates{
				(247.49,0.156969)
				(335.32,0.139838)
				(465.54,0.125235)
				(543.205,0.117281)
				(621.41,0.111133)
				(862.475,0.0986539)
				(1059.11,0.0920731)
				(1359.22,0.0849422)
				(1871.67,0.076739)
				(2930.78,0.0663083)
				(3532,0.0618438)
				(4508.69,0.0587328)
				(4998.21,0.0585372)
			};

		\end{axis}
	\end{tikzpicture}
\end{subfigure}
\begin{subfigure}[b]{\halfthreeone\textwidth}\phantomsubcaption\label{subfig:experiments-implementations-bsds500-t}
	\begin{tikzpicture}
		\begin{axis}[EIBSDS500t,ymode=log]

			\addplot[RESEEDSI] coordinates{
				(401.465,0.03832495)
				(1201.34,0.03862495)
				(3840.22,0.03812495)
			};

			\addplot[SEEDS] coordinates{
				(468.81,0.450675)
				(1270.11,0.45755)
				(3895.78,0.452425)
			};

			\addplot[FH] coordinates{
				(782.42,0.03667495)
				(799.09,0.034075)
				(3219.63,0.02397505)
			};

			\addplot[PRESLIC] coordinates{
				(369,0.01435)
				(1229.22,0.01415)
				(3059.43,0.01335)
			};

			\addplot[SLIC] coordinates{
				(368.57,0.072925)
				(1203.61,0.07685)
				(3038.13,0.078375)
			};

			\addplot[VLSLICI] coordinates{
				(763.62,0.0588)
				(1348.08,0.05875)
				(2954.18,0.05895)
			};

			\addplot[REFHI] coordinates{
				(465.54, 0.094)
				(862.475, 0.093)
				(2930.78, 0.086)
			};

		\end{axis}
	\end{tikzpicture}
\end{subfigure}
	\caption{\Rec, \UE and runtime in seconds $t$ on the \BSDS dataset for different implementations of \SLIC, \SEEDS and \FH.
	In particular, \reSEEDS and \reFH show slightly better performance which may be explained by
	improved connectivity. \vlSLIC, in contrast, shows similar performance to \SLIC and, indeed,
	the implementations are very similar. Finally, \preSLIC reduces runtime by reducing the number
	of iterations spend on individual superpixels.
    \textbf{Best viewed in color.}}
	\label{fig:experiments-implementations-bsds500}
    \vskip 12px
	{\scriptsize
		\begin{tabularx}{0.475\textwidth}{X X X l}
			\ref{plot:fh} \FH &
			\ref{plot:refh} \reFH &
			\ref{plot:slic} \SLIC &
			\ref{plot:vlslic} \vlSLIC\\
			\ref{plot:seeds} \SEEDS &
			\ref{plot:reseeds} \reSEEDS &
			\ref{plot:preslic} \preSLIC
		\end{tabularx}
	}

\end{figure}

\begin{figure}[t]
    \centering
    \input{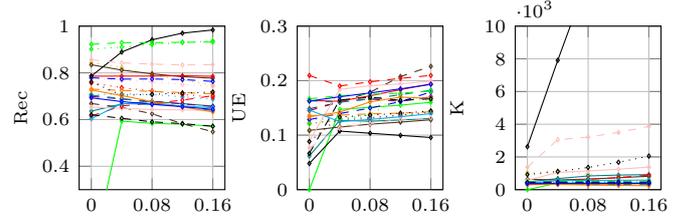}
    \caption{The influence of salt and pepper noise for $p \in \{0, 0.04, 0.08, 0.12, 0.16\}$ being the
    probability of salt or pepper. Regarding \Rec and \UE, most algorithms are not
    significantly influence by salt and pepper noise. Algorithms such as
    \QS and \VC compensate the noise by generating additional superpixels.
    \textbf{Best viewed in color.}}
    \label{subfig:experiments-robustness-noise}
\end{figure}

\begin{figure}[t]
	\centering
    \input{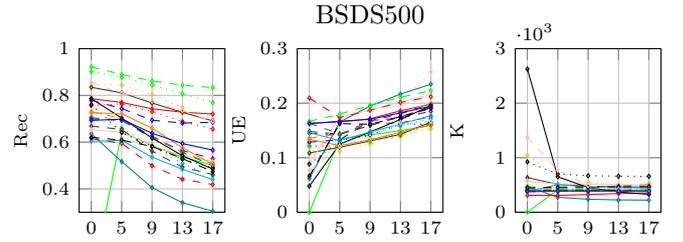}
    \caption{The influence of average blur for $k \in \{0, 5, 9, 13, 17\}$ being
    the filter size. As can be seen, blurring gradually reduces performance
    -- which may be explained by vanishing image boundaries. In addition, for
    algorithms such as \VC and \QS, blurring also leads to fewer superpixels being generated.
    \textbf{Best viewed in color.}}
    \label{subfig:experiments-robustness-blur}
    \vskip 12px
	{\scriptsize
		\begin{tabularx}{0.475\textwidth}{X X X l}
			\ref{plot:w} \W &
			\ref{plot:eams} \EAMS &
			\ref{plot:nc} \NC &
			\ref{plot:fh} \FH\\
			\ref{plot:rw} \RW &
			\ref{plot:qs} \QS &
			\ref{plot:pf} \PF &
			\ref{plot:tp} \TP\\
			\ref{plot:cis} \CIS &
			\ref{plot:slic} \SLIC &
			\ref{plot:crs} \CRS &
			\ref{plot:ers} \ERS\\
			\ref{plot:pb} \PB &
			\ref{plot:seeds} \SEEDS &
			\ref{plot:tps} \TPS &
			\ref{plot:vc} \VC\\
			\ref{plot:ccs} \CCS &
			\ref{plot:cw} \CW &
			\ref{plot:ergc} \ERGC &
			\ref{plot:mss} \MSS\\
			\ref{plot:preslic} \preSLIC &
			\ref{plot:wp} \WP &
			\ref{plot:etps} \ETPS &
			\ref{plot:lsc} \LSC\\
			\ref{plot:poise} \POISE &
			\ref{plot:seaw} \SEAW &
		\end{tabularx}
	}

\end{figure}

Most algorithms are robust to salt and pepper noise; blurring, in contrast, tends
to reduce performance. We consider Figure \ref{subfig:experiments-robustness-noise} showing \Rec, \UE and \K
for $p \in \{0, 0.04,$ $0.08, 0.12, 0.16\}$ being the probability of a pixel being salt or pepper.
Note that Figure \ref{subfig:experiments-robustness-noise} shows the number of
superpixels $K$ before enforcing connectivity as described in Section \ref{subsec:parameter-optimization-connectivity}.
As we can deduce, salt and pepper noise only slightly reduces \Rec and \UE for most algorithms.
Some algorithms compensate the noise by generating more superpixels such as \VCr or \SEAWr
while only slightly reducing performance. In contrast, for \QSr the performance even increases
-- a result of the strongly increasing number of superpixels.
Similar results can be obtained for Gaussian additive noise.
Turning to Figure \ref{subfig:experiments-robustness-blur}
showing \Rec, \UE and \K for $k \in \{0, 5, 9, 13, 17\}$ being the size of a box
filter used for average blurring. As expected, blurring leads to reduced performance with
respect to both \Rec and \UE. Furthermore, it leads to a reduced number of generated
superpixels for algorithms such as \QSr or \VCr. Similar observations can be made for motion blur as well as Gaussian blur.


Overall, most superpixel algorithms are robust to the considered noise models,
while blurring tends to reduce performance. Although the corresponding experiments are omitted for brevity,
we found that affine transformations do not influence performance.
\subsection{More Superpixels}
\label{subsec:experiments-k}

\begin{table}[t]
	\centering
	{\scriptsize
	\begin{tabular}{| r | l | l | l | l | l |}
		\hline
		& \Rec & \UE & \EV & \K & $t$\\\hline
		\W & 0.9998 & 0.0377 & 0.9238 & 20078.2 & 0.0090\\
		\PF & 0.9512 & 0.0763 & 0.8834 & 15171.5 & 0.0105\\
		\CIS & 0.9959 & 0.0353 & 0.9792 & 15988.6 & 7.617\\
		\SLIC & 0.9997 & 0.0339 & 0.9570 & 17029.7 & 0.1633\\
		\CRS & 0.9999 & 0.0265 & 0.9587 & 24615.8 & 4.7014\\
		\ERS & 0.9999 & 0.0297 & 0.9564 & 20000 & 0.5935\\
		\PB & 0.9988 & 0.0398 & 0.9477 & 15058.9 & 0.0531\\
		\SEEDS & 0.9997 & 0.0340 & 0.9729 & 18982.5 & 0.2267\\
		\VC & 0.9482 & 0.04621 & 0.9631 & 10487.5 & 2.3174\\
		\CCS & 0.9999 & 0.0223 & 0.9637 & 17676.9 & 0.227\\
		\CW & 0.9999 & 0.0322 & 0.9362 & 26319.8 & 0.0049\\
		\ERGC & 0.9999 & 0.0316 & 0.9744 & 21312 & 0.1217\\
		\MSS & 0.9989 & 0.0372 & 0.9171 & 25890.5 & 0.0894\\
		\preSLIC & 0.9999 & 0.0359 & 0.958 & 17088.3 & 0.021\\
		\WP & 0.9980 & 0.0411 & 0.9463 & 15502.7 & 0.6510\\
		\ETPS & 0.9999 & 0.0311 & 0.9793 & 17227 & 1.1657\\\hline
	\end{tabular}
	}
	\caption{\Rec, \UE, \EV, \K and runtime in seconds $t$ for $\K \approx 20000$
	on the \BSDS dataset including all algorithms able to generate $\K \gg 5000$ superpixels.
	The experiments demonstrate that nearly all superpixel algorithms are able to
	capture the image content without loss of information -- with $\Rec \approx 0.99$ and $\UE \approx 0.03$
	-- while reducing the number of primitives from $481 \cdot 321 = 154401$ to $\K \approx 20000$.}
	\label{table:experiments-k}
\end{table}

Up to now, we used \ARec, \AUE and \AEV to summarize experimental results for $\K \in [200,5200]$.
However, for some applications, generating $\K \gg 5200$ superpixels may be interesting.

For $\K \approx 20000$, superpixel algorithms can be used to dramatically reduce the
number of primitives for subsequent processing steps with negligible loss of information.
We consider Table \ref{table:experiments-k}
presenting \Rec, \UE, \EV, runtime in seconds $t$ and \K for $\K \approx 20,000$ on
the \BSDS dataset. We note that some algorithms were not able to generate $\K \gg 5200$
superpixels and are, therefore, excluded. Similarly, we excluded algorithms not offering
control over the number of generated superpixels. We observe that except for \VC and \PF all
algorithms achive $\Rec \geq 0.99$, $\UE \approx 0.03$, and $\EV > 0.9$. Furthermore, the runtime of
many algorithms is preserved. For example \W and \CW still run in below $10ms$ and the
runtime for \preSLIC and \SLIC increases only slightly. Obviously, the number of
generated superpixels varies more strongly for large~\K. Overall, most algorithms
are able to capture the image content nearly perfectly while reducing
 $321 \times 481 = 154401$ pixels to $\K \approx 20000$ superpixels.

\subsection{Ranking}
\label{subsec:experiments-ranking}

We conclude the experimental part of this paper with a ranking with respect to \ARec and \AUE
 -- reflecting the objective used for parameter optimization. Unfortunately, the high number
of algorithms as well as the low number of datasets prohibits using statistical
tests to extract rankings, as done in other benchmarks (\eg \cite{Demsar:2006,DollarWojekSchielePerona:2009}).
Therefore, Table \ref{table:ranking} presents average \ARec and \AUE, average ranks
as well as the corresponding rank matrix.
On each dataset, the algorithms were ranked according to $\ARec + \AUE$ where
lowest $\ARec + \AUE$ corresponds to the best rank, \ie rank one.
The corresponding rank matrix represents the rank distribution
(\ie the frequencies of the attained ranks) for each algorithm.
We find that the presented average ranks provide a founded overview of the evaluated 
algorithms, summarizing many of the observations discussed before.
In the absence of additional constraints, Table \ref{table:ranking}
may be used to select suitable superpixel algorithms.
\begin{table*}[ht]
	\centering
	{\scriptsize
	\begin{tabular}{| c | c || c || c | cccccccccccccccccccccccccccc |}
		\hline
		\ARec & \AUE && Rank & 1\hphantom{1} & 2\hphantom{1} & 3\hphantom{1} & 4\hphantom{1} & 5\hphantom{1} & 6\hphantom{1} & 7\hphantom{1} & 8\hphantom{1} & 9\hphantom{1} & 10 & 11 & 12 & 13 & 14 & 15 & 16 & 17 & 18 & 19 & 20 & 21 & 22 & 23 & 24 & 25 & 26 & 27 & 28\\
		\hline
		2.05 & 6.07 & \ETPS & 1 & \cellcolor{black!65}5 & 0 & 0 & 0 & 0 & 0 & 0 & 0 & 0 & 0 & 0 & 0 & 0 & 0 & 0 & 0 & 0 & 0 & 0 & 0 & 0 & 0 & 0 & 0 & 0 & 0 & 0 & 0\\
		1.57 & 7.98 & \SEEDS & 3.8 & 0 & \cellcolor{black!35}2 & \cellcolor{black!25}1 & \cellcolor{black!25}1 & 0 & 0 & 0 & \cellcolor{black!25}1 & 0 & 0 & 0 & 0 & 0 & 0 & 0 & 0 & 0 & 0 & 0 & 0 & 0 & 0 & 0 & 0 & 0 & 0 & 0 & 0\\
		3.38 & 6.28 & \ERS & 3.8 & 0 & \cellcolor{black!25}1 & \cellcolor{black!25}1 & \cellcolor{black!35}2 & 0 & \cellcolor{black!25}1 & 0 & 0 & 0 & 0 & 0 & 0 & 0 & 0 & 0 & 0 & 0 & 0 & 0 & 0 & 0 & 0 & 0 & 0 & 0 & 0 & 0 & 0\\
		4.16 & 6.29 & \CRS & 4.8 & 0 & 0 & \cellcolor{black!35}2 & \cellcolor{black!25}1 & \cellcolor{black!25}1 & 0 & 0 & 0 & \cellcolor{black!25}1 & 0 & 0 & 0 & 0 & 0 & 0 & 0 & 0 & 0 & 0 & 0 & 0 & 0 & 0 & 0 & 0 & 0 & 0 & 0\\
		4.18 & 6.77 & \EAMS & 5.4 & 0 & \cellcolor{black!25}1 & \cellcolor{black!25}1 & 0 & 0 & \cellcolor{black!25}1 & \cellcolor{black!25}1 & 0 & \cellcolor{black!25}1 & 0 & 0 & 0 & 0 & 0 & 0 & 0 & 0 & 0 & 0 & 0 & 0 & 0 & 0 & 0 & 0 & 0 & 0 & 0\\
		4.55 & 6.34 & \ERGC & 5.8 & 0 & 0 & 0 & 0 & \cellcolor{black!35}2 & \cellcolor{black!35}2 & \cellcolor{black!25}1 & 0 & 0 & 0 & 0 & 0 & 0 & 0 & 0 & 0 & 0 & 0 & 0 & 0 & 0 & 0 & 0 & 0 & 0 & 0 & 0 & 0\\
		4.91 & 6.50 & \SLIC & 6.2 & 0 & 0 & 0 & 0 & \cellcolor{black!35}2 & 0 & \cellcolor{black!45}3 & 0 & 0 & 0 & 0 & 0 & 0 & 0 & 0 & 0 & 0 & 0 & 0 & 0 & 0 & 0 & 0 & 0 & 0 & 0 & 0 & 0\\
		5.40 & 7.24 & \LSC & 9.2 & 0 & 0 & 0 & \cellcolor{black!25}1 & 0 & \cellcolor{black!25}1 & 0 & 0 & \cellcolor{black!35}2 & 0 & 0 & 0 & 0 & 0 & 0 & 0 & 0 & \cellcolor{black!25}1 & 0 & 0 & 0 & 0 & 0 & 0 & 0 & 0 & 0 & 0\\
		5.49 & 7.02 & \preSLIC & 9.2 & 0 & 0 & 0 & 0 & 0 & 0 & 0 & \cellcolor{black!35}2 & \cellcolor{black!25}1 & \cellcolor{black!25}1 & \cellcolor{black!25}1 & 0 & 0 & 0 & 0 & 0 & 0 & 0 & 0 & 0 & 0 & 0 & 0 & 0 & 0 & 0 & 0 & 0\\
		6.60 & 6.41 & \CCS & 10.6 & 0 & 0 & 0 & 0 & 0 & 0 & 0 & \cellcolor{black!35}2 & 0 & \cellcolor{black!25}1 & 0 & \cellcolor{black!25}1 & 0 & 0 & \cellcolor{black!25}1 & 0 & 0 & 0 & 0 & 0 & 0 & 0 & 0 & 0 & 0 & 0 & 0 & 0\\
		6.49 & 7.19 & \CW & 11 & 0 & 0 & 0 & 0 & 0 & 0 & 0 & 0 & 0 & \cellcolor{black!35}2 & \cellcolor{black!35}2 & 0 & \cellcolor{black!25}1 & 0 & 0 & 0 & 0 & 0 & 0 & 0 & 0 & 0 & 0 & 0 & 0 & 0 & 0 & 0\\
		8.24 & 7.83 & \DASP & 12 & 0 & 0 & 0 & 0 & 0 & 0 & 0 & 0 & 0 & \cellcolor{black!25}1 & 0 & 0 & 0 & \cellcolor{black!25}1 & 0 & 0 & 0 & 0 & 0 & 0 & 0 & 0 & 0 & 0 & 0 & 0 & 0 & 0\\
		6.51 & 7.44 & \W & 12.8 & 0 & 0 & 0 & 0 & 0 & 0 & 0 & 0 & 0 & 0 & \cellcolor{black!35}2 & \cellcolor{black!25}1 & 0 & \cellcolor{black!25}1 & 0 & \cellcolor{black!25}1 & 0 & 0 & 0 & 0 & 0 & 0 & 0 & 0 & 0 & 0 & 0 & 0\\
		7.44 & 6.97 & \WP & 13.4 & 0 & 0 & 0 & 0 & 0 & 0 & 0 & 0 & 0 & 0 & 0 & \cellcolor{black!25}1 & \cellcolor{black!35}2 & \cellcolor{black!25}1 & \cellcolor{black!25}1 & 0 & 0 & 0 & 0 & 0 & 0 & 0 & 0 & 0 & 0 & 0 & 0 & 0\\
		7.85 & 7.58 & \POISE & 13.8 & 0 & \cellcolor{black!25}1 & 0 & 0 & 0 & 0 & 0 & 0 & 0 & 0 & 0 & 0 & \cellcolor{black!25}1 & 0 & \cellcolor{black!25}1 & \cellcolor{black!25}1 & 0 & 0 & 0 & 0 & 0 & 0 & \cellcolor{black!25}1 & 0 & 0 & 0 & 0 & 0\\
		8.13 & 6.90 & \NC & 15.25 & 0 & 0 & 0 & 0 & 0 & 0 & 0 & 0 & 0 & 0 & 0 & \cellcolor{black!35}2 & 0 & 0 & 0 & 0 & 0 & \cellcolor{black!25}1 & \cellcolor{black!25}1 & 0 & 0 & 0 & 0 & 0 & 0 & 0 & 0 & 0\\
		7.66 & 7.43 & \MSS & 15.8 & 0 & 0 & 0 & 0 & 0 & 0 & 0 & 0 & 0 & 0 & 0 & 0 & \cellcolor{black!25}1 & \cellcolor{black!25}1 & 0 & 0 & \cellcolor{black!35}2 & \cellcolor{black!25}1 & 0 & 0 & 0 & 0 & 0 & 0 & 0 & 0 & 0 & 0\\
		8.90 & 7.67 & \VC & 17.8 & 0 & 0 & 0 & 0 & 0 & 0 & 0 & 0 & 0 & 0 & 0 & 0 & 0 & 0 & \cellcolor{black!25}1 & \cellcolor{black!35}2 & 0 & 0 & \cellcolor{black!25}1 & 0 & 0 & 0 & \cellcolor{black!25}1 & 0 & 0 & 0 & 0 & 0\\
		8.31 & 7.85 & \PB & 18.4 & 0 & 0 & 0 & 0 & 0 & 0 & 0 & 0 & 0 & 0 & 0 & 0 & 0 & \cellcolor{black!25}1 & \cellcolor{black!25}1 & 0 & 0 & 0 & 0 & \cellcolor{black!25}1 & \cellcolor{black!25}1 & \cellcolor{black!25}1 & 0 & 0 & 0 & 0 & 0 & 0\\
		8.53 & 8.15 & \FH & 19 & 0 & 0 & 0 & 0 & 0 & 0 & 0 & 0 & 0 & 0 & 0 & 0 & 0 & 0 & 0 & 0 & \cellcolor{black!25}1 & 0 & \cellcolor{black!45}3 & 0 & \cellcolor{black!25}1 & 0 & 0 & 0 & 0 & 0 & 0 & 0\\
		8.35 & 7.32 & \RW & 19 & 0 & 0 & 0 & 0 & 0 & 0 & 0 & 0 & 0 & 0 & 0 & 0 & 0 & 0 & 0 & \cellcolor{black!25}1 & \cellcolor{black!25}1 & 0 & 0 & \cellcolor{black!25}1 & 0 & 0 & \cellcolor{black!25}1 & 0 & 0 & 0 & 0 & 0\\
		11.45 & 6.70 & \CIS & 20.8 & 0 & 0 & 0 & 0 & 0 & 0 & 0 & 0 & 0 & 0 & 0 & 0 & 0 & 0 & 0 & 0 & \cellcolor{black!25}1 & 0 & 0 & \cellcolor{black!25}1 & \cellcolor{black!35}2 & 0 & 0 & 0 & \cellcolor{black!25}1 & 0 & 0 & 0\\
		11.45 & 7.19 & \TPS & 21 & 0 & 0 & 0 & 0 & 0 & 0 & 0 & 0 & 0 & 0 & 0 & 0 & 0 & 0 & 0 & 0 & 0 & \cellcolor{black!25}1 & 0 & \cellcolor{black!25}1 & \cellcolor{black!25}1 & \cellcolor{black!25}1 & 0 & \cellcolor{black!25}1 & 0 & 0 & 0 & 0\\
		11.05 & 7.93 & \TP & 21.6 & 0 & 0 & 0 & 0 & 0 & 0 & 0 & 0 & 0 & 0 & 0 & 0 & 0 & 0 & 0 & 0 & 0 & \cellcolor{black!25}1 & 0 & \cellcolor{black!25}1 & 0 & \cellcolor{black!25}1 & \cellcolor{black!25}1 & 0 & \cellcolor{black!25}1 & 0 & 0 & 0\\
		14.22 & 13.54 & \VCCS & 23 & 0 & 0 & 0 & 0 & 0 & 0 & 0 & 0 & 0 & 0 & 0 & 0 & 0 & 0 & 0 & 0 & 0 & 0 & 0 & 0 & 0 & \cellcolor{black!25}1 & 0 & \cellcolor{black!25}1 & 0 & 0 & 0 & 0\\
		11.97 & 8.36 & \SEAW & 24 & 0 & 0 & 0 & 0 & 0 & 0 & 0 & 0 & 0 & 0 & 0 & 0 & 0 & 0 & 0 & 0 & 0 & 0 & 0 & 0 & 0 & \cellcolor{black!25}1 & 0 & \cellcolor{black!35}2 & 0 & \cellcolor{black!25}1 & 0 & 0\\
		15.72 & 10.75 & \QS & 24.8 & 0 & 0 & 0 & 0 & 0 & 0 & 0 & 0 & 0 & 0 & 0 & 0 & 0 & 0 & 0 & 0 & 0 & 0 & 0 & 0 & 0 & 0 & \cellcolor{black!25}1 & \cellcolor{black!25}1 & \cellcolor{black!35}2 & 0 & \cellcolor{black!25}1 & 0\\
		24.64 & 14.91 & \PF & 26.2 & 0 & 0 & 0 & 0 & 0 & 0 & 0 & 0 & 0 & 0 & 0 & 0 & 0 & 0 & 0 & 0 & 0 & 0 & 0 & 0 & 0 & 0 & 0 & 0 & \cellcolor{black!25}1 & \cellcolor{black!45}3 & 0 & \cellcolor{black!25}1\\
		\hline
	\end{tabular}
	}
	\caption{Average \ARec and \AUE, average ranks and rank distribution for each evaluated algorithm.
	To compute average ranks, the algorithms were ranked according to $\ARec + \AUE$ 
	(where lowest $\ARec + \AUE$ corresponds to the best rank, \ie $1$)
	on each dataset separately.
	For all algorithms (rows), the rank distribution (columns $1$ through $28$) illustrates the frequency 
	a particular rank was attained over all considered datasets.
	We note that we could not evaluate \RW, \NC and \SEAW on the \SUNRGBD dataset and
	\DASP and \VCCS cannot be evaluated on the \BSDS, \SBD and \Fash datasets.}
	\label{table:ranking}
\end{table*}

\section{Conclusion}
\label{sec:conclusion}

In this paper, we presented a large-scale comparison of superpixel algorithms
taking into account visual quality, ground truth dependent and independent metrics,
runtime, implementation details as well as robustness to noise, blur and affine transformations.
For fairness, we systematically optimized parameters while strictly
enforcing connectivity. Based on the obtained parameters, we presented experiments
based on five different datasets including indoor and outdoor scenes as well as persons.
In contrast to existing work \cite{SchickFischerStiefelhagen:2012,AchantaShajiSmithLucchiFuaSuesstrunk:2012,NeubertProtzel:2012,NeubertProtzel:2013},
we considered minimum/maximum as well as the standard deviation in addition to simple metric averages.
We further proposed \AvgRec (\ARec), \AvgUE (\AUE) and \AvgEV (\AEV) to summarize algorithm performance
independent of the number of generated superpixels.
This enabled us to present an overall ranking of superpixel algorithms aimed to simplify and guide algorithm selection.

Regarding the mentioned aspects of superpixel algorithms, we made several observations
relevant for applications and future research. Considering visual quality, we found that
the majority of algorithms provides good boundary adherence; some algorithms are
able to capture even small details. However, better boundary adherence may influence compactness,
regularity and smoothness. While regularity and smoothness strongly depends on the individual algorithms,
a compactness parameter is beneficial to trade-off boundary adherence for compactness.
Regarding performance, Boundary Recall (\Rec) \cite{MartinFowlkesMalik:2004},
Undersegmentation Error (\UE) \cite{LevinshteinStereKutulakosFleetDickinsonSiddiqi:2009, AchantaShajiSmithLucchiFuaSuesstrunk:2012, NeubertProtzel:2012}
and Explained Variation (\EV) \cite{MoorePrinceWarrellMohammedJones:2008} provide a
good overview but are not sufficient to discriminate algorithms reliably.
Therefore, we used the minimum/maximum as well as the standard deviation of these metrics
to identify stable algorithms, \ie algoritms providing monotonically increasing
performance with regard to the number of generated superpixels. Furthermore, we were
able to relate poor performance to a high standard deviation in the number of generated superpixels,
justifying the need to strictly control connectivity. Concerning runtime, we identified
several algorithms providing realtime capabilities, \ie roughly $30\text{fps}$, and showed
that iterative algorithms allow to reduce runtime while only gradually reducing performance.
Implementation details are rarely discussed in the literature; on three examples, we highlighted
the advantage of ensuring connectivity and showed that revisiting implementations may benefit performance and runtime.
We further demonstrated that generating a higher number of superpixels, \eg roughly $20000$,
results in nearly no loss of information while reducing the high number of pixels to only $\sim 20000$ superpixels.
Finally, we experimentally argued that superpixel algorithms are robust against noise and
affine transformations before providing a final ranking of the algorithms based on the
proposed metrics \AvgRec and \AvgUE.

From the ranking in Table \ref{table:ranking}, we recommend 6 algorithms for use in practice,
thereby covering a wide range of application scenarios: \ETPS \cite{YaoBobenFidlerUrtasun:2015},
\SEEDS \cite{VanDenBerghBoixRoigCapitaniVanGool:2012}, \ERS \cite{LiuTuzelRamalingamChellappa:2011},
\CRS \cite{MesterConradGuevara:2011,ConradMertzMester:2013}, \ERGC \cite{BuyssensGardinRuan:2014}
and \SLIC \cite{AchantaShajiSmithLucchiFuaSuesstrunk:2010}.
These algorithms show superior performance regarding Boundary Recall, Undersegmentation
Error and Explained Variation and can be considered stable.
Furthermore, they are iterative (except for \ERGC and \ERS) and provide a compactness parameter (except for \SEEDS and \ERS).
Except for \ERS and \CRS, they provide runtimes below $100\text{ms}$ -- depending on the implementation --
and \preSLIC \cite{NeubertProtzel:2014}, which we see as a variant of \SLIC, provides realtime capabilities. Finally,
the algorithms provide control over the number of generated superpixels
(therefore, \EAMS, ranked $5$th in Table \ref{table:ranking}, is not recommended), are able to generate
mostly connected superpixels and exhibit a very low standard deviation in the number of generated superpixels.


\textbf{Software.} The individual implementations, together with the used benchmark,
are made publicly available at \url{davidstutz.de/projects/superpixel-benchmark/}.

\textbf{Acknowledgements.} The work in this paper was funded by the EU project STRANDS (ICT-2011-600623).
We are also grateful for the implementations provided by many authors.

\section*{References}
\bibliography{references/abbreviations,references/references}

\FloatBarrier
\clearpage
\newpage
\pagebreak
\FloatBarrier

\begin{appendix}
	\section{Algorithms}
\label{sec:appendix-algorithms}

Complementing the information presented in Section \ref{sec:algorithms}, Table \ref{table:algorithms}
gives a complete overview of all algorithms.

\begin{table*}[t]
    \centering
    {\scriptsize
        \begin{tabular}{r | c c c | c c | c c c c c c | c c c c | c c | c}
            & \rot{Reference} & \rot{Year} & \rot{Citations} & \rot{Categorization} & \rot{Implementation} & \rot{GRAY} & \rot{RGB} & \rot{Lab} & \rot{Luv} & \rot{HSV} & \rot{YCrCb} & \rot{\#Parameters} & \rot{\#Superpixels} & \rot{\#Iterations} & \rot{Compactness} & \rot{Depth} & \rot{Edges} & \\\hline
           \W & \cite{Meyer:1992} & 1992 & 234 & watershed & C/C++ & -- & \bcheckmark & \checkmark & \checkmark & \checkmark & \checkmark & 1 & \checkmark & -- & -- & -- & -- & \ref{plot:w}\\
           \EAMS & \cite{ComaniciuMeer:2002} & 2002 & 9631 & density & MatLab/C & -- & \bcheckmark & -- & \checkmark & -- & -- & 2 & -- & -- & -- & -- & \checkmark & \ref{plot:eams}\\
           \NC & \cite{RenMalik:2003} & 2003 & 996 & graph & MatLab/C & -- & \bcheckmark & -- & -- & -- & -- & 3 & \checkmark & -- & -- & -- & -- & \ref{plot:nc}\\
           \FH & \cite{FelzenswalbHuttenlocher:2004} & 2004 & 4144 & graph & C/C++ & -- & \bcheckmark & -- & -- & -- & -- & 3 & -- & -- & -- & -- & -- & \ref{plot:fh}\\
           \reFH & \multicolumn{3}{c|}{\dittostraight} & \multicolumn{1}{c}{\dittostraight} & C/C++ & -- & \bcheckmark & -- & -- & -- & -- & 3 & -- & -- & -- & -- & -- & \ref{plot:refh}\\
           \RW & \cite{GradyFunkaLea:2004,Grady:2006} & 2004 & 189 + 1587 & graph & MatLab/C & -- & \bcheckmark & -- & -- & -- & -- & 2 & \checkmark & -- & -- & -- & -- & \ref{plot:rw}\\
           \QS & \cite{VedaldiSoatto:2008} & 2008 & 376 & density & MatLab/C & -- & \checkmark & \bcheckmark & -- & -- & -- & 3 & -- & -- & -- & -- & -- & \ref{plot:qs}\\
           \PF & \cite{DruckerMacCormick:2009} & 2009 & 18 & path & Java & -- & \bcheckmark & -- & -- & -- & -- & 2 & \checkmark & -- & -- & -- & -- & \ref{plot:pf}\\
           \TP & \cite{LevinshteinStereKutulakosFleetDickinsonSiddiqi:2009} & 2009 & 559 & contour evolution & MatLab/C & -- & \bcheckmark & -- & -- & -- & -- & 4 & \checkmark & -- & -- & -- & -- & \ref{plot:tp}\\
           \CIS & \cite{VekslerBoykovMehrani:2010} & 2010 & 223 & graph & C/C++ & \bcheckmark & \checkmark & -- & -- & -- & -- & 4 & \checkmark & \checkmark & -- & -- & -- & \ref{plot:cis}\\
           \SLIC & \cite{AchantaShajiSmithLucchiFuaSuesstrunk:2010,AchantaShajiSmithLucchiFuaSuesstrunk:2012} & 2010 & 438 + 1843 & clustering & C/C++ & -- & \checkmark & \bcheckmark & -- & -- & -- & 4 & \checkmark & \checkmark & \checkmark & -- & -- & \ref{plot:slic}\\
           \vlSLIC & \multicolumn{3}{c|}{\dittostraight} & \multicolumn{1}{c}{\dittostraight} & C/C++ & -- & \bcheckmark & -- & -- & -- & -- & 4 & \checkmark & \checkmark & \checkmark & -- & -- & \ref{plot:vlslic}\\
           \CRS & \cite{MesterConradGuevara:2011,ConradMertzMester:2013} & 2011 & 14 + 4 & energy optimization & C/C++ & \checkmark & \checkmark & -- & -- & -- & \bcheckmark & 4 & \checkmark & \checkmark & \checkmark & -- & -- & \ref{plot:crs}\\
           \ERS & \cite{LiuTuzelRamalingamChellappa:2011} & 2011 & 216 & graph & C/C++ & -- & \bcheckmark & -- & -- & -- & -- & 3 & \checkmark & -- & -- & -- & -- & \ref{plot:ers}\\
           \PB & \cite{ZhangHartleyMashfordBurn:2011} & 2011 & 36 & graph & C/C++ & -- & \bcheckmark & -- & -- & -- & -- & 3 & \checkmark & -- & -- & -- & -- & \ref{plot:pb}\\
           \DASP & \cite{WeikersdorferGossowBeetz:2012} & 2012 & 22 & clustering & C/C++ & -- & \bcheckmark & -- & -- & -- & -- & 5 & \checkmark & \checkmark & \checkmark & -- & \checkmark & \ref{plot:dasp}\\
           \SEEDS & \cite{VanDenBerghBoixRoigCapitaniVanGool:2012} & 2012 & 98 & energy optimization & C/C++ & -- & \checkmark & \bcheckmark & -- & \checkmark & -- & 6 & \checkmark & \checkmark & -- & -- & -- & \ref{plot:seeds}\\
           \reSEEDS & \multicolumn{3}{c|}{\dittostraight} & \multicolumn{1}{c}{\dittostraight} & C/C++ & -- & \bcheckmark & \checkmark &  -- & \checkmark & -- & 6 & \checkmark & \checkmark & \checkmark & -- & -- & \ref{plot:reseeds}\\
           \TPS & \cite{DaiTangHuazhaFuXiaochunCao:2012,HuazhuFuXiaochunCaoDaiTangYahongHanDongXu:2014} & 2012 & 8 + 1 & path & MatLab/C & -- & \bcheckmark & -- & -- & -- & -- & 4 & \checkmark & -- & -- & -- & \checkmark & \ref{plot:tps}\\
           \VC & \cite{WangWang:2012} & 2012 & 36 & clustering & C/C++ & -- & \checkmark & \bcheckmark & -- & -- & -- & 6 & \checkmark & -- & \checkmark & -- & -- & \ref{plot:vc}\\
           \CCS & \cite{TasliCiglaGeversAlatan:2013,TasliCiglaAlatan:2015} & 2013 & 6 + 4 & energy optimization & C/C++ & -- & \checkmark & \bcheckmark & -- & -- & -- & 3 & \checkmark & \checkmark & \checkmark & -- & -- & \ref{plot:ccs}\\
           \VCCS & \cite{PaponAbramovSchoelerWoergoetter:2013} & 2013 & 87 & clustering & C/C++ & -- & \bcheckmark & -- & -- & -- & -- & 4 & -- & -- & \checkmark & \checkmark & -- & \ref{plot:vccs}\\
           \CW & \cite{NeubertProtzel:2014} & 2014 & 11 & watershed & C/C++ & -- & \bcheckmark & -- & -- & -- & -- & 2 & \checkmark & -- & \checkmark & -- & -- & \ref{plot:cw}\\
           \ERGC & \cite{BuyssensGardinRuan:2014,BuyssensToutainElmoatazLezoray:2014} & 2014 & 2 + 1 & contour evolution & C/C++ & -- & \checkmark & \bcheckmark & -- & -- & -- & 3 & \checkmark & -- & \checkmark & -- & -- & \ref{plot:ergc}\\
           \MSS & \cite{BenesovaKottman:2014} & 2014 & 4 & watershed & C/C++ & -- & \bcheckmark & -- & -- & -- & -- & 5 & \checkmark & -- & -- & -- & -- & \ref{plot:mss}\\
           \preSLIC & \cite{NeubertProtzel:2014} & 2014 & 11 & clustering & C/C++ & -- & \checkmark & \bcheckmark & -- & -- & -- & 4 & \checkmark & \checkmark & \checkmark & -- & -- & \ref{plot:preslic}\\
           \WP & \cite{MachairasDecenciereWalter:2014,MachairesFaesselCardenasPenaChabardesWalterDecenciere:2015} & 2014 & 5 + 8 & watershed & Python & -- & \bcheckmark & -- & -- & -- & -- & 2 & \checkmark & -- & \checkmark & -- & -- & \ref{plot:wp}\\
           \ETPS & \cite{YaoBobenFidlerUrtasun:2015} & 2015 & 6 & energy optimization & C/C++ & -- & \bcheckmark & -- & -- & -- & -- & 5 & \checkmark & \checkmark & \checkmark & -- & -- & \ref{plot:etps}\\
           \LSC & \cite{LiChen:2015} & 2015 & 2 & clustering & C/C++ & -- & \checkmark & \bcheckmark & -- & -- & -- & 4 & \checkmark & \checkmark & \checkmark & -- & -- & \ref{plot:lsc}\\
           \POISE & \cite{HumayunLiRehg:2015} & 2015 & 3 & graph & MatLab/C & -- & \bcheckmark & -- & -- & -- & -- & 5 & \checkmark & -- & -- & -- & \checkmark & \ref{plot:poise}\\
           \SEAW & \cite{StrassburgGrzeszickRothackerFink:2015} & 2015 & 0 & wavelet & MatLab/C & -- & \bcheckmark & -- & -- & -- & -- & 3 & -- & -- & -- & -- & -- & \ref{plot:seaw}
        \end{tabular}
    }
    \caption{List of evaluated superpixel algorithms. First of all we present the
    used acronym (in parts consistent with \cite{AchantaShajiSmithLucchiFuaSuesstrunk:2012}
    and \cite{NeubertProtzel:2012}), the corresponding publication, the year of
    publication and the number of Google Scholar citations as of October 13, 2016.
    We present a coarse categorization which is discussed in Section \ref{sec:algorithms}.
    We additionally present implementation details such as the programming language,
    supported color spaces and provided parameters. The color space used for evaluation is marked by a square.
    \textbf{Best viewed in color.}}
    \label{table:algorithms}
\end{table*}
	\section{Datasets}
\label{sec:appendix-datasets}

\begin{figure}[t]
	\centering
	\begin{subfigure}[b]{0.29\textwidth}
		\begin{subfigure}[t]{0.5\textwidth}%
			\includegraphics[height=1.75cm]{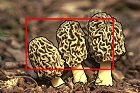}\phantomsubcaption\label{subfig:datasets-bsds500}%
		\end{subfigure}
		\begin{subfigure}[t]{0.425\textwidth}%
			\includegraphics[height=1.75cm]{pictures/nyuv2/gt/contours/marked/00000561}\phantomsubcaption\label{subfig:datasets-nyuv2}%
		\end{subfigure}
		\\[4px]
		\begin{subfigure}[t]{0.5\textwidth}%
			\includegraphics[height=1.75cm]{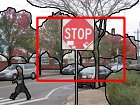}\phantomsubcaption\label{subfig:datasets-sbd}%
		\end{subfigure}
		\begin{subfigure}[t]{0.425\textwidth}%
			\includegraphics[height=1.75cm]{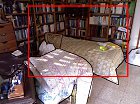}\phantomsubcaption\label{subfig:datasets-sunrgbd}%
		\end{subfigure}
	\end{subfigure}
	\hskip -3px
	\begin{subfigure}[t]{0.12\textwidth}%
		\includegraphics[height=3.68cm]{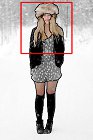}\phantomsubcaption\label{subfig:datasets-fash}%
	\end{subfigure}
	\caption{Example images from the used datasets. From left ro right: \BSDS; \SBD; \NYU; \SUNRGBD; and \Fash.
	Black contours represent ground truth and red rectangles indicate excerpts
	used for qualitative comparison in Figures \ref{fig:appendix-experiments-qualitative-bsds500-sbd-fash}
	and \ref{fig:appendix-experiments-qualitative-nyuv2-sunrgbd}.
	\textbf{Best viewed in color.}}
	\label{fig:appendix-datasets}
\end{figure}

The \BSDS dataset is the only dataset providing several ground truth segmentations per image.
Therefore, we briefly discuss evaluation on the \BSDS dataset in detail. Furthermore, 
additional example images from all used datasets are shown in Figure \ref{fig:appendix-datasets} and 
are used for qualitative results in \ref{subsec:appendix-experiments-qualitative}.

Assuming at least two ground truth segmentations per image, Arbel{\'a}ez \etal \cite{ArbelaezMaireFowlkesMalik:2011}
consider two methodologies of computing average \Rec: computing the average \Rec over all ground truth
segmentations per image, and subsequently taking the worst average (\ie the lowest \Rec); or taking the
lowest \Rec over all ground truth segmentations per image and averaging these. We follow Arbel{\'a}ez \etal
and pick the latter approach. The same methodology is then applied to \UE, \EV, \ASA and \UEL.
In this sense, we never overestimate performance with respect to the provided ground truth
segmentations.

	\section{Benchmark}
\label{sec:appendix-benchmark}

In the following section we discuss the expressiveness of the used metrics
and details regarding the computation of \ARec, \AUE and \AEV.

\subsection{Expressiveness and Correlation}
\label{subsec:appendix-benchmark-expressiveness}

Complementing Section \ref{subsec:benchmark-correlation}, we exemplarily discuss
the correlation computed for \SEEDS with $\K \approx 400$ on the \BSDS dataset as shown
in Table \ref{table:benchmark-correlation}. We note that the following observations
can be confirmed when considering different algorithms. Still, \SEEDS has the
advantage of showing good overall performance (see the ranking in Section \ref{subsec:experiments-ranking})
and low standard deviation in \Rec, \UE and \EV. We observe a correlation of $-0.47$
between \Rec and \UE reflecting that \SEEDS exhibits high \Rec but comparably lower
\UE on the \BSDS dataset. This justifies the choice of using both \Rec and \UE for quantitative
comparison. The correlation of \UE and \ASA is $1$, which we explained with the
respective definitions. More interestingly, the correlation of \UE and \UEL is $-0.7$.
Therefore, we decided to discuss results regarding \UEL in more detail in
\ref{sec:appendix-experiments}. Nevertheless, this may also confirm the observations
by Neubert and Protzel \cite{NeubertProtzel:2012} as well as
Achanta et al. \cite{AchantaShajiSmithLucchiFuaSuesstrunk:2012} that \UEL
unjustly penalizes large superpixels.
The high correlation of $-0.97$ between \MDE and \Rec has also been explained using
the respective definitions. Interestingly, the correlation decreases with increased $\K$.
This can, however, be explained by the implementation of \Rec, allowing a fixed tolerance
of $0.0025$ times the image diagonal. The correlation of $-0.42$ between \ICV and \EV
was explained by the missing normalization of \ICV compared to~\EV. This observation
is confirmed by the decreasing correlation for larger $\K$ as, on average, superpixels
get smaller thereby reducing the influence of normalization.

\pgfplotstableset{
    col sep=comma,
    string type,
    before row=\hline,
    every last row/.style={after row=\hline},
    every row/.style={after row=\hline},
    columns/x/.style={column type=|l,column name={$\K = 400$}},
    columns/rec/.style={column type=|l,column name=\Rec},
    columns/uenp/.style={column type=|l,column name=\UE},
    columns/uelevin/.style={column type=|l,column name=\UEL},
    columns/asa/.style={column type=|l,column name=\ASA},
    columns/co/.style={column type=|l,column name=\CO},
    columns/ev/.style={column type=|l,column name=\EV},
    columns/mde/.style={column type=|l,column name=\MDE},
    columns/icv/.style={column type=|l|,column name=\ICV},
}

\begin{table}[t]
	\centering
	{\scriptsize
		\pgfplotstabletypeset[]{
			x,rec,uenp,uelevin,asa,co,ev,mde,icv
			\Rec,1,-0.47,-0.07,0.46,-0.08,0.11,-0.97,-0.04
			\UE,-0.47,1,0.11,-1,0.07,-0.19,0.47,0.25
			\UEL,-0.07,0.11,1,-0.1,-0.01,-0.03,0.09,0.05
			\ASA,0.46,-1,-0.1,1,-0.07,0.18,-0.47,-0.24
			\CO,-0.08,0.07,-0.01,-0.07,1,0.34,0.06,-0.05
			\EV,0.11,-0.19,-0.03,0.18,0.34,1,-0.16,-0.42
			\MDE,-0.97,0.47,0.09,-0.47,0.06,-0.16,1,0.06
			\ICV,-0.04,0.25,0.05,-0.24,-0.05,-0.42,0.06,1
			\CD,0.05,0,0,0,-0.86,-0.45,-0.03,0.21
		}\\[4px]
		\pgfplotstabletypeset[]{
			x,rec,uenp,uelevin,asa,co,ev,mde,icv
			\Rec,1,-0.4,-0.06,0.4,-0.11,0.16,-0.92,-0.09
			\UE,-0.4,1,0.14,-1,-0.01,-0.22,0.45,0.27
			\UEL,-0.06,0.14,1,-0.14,-0.07,-0.07,0.08,0.09
			\ASA,0.4,-1,-0.14,1,0.01,0.21,-0.45,-0.27
			\CO,-0.11,-0.01,-0.07,0.01,1,0.24,0.12,-0.15
			\EV,0.16,-0.22,-0.07,0.21,0.24,1,-0.3,-0.52
			\MDE,-0.92,0.45,0.08,-0.45,0.12,-0.3,1,0.15
			\ICV,-0.09,0.27,0.09,-0.27,-0.15,-0.52,0.15,1
			\CD,0.13,0,0.03,-0.01,-0.91,-0.31,-0.13,0.15
		}\\[4px]
		\pgfplotstabletypeset[]{
			x,rec,uenp,uelevin,asa,co,ev,mde,icv
			\Rec,1,-0.26,-0.05,0.28,-0.08,0.12,-0.66,-0.1
			\UE,-0.26,1,0.16,-1,0.02,-0.2,0.41,0.28
			\UEL,-0.05,0.16,1,-0.16,-0.02,-0.07,0.1,0.07
			\ASA,0.28,-1,-0.16,1,-0.02,0.2,-0.41,-0.28
			\CO,-0.08,0.02,-0.02,-0.02,1,0.19,0.13,-0.17
			\EV,0.12,-0.2,-0.07,0.2,0.19,1,-0.36,-0.61
			\MDE,-0.66,0.41,0.1,-0.41,0.13,-0.36,1,0.22
			\ICV,-0.1,0.28,0.07,-0.28,-0.17,-0.61,0.22,1
		}
	}
	\caption{Pearson correlation coefficient of all discussed metrics exemplarily shown for \SEEDS with $\K \approx 400$, $\K \approx 1200$ and $\K \approx 3600$.}
	\label{table:benchmark-correlation}
\end{table}

\subsection{\AvgRec, \AvgUE and \AvgEV}

As introduced in Section \ref{subsec:benchmark-average}, \ARec, \AUE and \AEV are intended to
summarize algorithm performance independent of \K. To this end, we
compute the area below the $\MR = (1 - \Rec)$, $\UE$ and $\UEV = (1 - \EV)$ curves within the interval
$[\K_{\min}, \K_{\max}] = [200, 5200]$. As introduced before, the first corresponds to the
Boundary \underline{M}iss \underline{R}ate (\MR) and the last is referred to as \underline{U}nexplained Variation (\UEV).
In particular, we use the trapezoidal rule for integration. As the algorithms do not necessarily generate the desired number
of superpixels, we additionally considered the following two cases for special treatment.
First, if an algorithm generates more that $\K_{\max}$ superpixels (or less than $\K_{\min}$),
we interpolate linearly to determine the value for $\K_{\max}$ ($\K_{\min}$).
Second, if an algorithm consistently generates less that $\K_{\max}$ (or more than $\K_{\min}$)
superpixels, we take the value lower or equal (greater or equal) and closest to $\K_{\max}$ ($\K_{\min}$).
In the second case, a superpixel algorithm is penalized if it is not able to generate
very few (\ie $\K_{\min}$) or very many (\ie $\K_{\max}$) superpixels.

	\section{Parameter Optimization}

We discuss the following two topics concerning parameter optimization in more detail:
color spaces and controlling the number of superpixels in a consistent manner.
Overall, we find that together with Section \ref{sec:parameter-optimization},
the described parameter optimization procedure ensures fair comparison as far as possible.

\subsection{Color Spaces}

The used color space inherently influences the performance of superpixel algorithms as
the majority of superpixel algorithms depend on comparing pixels within this color space.
To ensure fair comparison, we included the color space in parameter optimization.
In particular, we ensured that all algorithms support RGB color space and considered
different color spaces only if reported in the corresponding publications or
supported by the respective implementation. While some algorithms may benefit from
different color spaces not mentioned in the corresponding publications, we decided to
not consider additional color spaces for simplicity and to avoid additional overhead
during parameter optimization. Parameter optimization yielded the color spaces
highlighted in Table \ref{table:algorithms}.

\subsection{Controlling the Number of Generated Superpixels}

\DeclarePairedDelimiter\ceil{\lceil}{\rceil}
\DeclarePairedDelimiter\floor{\lfloor}{\rfloor}

Some implementations, for example \ERS and \POISE, control the number of superpixels directly
-- for example by stopping the merging of pixels as soon as the desired number of superpixels is met.
In contrast, clustering-based algorithms (except for \DASP), contour evolution algorithms, watershed-based algorithms
as well as path-based algorithms utilize a regular grid to initialize superpixels.
Some algorithms allow to adapt the grid in both horizontal and vertical direction,
while others require a Cartesian grid. We expected this difference to be reflected in
the experimental results, however, this is not the case. We standardized initialization in both cases.
	\section{Experiments}
\label{sec:appendix-experiments}

We complement Section \ref{sec:experiments} with additional experimental results.
In particular, we provide additional qualitative results to better judge the visual
quality of individual superpixel algorithms. Furthermore, we explicitly present \ASA and \UEL
on the \BSDS and \NYU datasets as well as \Rec, \UE and \EV on the \SBD, \SUNRGBD and \Fash datasets.

\subsection{Qualitative}
\label{subsec:appendix-experiments-qualitative}

\begin{figure*}
	\centering
	\vspace{-0.5cm}
	\begin{subfigure}[b]{0.02\textwidth}
		\rotatebox{90}{\small\hphantom{aaai}\W}
	\end{subfigure}
	\begin{subfigure}[b]{0.16\textwidth}
        \begin{center}
            \BSDS
        \end{center}
        \vskip -6px
		\includegraphics[height=1.65cm]{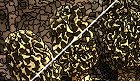}
	\end{subfigure}
	\begin{subfigure}[b]{0.129\textwidth}
        \begin{center}
            \SBD
        \end{center}
        \vskip -6px
		\includegraphics[height=1.65cm]{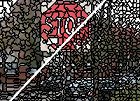}
	\end{subfigure}
	\begin{subfigure}[b]{0.10\textwidth}
        \begin{center}
            \Fash
        \end{center}
        \vskip -6px
		\includegraphics[height=1.65cm]{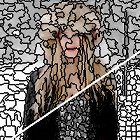}
	\end{subfigure}
	\begin{subfigure}[b]{0.02\textwidth}
		\rotatebox{90}{\small\hphantom{ai}\EAMS}
	\end{subfigure}
	\begin{subfigure}[b]{0.16\textwidth}
        \begin{center}
            \BSDS
        \end{center}
        \vskip -6px
		\includegraphics[height=1.65cm]{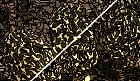}
	\end{subfigure}
	\begin{subfigure}[b]{0.129\textwidth}
        \begin{center}
            \SBD
        \end{center}
        \vskip -6px
		\includegraphics[height=1.65cm]{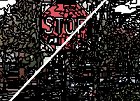}
	\end{subfigure}
	\begin{subfigure}[b]{0.10\textwidth}
        \begin{center}
            \Fash
        \end{center}
        \vskip -6px
		\includegraphics[height=1.65cm]{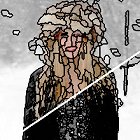}
	\end{subfigure}\\
	\begin{subfigure}[b]{0.02\textwidth}
		\rotatebox{90}{\small\hphantom{aaa}\NC}
	\end{subfigure}
	\begin{subfigure}[b]{0.16\textwidth}
		\includegraphics[height=1.65cm]{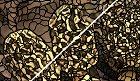}
	\end{subfigure}
	\begin{subfigure}[b]{0.129\textwidth}
		\includegraphics[height=1.65cm]{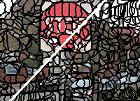}
	\end{subfigure}
	\begin{subfigure}[b]{0.10\textwidth}
		\includegraphics[height=1.65cm]{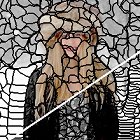}
	\end{subfigure}
	\begin{subfigure}[b]{0.02\textwidth}
		\rotatebox{90}{\small\hphantom{aaa}\FH}
	\end{subfigure}
	\begin{subfigure}[b]{0.16\textwidth}
		\includegraphics[height=1.65cm]{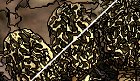}
	\end{subfigure}
	\begin{subfigure}[b]{0.129\textwidth}
		\includegraphics[height=1.65cm]{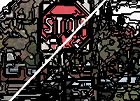}
	\end{subfigure}
	\begin{subfigure}[b]{0.10\textwidth}
		\includegraphics[height=1.65cm]{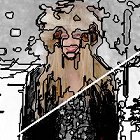}
	\end{subfigure}\\
	\begin{subfigure}[b]{0.02\textwidth}
		\rotatebox{90}{\small\hphantom{aaa}\RW}
	\end{subfigure}
	\begin{subfigure}[b]{0.16\textwidth}
		\includegraphics[height=1.65cm]{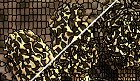}
	\end{subfigure}
	\begin{subfigure}[b]{0.129\textwidth}
		\includegraphics[height=1.65cm]{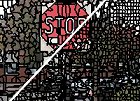}
	\end{subfigure}
	\begin{subfigure}[b]{0.10\textwidth}
		\includegraphics[height=1.65cm]{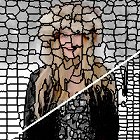}
	\end{subfigure}
	\begin{subfigure}[b]{0.02\textwidth}
		\rotatebox{90}{\small\hphantom{aaa}\QS}
	\end{subfigure}
	\begin{subfigure}[b]{0.16\textwidth}
		\includegraphics[height=1.65cm]{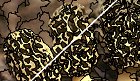}
	\end{subfigure}
	\begin{subfigure}[b]{0.129\textwidth}
		\includegraphics[height=1.65cm]{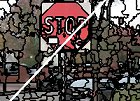}
	\end{subfigure}
	\begin{subfigure}[b]{0.10\textwidth}
		\includegraphics[height=1.65cm]{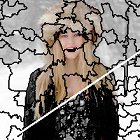}
	\end{subfigure}\\
	\begin{subfigure}[b]{0.02\textwidth}
		\rotatebox{90}{\small\hphantom{aaa}\PF}
	\end{subfigure}
	\begin{subfigure}[b]{0.16\textwidth}
		\includegraphics[height=1.65cm]{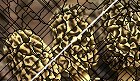}
	\end{subfigure}
	\begin{subfigure}[b]{0.129\textwidth}
		\includegraphics[height=1.65cm]{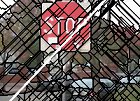}
	\end{subfigure}
	\begin{subfigure}[b]{0.10\textwidth}
		\includegraphics[height=1.65cm]{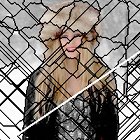}
	\end{subfigure}
	\begin{subfigure}[b]{0.02\textwidth}
		\rotatebox{90}{\small\hphantom{aaa}\TP}
	\end{subfigure}
	\begin{subfigure}[b]{0.16\textwidth}
		\includegraphics[height=1.65cm]{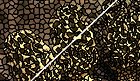}
	\end{subfigure}
	\begin{subfigure}[b]{0.129\textwidth}
		\includegraphics[height=1.65cm]{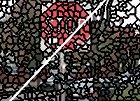}
	\end{subfigure}
	\begin{subfigure}[b]{0.10\textwidth}
		\includegraphics[height=1.65cm]{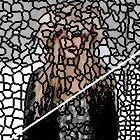}
	\end{subfigure}\\
	\begin{subfigure}[b]{0.02\textwidth}
		\rotatebox{90}{\small\hphantom{aai}\CIS}
	\end{subfigure}
	\begin{subfigure}[b]{0.16\textwidth}
		\includegraphics[height=1.65cm]{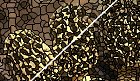}
	\end{subfigure}
	\begin{subfigure}[b]{0.129\textwidth}
		\includegraphics[height=1.65cm]{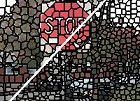}
	\end{subfigure}
	\begin{subfigure}[b]{0.10\textwidth}
		\includegraphics[height=1.65cm]{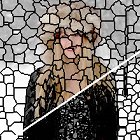}
	\end{subfigure}
	\begin{subfigure}[b]{0.02\textwidth}
		\rotatebox{90}{\small\hphantom{aa}\SLIC}
	\end{subfigure}
	\begin{subfigure}[b]{0.16\textwidth}
		\includegraphics[height=1.65cm]{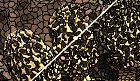}
	\end{subfigure}
	\begin{subfigure}[b]{0.129\textwidth}
		\includegraphics[height=1.65cm]{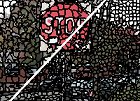}
	\end{subfigure}
	\begin{subfigure}[b]{0.10\textwidth}
		\includegraphics[height=1.65cm]{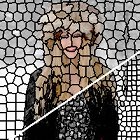}
	\end{subfigure}\\
	\begin{subfigure}[b]{0.02\textwidth}
		\rotatebox{90}{\small\hphantom{aai}\CRS}
	\end{subfigure}
	\begin{subfigure}[b]{0.16\textwidth}
		\includegraphics[height=1.65cm]{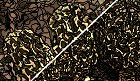}
	\end{subfigure}
	\begin{subfigure}[b]{0.129\textwidth}
		\includegraphics[height=1.65cm]{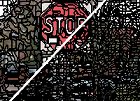}
	\end{subfigure}
	\begin{subfigure}[b]{0.10\textwidth}
		\includegraphics[height=1.65cm]{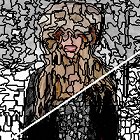}
	\end{subfigure}
	\begin{subfigure}[b]{0.02\textwidth}
		\rotatebox{90}{\small\hphantom{aai}\ERS}
	\end{subfigure}
	\begin{subfigure}[b]{0.16\textwidth}
		\includegraphics[height=1.65cm]{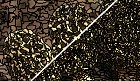}
	\end{subfigure}
	\begin{subfigure}[b]{0.129\textwidth}
		\includegraphics[height=1.65cm]{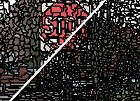}
	\end{subfigure}
	\begin{subfigure}[b]{0.10\textwidth}
		\includegraphics[height=1.65cm]{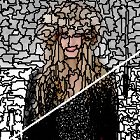}
	\end{subfigure}\\
	\begin{subfigure}[b]{0.02\textwidth}
		\rotatebox{90}{\small\hphantom{aaa}\PB}
	\end{subfigure}
	\begin{subfigure}[b]{0.16\textwidth}
		\includegraphics[height=1.65cm]{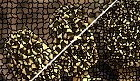}
	\end{subfigure}
	\begin{subfigure}[b]{0.129\textwidth}
		\includegraphics[height=1.65cm]{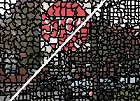}
	\end{subfigure}
	\begin{subfigure}[b]{0.10\textwidth}
		\includegraphics[height=1.65cm]{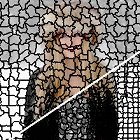}
	\end{subfigure}
	\begin{subfigure}[b]{0.02\textwidth}
		\rotatebox{90}{\small\hphantom{a}\SEEDS}
	\end{subfigure}
	\begin{subfigure}[b]{0.16\textwidth}
		\includegraphics[height=1.65cm]{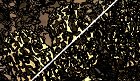}
	\end{subfigure}
	\begin{subfigure}[b]{0.129\textwidth}
		\includegraphics[height=1.65cm]{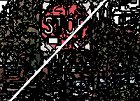}
	\end{subfigure}
	\begin{subfigure}[b]{0.10\textwidth}
		\includegraphics[height=1.65cm]{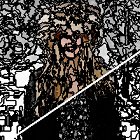}
	\end{subfigure}\\
	\begin{subfigure}[b]{0.02\textwidth}
		\rotatebox{90}{\small\hphantom{aai}\TPS}
	\end{subfigure}
	\begin{subfigure}[b]{0.16\textwidth}
		\includegraphics[height=1.65cm]{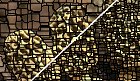}
	\end{subfigure}
	\begin{subfigure}[b]{0.129\textwidth}
		\includegraphics[height=1.65cm]{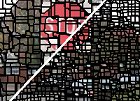}
	\end{subfigure}
	\begin{subfigure}[b]{0.10\textwidth}
		\includegraphics[height=1.65cm]{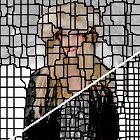}
	\end{subfigure}
	\begin{subfigure}[b]{0.02\textwidth}
		\rotatebox{90}{\small\hphantom{aaa}\VC}
	\end{subfigure}
	\begin{subfigure}[b]{0.16\textwidth}
		\includegraphics[height=1.65cm]{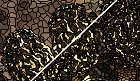}
	\end{subfigure}
	\begin{subfigure}[b]{0.129\textwidth}
		\includegraphics[height=1.65cm]{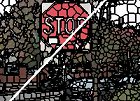}
	\end{subfigure}
	\begin{subfigure}[b]{0.10\textwidth}
		\includegraphics[height=1.65cm]{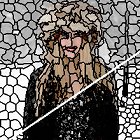}
	\end{subfigure}\\
	\begin{subfigure}[b]{0.02\textwidth}
		\rotatebox{90}{\small\hphantom{aai}\CCS}
	\end{subfigure}
	\begin{subfigure}[b]{0.16\textwidth}
		\includegraphics[height=1.65cm]{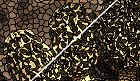}
	\end{subfigure}
	\begin{subfigure}[b]{0.129\textwidth}
		\includegraphics[height=1.65cm]{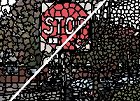}
	\end{subfigure}
	\begin{subfigure}[b]{0.10\textwidth}
		\includegraphics[height=1.65cm]{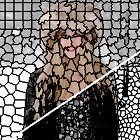}
	\end{subfigure}
	\begin{subfigure}[b]{0.02\textwidth}
		\rotatebox{90}{\small\hphantom{aaa}\CW}
	\end{subfigure}
	\begin{subfigure}[b]{0.16\textwidth}
		\includegraphics[height=1.65cm]{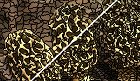}
	\end{subfigure}
	\begin{subfigure}[b]{0.129\textwidth}
		\includegraphics[height=1.65cm]{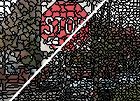}
	\end{subfigure}
	\begin{subfigure}[b]{0.10\textwidth}
		\includegraphics[height=1.65cm]{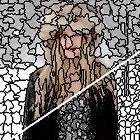}
	\end{subfigure}\\
	\begin{subfigure}[b]{0.02\textwidth}
		\rotatebox{90}{\small\hphantom{aa}\ERGC}
	\end{subfigure}
	\begin{subfigure}[b]{0.16\textwidth}
		\includegraphics[height=1.65cm]{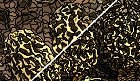}
	\end{subfigure}
	\begin{subfigure}[b]{0.129\textwidth}
		\includegraphics[height=1.65cm]{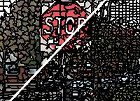}
	\end{subfigure}
	\begin{subfigure}[b]{0.10\textwidth}
		\includegraphics[height=1.65cm]{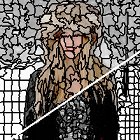}
	\end{subfigure}
	\begin{subfigure}[b]{0.02\textwidth}
		\rotatebox{90}{\small\hphantom{aai}\MSS}
	\end{subfigure}
	\begin{subfigure}[b]{0.16\textwidth}
		\includegraphics[height=1.65cm]{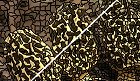}
	\end{subfigure}
	\begin{subfigure}[b]{0.129\textwidth}
		\includegraphics[height=1.65cm]{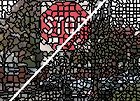}
	\end{subfigure}
	\begin{subfigure}[b]{0.10\textwidth}
		\includegraphics[height=1.65cm]{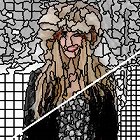}
	\end{subfigure}\\
	\begin{subfigure}[b]{0.02\textwidth}
		\rotatebox{90}{\small\hphantom{a}\preSLIC}
	\end{subfigure}
	\begin{subfigure}[b]{0.16\textwidth}
		\includegraphics[height=1.65cm]{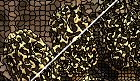}
	\end{subfigure}
	\begin{subfigure}[b]{0.129\textwidth}
		\includegraphics[height=1.65cm]{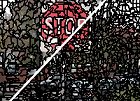}
	\end{subfigure}
	\begin{subfigure}[b]{0.10\textwidth}
		\includegraphics[height=1.65cm]{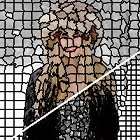}
	\end{subfigure}
	\begin{subfigure}[b]{0.02\textwidth}
		\rotatebox{90}{\small\hphantom{aaa}\WP}
	\end{subfigure}
	\begin{subfigure}[b]{0.16\textwidth}
		\includegraphics[height=1.65cm]{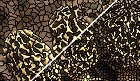}
	\end{subfigure}
	\begin{subfigure}[b]{0.129\textwidth}
		\includegraphics[height=1.65cm]{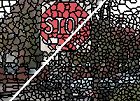}
	\end{subfigure}
	\begin{subfigure}[b]{0.10\textwidth}
		\includegraphics[height=1.65cm]{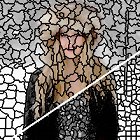}
	\end{subfigure}\\
	\begin{subfigure}[b]{0.02\textwidth}
		\rotatebox{90}{\small\hphantom{aa}\ETPS}
	\end{subfigure}
	\begin{subfigure}[b]{0.16\textwidth}
		\includegraphics[height=1.65cm]{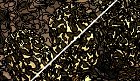}
	\end{subfigure}
	\begin{subfigure}[b]{0.129\textwidth}
		\includegraphics[height=1.65cm]{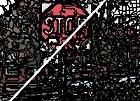}
	\end{subfigure}
	\begin{subfigure}[b]{0.10\textwidth}
		\includegraphics[height=1.65cm]{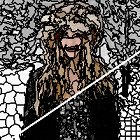}
	\end{subfigure}
	\begin{subfigure}[b]{0.02\textwidth}
		\rotatebox{90}{\small\hphantom{aai}\LSC}
	\end{subfigure}
	\begin{subfigure}[b]{0.16\textwidth}
		\includegraphics[height=1.65cm]{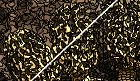}
	\end{subfigure}
	\begin{subfigure}[b]{0.129\textwidth}
		\includegraphics[height=1.65cm]{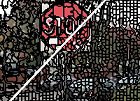}
	\end{subfigure}
	\begin{subfigure}[b]{0.10\textwidth}
		\includegraphics[height=1.65cm]{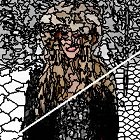}
	\end{subfigure}\\
	\begin{subfigure}[b]{0.02\textwidth}
		\rotatebox{90}{\small\hphantom{a}\POISE}
	\end{subfigure}
	\begin{subfigure}[b]{0.16\textwidth}
		\includegraphics[height=1.65cm]{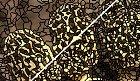}
	\end{subfigure}
	\begin{subfigure}[b]{0.129\textwidth}
		\includegraphics[height=1.65cm]{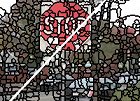}
	\end{subfigure}
	\begin{subfigure}[b]{0.10\textwidth}
		\includegraphics[height=1.65cm]{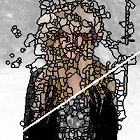}
	\end{subfigure}
	\begin{subfigure}[b]{0.02\textwidth}
		\rotatebox{90}{\small\hphantom{ai}\SEAW}
	\end{subfigure}
	\begin{subfigure}[b]{0.16\textwidth}
		\includegraphics[height=1.65cm]{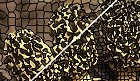}
	\end{subfigure}
	\begin{subfigure}[b]{0.129\textwidth}
		\includegraphics[height=1.65cm]{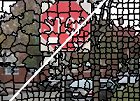}
	\end{subfigure}
	\begin{subfigure}[b]{0.10\textwidth}
		\includegraphics[height=1.65cm]{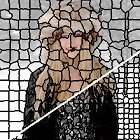}
	\end{subfigure}
	\caption{Qualitative results on the \BSDS, \SBD and \Fash datasets; excerpts from the
	images in Figure \ref{fig:datasets} are shown for $K \approx 1200$, in the upper left corner,
	and $K \approx 3600$, in the lower right corner. Superpixel boundaries are depicted in black;
	best viewed in color. We observe that with higher $\K$ both boundary adherence and
	compactness increases, even for algorithms not offering a compactness parameter.
	\textbf{Best viewed in color.}}
	\label{fig:appendix-experiments-qualitative-bsds500-sbd-fash}
\end{figure*}

\def\NYUCroppedScale{0.18}
\def\SUNRGBDCroppedScale{0.14}
\begin{figure*}
    \centering
    \begin{subfigure}[b]{0.02\textwidth}
        \rotatebox{90}{\small\hphantom{aaa}\NC}
    \end{subfigure}
    \begin{subfigure}[b]{0.1375\textwidth}
        \begin{center}
            \NYU
        \end{center}
        \vskip -6px
        \includegraphics[height=1.65cm]{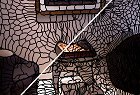}
    \end{subfigure}
    \begin{subfigure}[b]{0.129\textwidth}
        \begin{center}
            \NYU
        \end{center}
        \vskip -6px
        \includegraphics[height=1.65cm]{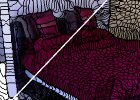}
    \end{subfigure}
    \begin{subfigure}[b]{0.02\textwidth}
        \rotatebox{90}{\small\hphantom{aaa}\RW}
    \end{subfigure}
    \begin{subfigure}[b]{0.1375\textwidth}
        \begin{center}
            \NYU
        \end{center}
        \vskip -6px
        \includegraphics[height=1.65cm]{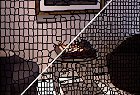}
    \end{subfigure}
    \begin{subfigure}[b]{0.129\textwidth}
        \begin{center}
            \NYU
        \end{center}
        \vskip -6px
        \includegraphics[height=1.65cm]{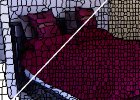}
    \end{subfigure}
    \begin{subfigure}[b]{0.02\textwidth}
        \rotatebox{90}{\small\hphantom{ai}\SEAW}
    \end{subfigure}
    \begin{subfigure}[b]{0.1375\textwidth}
        \begin{center}
            \NYU
        \end{center}
        \vskip -6px
        \includegraphics[height=1.65cm]{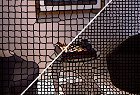}
    \end{subfigure}
    \begin{subfigure}[b]{0.129\textwidth}
        \begin{center}
            \NYU
        \end{center}
        \vskip -6px
        \includegraphics[height=1.65cm]{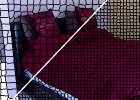}
    \end{subfigure}\\[4px]
    \begin{subfigure}[b]{0.02\textwidth}
        \rotatebox{90}{\small\hphantom{aaai}\W}
    \end{subfigure}
    \begin{subfigure}[b]{0.1375\textwidth}
        \begin{center}
			\NYU
		\end{center}
		\vskip -6px
        \includegraphics[height=1.65cm]{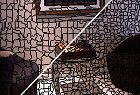}
    \end{subfigure}
    \begin{subfigure}[b]{0.129\textwidth}
        \begin{center}
			\SUNRGBD
		\end{center}
		\vskip -6px
        \includegraphics[height=1.65cm]{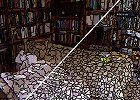}
    \end{subfigure}
    \begin{subfigure}[b]{0.02\textwidth}
        \rotatebox{90}{\small\hphantom{ai}\EAMS}
    \end{subfigure}
    \begin{subfigure}[b]{0.1375\textwidth}
        \begin{center}
			\NYU
		\end{center}
		\vskip -6px
        \includegraphics[height=1.65cm]{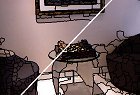}
    \end{subfigure}
    \begin{subfigure}[b]{0.129\textwidth}
        \begin{center}
			\SUNRGBD
		\end{center}
		\vskip -6px
        \includegraphics[height=1.65cm]{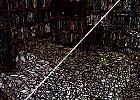}
    \end{subfigure}
    \begin{subfigure}[b]{0.02\textwidth}
        \rotatebox{90}{\small\hphantom{aaa}\FH}
    \end{subfigure}
    \begin{subfigure}[b]{0.1375\textwidth}
        \begin{center}
			\NYU
		\end{center}
		\vskip -6px
        \includegraphics[height=1.65cm]{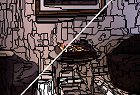}
    \end{subfigure}
    \begin{subfigure}[b]{0.129\textwidth}
        \begin{center}
			\SUNRGBD
		\end{center}
		\vskip -6px
        \includegraphics[height=1.65cm]{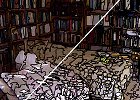}
    \end{subfigure}\\
    \begin{subfigure}[b]{0.02\textwidth}
        \rotatebox{90}{\small\hphantom{aaa}\QS}
    \end{subfigure}
    \begin{subfigure}[b]{0.1375\textwidth}
        \includegraphics[height=1.65cm]{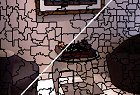}
    \end{subfigure}
    \begin{subfigure}[b]{0.129\textwidth}
        \includegraphics[height=1.65cm]{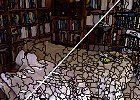}
    \end{subfigure}
    \begin{subfigure}[b]{0.02\textwidth}
        \rotatebox{90}{\small\hphantom{aaa}\PF}
    \end{subfigure}
    \begin{subfigure}[b]{0.1375\textwidth}
        \includegraphics[height=1.65cm]{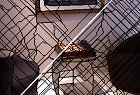}
    \end{subfigure}
    \begin{subfigure}[b]{0.129\textwidth}
        \includegraphics[height=1.65cm]{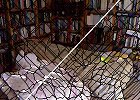}
    \end{subfigure}
    \begin{subfigure}[b]{0.02\textwidth}
        \rotatebox{90}{\small\hphantom{aaa}\TP}
    \end{subfigure}
    \begin{subfigure}[b]{0.1375\textwidth}
        \includegraphics[height=1.65cm]{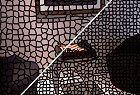}
    \end{subfigure}
    \begin{subfigure}[b]{0.129\textwidth}
        \includegraphics[height=1.65cm]{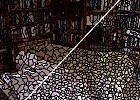}
    \end{subfigure}\\
    \begin{subfigure}[b]{0.02\textwidth}
        \rotatebox{90}{\small\hphantom{aa}\CIS}
    \end{subfigure}
    \begin{subfigure}[b]{0.1375\textwidth}
        \includegraphics[height=1.65cm]{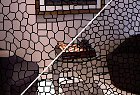}
    \end{subfigure}
    \begin{subfigure}[b]{0.129\textwidth}
        \includegraphics[height=1.65cm]{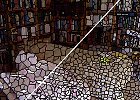}
    \end{subfigure}
    \begin{subfigure}[b]{0.02\textwidth}
        \rotatebox{90}{\small\hphantom{aa}\SLIC}
    \end{subfigure}
    \begin{subfigure}[b]{0.1375\textwidth}
        \includegraphics[height=1.65cm]{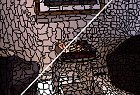}
    \end{subfigure}
    \begin{subfigure}[b]{0.129\textwidth}
        \includegraphics[height=1.65cm]{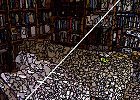}
    \end{subfigure}
    \begin{subfigure}[b]{0.02\textwidth}
        \rotatebox{90}{\small\hphantom{aai}\CRS}
    \end{subfigure}
    \begin{subfigure}[b]{0.1375\textwidth}
        \includegraphics[height=1.65cm]{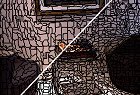}
    \end{subfigure}
    \begin{subfigure}[b]{0.129\textwidth}
        \includegraphics[height=1.65cm]{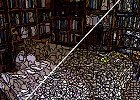}
    \end{subfigure}\\
    \begin{subfigure}[b]{0.02\textwidth}
        \rotatebox{90}{\small\hphantom{aai}\ERS}
    \end{subfigure}
    \begin{subfigure}[b]{0.1375\textwidth}
        \includegraphics[height=1.65cm]{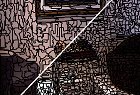}
    \end{subfigure}
    \begin{subfigure}[b]{0.129\textwidth}
        \includegraphics[height=1.65cm]{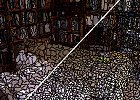}
    \end{subfigure}
    \begin{subfigure}[b]{0.02\textwidth}
        \rotatebox{90}{\small\hphantom{aaa}\PB}
    \end{subfigure}
    \begin{subfigure}[b]{0.1375\textwidth}
        \includegraphics[height=1.65cm]{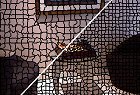}
    \end{subfigure}
    \begin{subfigure}[b]{0.129\textwidth}
        \includegraphics[height=1.65cm]{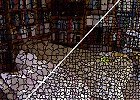}
    \end{subfigure}
    \begin{subfigure}[b]{0.02\textwidth}
        \rotatebox{90}{\small\hphantom{aa}\DASP}
    \end{subfigure}
    \begin{subfigure}[b]{0.1375\textwidth}
        \includegraphics[height=1.65cm]{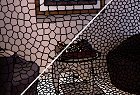}
    \end{subfigure}
    \begin{subfigure}[b]{0.129\textwidth}
        \includegraphics[height=1.65cm]{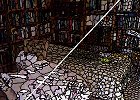}
    \end{subfigure}\\
    \begin{subfigure}[b]{0.02\textwidth}
        \rotatebox{90}{\small\hphantom{a}\SEEDS}
    \end{subfigure}
    \begin{subfigure}[b]{0.1375\textwidth}
        \includegraphics[height=1.65cm]{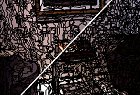}
    \end{subfigure}
    \begin{subfigure}[b]{0.129\textwidth}
        \includegraphics[height=1.65cm]{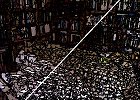}
    \end{subfigure}
    \begin{subfigure}[b]{0.02\textwidth}
        \rotatebox{90}{\small\hphantom{aai}\TPS}
    \end{subfigure}
    \begin{subfigure}[b]{0.1375\textwidth}
        \includegraphics[height=1.65cm]{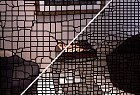}
    \end{subfigure}
    \begin{subfigure}[b]{0.129\textwidth}
        \includegraphics[height=1.65cm]{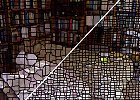}
    \end{subfigure}
    \begin{subfigure}[b]{0.02\textwidth}
        \rotatebox{90}{\small\hphantom{aaai}\VC}
    \end{subfigure}
    \begin{subfigure}[b]{0.1375\textwidth}
        \includegraphics[height=1.65cm]{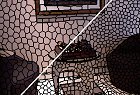}
    \end{subfigure}
    \begin{subfigure}[b]{0.129\textwidth}
        \includegraphics[height=1.65cm]{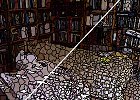}
    \end{subfigure}\\
    \begin{subfigure}[b]{0.02\textwidth}
        \rotatebox{90}{\small\hphantom{aai}\CCS}
    \end{subfigure}
    \begin{subfigure}[b]{0.1375\textwidth}
        \includegraphics[height=1.65cm]{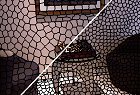}
    \end{subfigure}
    \begin{subfigure}[b]{0.129\textwidth}
        \includegraphics[height=1.65cm]{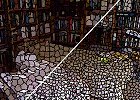}
    \end{subfigure}
    \begin{subfigure}[b]{0.02\textwidth}
        \rotatebox{90}{\small\hphantom{aa}\VCCS}
    \end{subfigure}
    \begin{subfigure}[b]{0.1375\textwidth}
        \includegraphics[height=1.65cm]{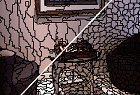}
    \end{subfigure}
    \begin{subfigure}[b]{0.129\textwidth}
        \includegraphics[height=1.65cm]{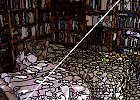}
    \end{subfigure}
    \begin{subfigure}[b]{0.02\textwidth}
        \rotatebox{90}{\small\hphantom{aaa}\CW}
    \end{subfigure}
    \begin{subfigure}[b]{0.1375\textwidth}
        \includegraphics[height=1.65cm]{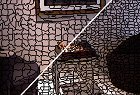}
    \end{subfigure}
    \begin{subfigure}[b]{0.129\textwidth}
        \includegraphics[height=1.65cm]{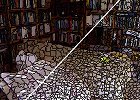}
    \end{subfigure}\\
    \begin{subfigure}[b]{0.02\textwidth}
        \rotatebox{90}{\small\hphantom{aa}\ERGC}
    \end{subfigure}
    \begin{subfigure}[b]{0.1375\textwidth}
        \includegraphics[height=1.65cm]{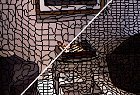}
    \end{subfigure}
    \begin{subfigure}[b]{0.129\textwidth}
        \includegraphics[height=1.65cm]{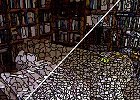}
    \end{subfigure}
    \begin{subfigure}[b]{0.02\textwidth}
        \rotatebox{90}{\small\hphantom{aai}\MSS}
    \end{subfigure}
    \begin{subfigure}[b]{0.1375\textwidth}
        \includegraphics[height=1.65cm]{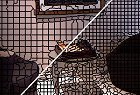}
    \end{subfigure}
    \begin{subfigure}[b]{0.129\textwidth}
        \includegraphics[height=1.65cm]{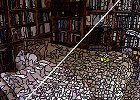}
    \end{subfigure}
    \begin{subfigure}[b]{0.02\textwidth}
        \rotatebox{90}{\small\hphantom{a}\preSLIC}
    \end{subfigure}
    \begin{subfigure}[b]{0.1375\textwidth}
        \includegraphics[height=1.65cm]{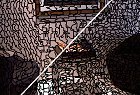}
    \end{subfigure}
    \begin{subfigure}[b]{0.129\textwidth}
        \includegraphics[height=1.65cm]{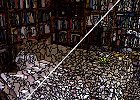}
    \end{subfigure}\\
    \begin{subfigure}[b]{0.02\textwidth}
        \rotatebox{90}{\small\hphantom{aaa}\WP}
    \end{subfigure}
    \begin{subfigure}[b]{0.1375\textwidth}
        \includegraphics[height=1.65cm]{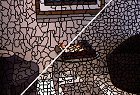}
    \end{subfigure}
    \begin{subfigure}[b]{0.129\textwidth}
        \includegraphics[height=1.65cm]{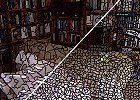}
    \end{subfigure}
    \begin{subfigure}[b]{0.02\textwidth}
        \rotatebox{90}{\small\hphantom{aa}\ETPS}
    \end{subfigure}
    \begin{subfigure}[b]{0.1375\textwidth}
        \includegraphics[height=1.65cm]{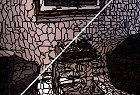}
    \end{subfigure}
    \begin{subfigure}[b]{0.129\textwidth}
        \includegraphics[height=1.65cm]{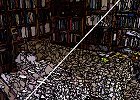}
    \end{subfigure}
    \begin{subfigure}[b]{0.02\textwidth}
        \rotatebox{90}{\small\hphantom{aaa}\LSC}
    \end{subfigure}
    \begin{subfigure}[b]{0.1375\textwidth}
        \includegraphics[height=1.65cm]{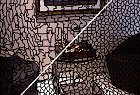}
    \end{subfigure}
    \begin{subfigure}[b]{0.129\textwidth}
        \includegraphics[height=1.65cm]{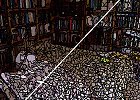}
    \end{subfigure}\\
    \begin{subfigure}[b]{0.02\textwidth}
        \rotatebox{90}{\small\hphantom{a}\POISE}
    \end{subfigure}
    \begin{subfigure}[b]{0.1375\textwidth}
        \includegraphics[height=1.65cm]{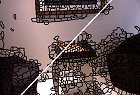}
    \end{subfigure}
    \begin{subfigure}[b]{0.129\textwidth}
        \includegraphics[height=1.65cm]{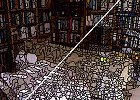}
    \end{subfigure}
    \begin{subfigure}[b]{0.02\textwidth}
        \hphantom{aa}
    \end{subfigure}
    \begin{subfigure}[b]{0.1375\textwidth}
        \hphantom{aaaaaaaaaaaaaaaaaaaaaaaaaaaaaii}
    \end{subfigure}
    \begin{subfigure}[b]{0.129\textwidth}
        \hphantom{aaaaaaaaaaaaaaaaaaaaaaaaaaa}
    \end{subfigure}
    \begin{subfigure}[b]{0.02\textwidth}
        \hphantom{aa}
    \end{subfigure}
    \begin{subfigure}[b]{0.1375\textwidth}
        \hphantom{aaaaaaaaaaaaaaaaaaaaaaaaaaaaii}
    \end{subfigure}
    \begin{subfigure}[b]{0.129\textwidth}
        \hphantom{aaaaaaaaaaaaaaaaaaaaaaaaaaa}
    \end{subfigure}
	\caption{Qualitative results on the \NYU and \SUNRGBD datasets; excerpts from the images
	in Figure \ref{fig:datasets} are shown for $K \approx 1200$, in the upper left corner,
	and $K \approx 3600$, in the lower right corner. Superpixel boundaries are depicted
	in black; best viewed in color.
    \NC, \RW and \SEAW could not be evaluated on the \SUNRGBD dataset due to
    exhaustive memory usage of the corresponding MatLab implementations.
    Therefore, results on the \NYU dataset are shown.
	\textbf{Best viewed in color.}
	}
	\label{fig:appendix-experiments-qualitative-nyuv2-sunrgbd}
\end{figure*}

\begin{figure}
	\centering
	\begin{subfigure}[b]{0.02\textwidth}
		\rotatebox{90}{\small\hphantom{aa}\SLIC}
	\end{subfigure}
	\begin{subfigure}[b]{0.141\textwidth}
		\includegraphics[height=1.525cm]{pictures/compactness/bsds500/slic/score/1/cropped/slic_35028_contours}
	\end{subfigure}
	\begin{subfigure}[b]{0.141\textwidth}
		\includegraphics[height=1.525cm]{pictures/compactness/bsds500/slic/score/10/cropped/slic_35028_contours}
	\end{subfigure}
	\begin{subfigure}[b]{0.141\textwidth}
		\includegraphics[height=1.525cm]{pictures/compactness/bsds500/slic/score/80/cropped/slic_35028_contours}
	\end{subfigure}\\
	\begin{subfigure}[b]{0.02\textwidth}
		\rotatebox{90}{\small\hphantom{a}\vlSLIC}
	\end{subfigure}
	\begin{subfigure}[b]{0.141\textwidth}
		\includegraphics[height=1.525cm]{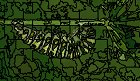}
	\end{subfigure}
	\begin{subfigure}[b]{0.141\textwidth}
		\includegraphics[height=1.525cm]{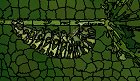}
	\end{subfigure}
	\begin{subfigure}[b]{0.141\textwidth}
		\includegraphics[height=1.525cm]{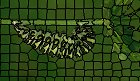}
	\end{subfigure}\\
	\begin{subfigure}[b]{0.02\textwidth}
		\rotatebox{90}{\small\hphantom{aai}\CRS}
	\end{subfigure}
	\begin{subfigure}[b]{0.141\textwidth}
		\includegraphics[height=1.525cm]{pictures/compactness/bsds500/crs/score/0.001/cropped/crs_35028_contours}
	\end{subfigure}
	\begin{subfigure}[b]{0.141\textwidth}
		\includegraphics[height=1.525cm]{pictures/compactness/bsds500/crs/score/0.01/cropped/crs_35028_contours}
	\end{subfigure}
	\begin{subfigure}[b]{0.141\textwidth}
		\includegraphics[height=1.525cm]{pictures/compactness/bsds500/crs/score/0.1/cropped/crs_35028_contours}
	\end{subfigure}\\
	\begin{subfigure}[b]{0.02\textwidth}
		\rotatebox{90}{\small\hphantom{a}\reSEEDS}
	\end{subfigure}
	\begin{subfigure}[b]{0.141\textwidth}
		\includegraphics[height=1.525cm]{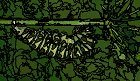}
	\end{subfigure}
	\begin{subfigure}[b]{0.141\textwidth}
		\includegraphics[height=1.525cm]{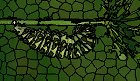}
	\end{subfigure}
	\begin{subfigure}[b]{0.141\textwidth}
		\includegraphics[height=1.525cm]{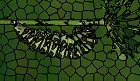}
	\end{subfigure}\\
	\begin{subfigure}[b]{0.02\textwidth}
		\rotatebox{90}{\small\hphantom{aaa}\VC}
	\end{subfigure}
	\begin{subfigure}[b]{0.141\textwidth}
		\includegraphics[height=1.525cm]{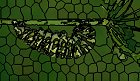}
	\end{subfigure}
	\begin{subfigure}[b]{0.141\textwidth}
		\includegraphics[height=1.525cm]{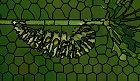}
	\end{subfigure}
	\begin{subfigure}[b]{0.141\textwidth}
		\includegraphics[height=1.525cm]{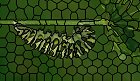}
	\end{subfigure}\\
	\begin{subfigure}[b]{0.02\textwidth}
		\rotatebox{90}{\small\hphantom{aai}\CCS}
	\end{subfigure}
	\begin{subfigure}[b]{0.141\textwidth}
		\includegraphics[height=1.525cm]{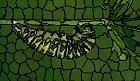}
	\end{subfigure}
	\begin{subfigure}[b]{0.141\textwidth}
		\includegraphics[height=1.525cm]{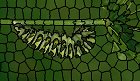}
	\end{subfigure}
	\begin{subfigure}[b]{0.141\textwidth}
		\includegraphics[height=1.525cm]{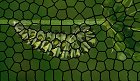}
	\end{subfigure}\\
	\begin{subfigure}[b]{0.02\textwidth}
		\rotatebox{90}{\small\hphantom{aaa}\CW}
	\end{subfigure}
	\begin{subfigure}[b]{0.141\textwidth}
		\includegraphics[height=1.525cm]{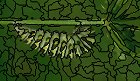}
	\end{subfigure}
	\begin{subfigure}[b]{0.141\textwidth}
		\includegraphics[height=1.525cm]{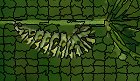}
	\end{subfigure}
	\begin{subfigure}[b]{0.141\textwidth}
		\includegraphics[height=1.525cm]{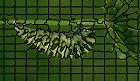}
	\end{subfigure}\\
	\begin{subfigure}[b]{0.02\textwidth}
		\rotatebox{90}{\small\hphantom{a}\preSLIC}
	\end{subfigure}
	\begin{subfigure}[b]{0.141\textwidth}
		\includegraphics[height=1.525cm]{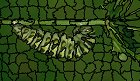}
	\end{subfigure}
	\begin{subfigure}[b]{0.141\textwidth}
		\includegraphics[height=1.525cm]{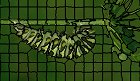}
	\end{subfigure}
	\begin{subfigure}[b]{0.141\textwidth}
		\includegraphics[height=1.525cm]{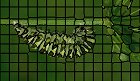}
	\end{subfigure}\\
	\begin{subfigure}[b]{0.02\textwidth}
		\rotatebox{90}{\small\hphantom{aaa}\WP}
	\end{subfigure}
	\begin{subfigure}[b]{0.141\textwidth}
		\includegraphics[height=1.525cm]{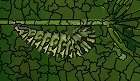}
	\end{subfigure}
	\begin{subfigure}[b]{0.141\textwidth}
		\includegraphics[height=1.525cm]{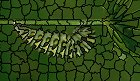}
	\end{subfigure}
	\begin{subfigure}[b]{0.141\textwidth}
		\includegraphics[height=1.525cm]{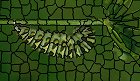}
	\end{subfigure}\\
	\begin{subfigure}[b]{0.02\textwidth}
		\rotatebox{90}{\small\hphantom{aa}\ERGC}
	\end{subfigure}
	\begin{subfigure}[b]{0.141\textwidth}
		\includegraphics[height=1.525cm]{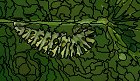}
	\end{subfigure}
	\begin{subfigure}[b]{0.141\textwidth}
		\includegraphics[height=1.525cm]{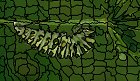}
	\end{subfigure}
	\begin{subfigure}[b]{0.141\textwidth}
		\includegraphics[height=1.525cm]{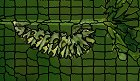}
	\end{subfigure}\\
	\begin{subfigure}[b]{0.02\textwidth}
		\rotatebox{90}{\small\hphantom{aai}\LSC}
	\end{subfigure}
	\begin{subfigure}[b]{0.141\textwidth}
		\includegraphics[height=1.525cm]{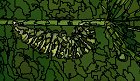}
	\end{subfigure}
	\begin{subfigure}[b]{0.141\textwidth}
		\includegraphics[height=1.525cm]{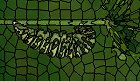}
	\end{subfigure}
	\begin{subfigure}[b]{0.141\textwidth}
		\includegraphics[height=1.525cm]{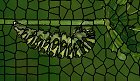}
	\end{subfigure}\\
	\begin{subfigure}[b]{0.02\textwidth}
		\rotatebox{90}{\small\hphantom{aa}\ETPS}
	\end{subfigure}
	\begin{subfigure}[b]{0.141\textwidth}
		\includegraphics[height=1.525cm]{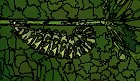}
	\end{subfigure}
	\begin{subfigure}[b]{0.141\textwidth}
		\includegraphics[height=1.525cm]{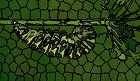}
	\end{subfigure}
	\begin{subfigure}[b]{0.141\textwidth}
		\includegraphics[height=1.525cm]{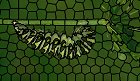}
	\end{subfigure}
	\caption{The influence of a low, on the left, and high, on the right, compactness parameter
	demonstrated on the caterpillar image from the \BSDS dataset for $\K \approx 400$.
	Superpixel boundaries are depicted in black; best viewed in color. For all shown algorithms,
	the compactness parameter allows to gradually trade boundary adherence for compactness.
	\textbf{Best viewed in color.}}
	\label{fig:appendix-experiments-qualitative-compactness}
\end{figure}

We briefly discuss visual quality on additional examples provided in Figures
\ref{fig:appendix-experiments-qualitative-bsds500-sbd-fash} and \ref{fig:appendix-experiments-qualitative-nyuv2-sunrgbd}.
Additionally, Figure \ref{fig:appendix-experiments-qualitative-compactness} shows
the influence of the compactness parameter on superpixel algorithms not discussed
in Section \ref{subsec:experiments-qualitative}.

Most algorithms exhibit good boundary adherence, especially for large \K.
In contrast to the discussion in Section \ref{subsec:experiments-qualitative} focussing
on qualitative results with $\K \approx 400$ and $\K \approx 1200$, Figures
\ref{fig:appendix-experiments-qualitative-bsds500-sbd-fash} and \ref{fig:appendix-experiments-qualitative-nyuv2-sunrgbd}
also show results for $\K \approx 3600$. We observe that with rising \K,
most algorithms exhibit better boundary adherence. Exceptions are, again, easily
identified: \FH, \QS, \CIS, \PF, \PB, \TPS and \SEAW. Still, due to higher \K,
the effect of missed image boundaries is not as serious as with less superpixels.
Overall, the remaining algorithms show good boundary adherence, especially for high \K.

Compactness increases with higher \K; still, a compactness parameter is beneficial.
While for higher \K, superpixels tend to be more compact in general, the
influence of parameter optimization with respect to \Rec and \UE is still visible
-- also for algorithms providing a compactness parameter.
For example, \ERGC or \ETPS exhibit more irregular superpixels compared to \SLIC or \CCS.
Complementing this discussion, Figure \ref{fig:appendix-experiments-qualitative-compactness}
shows the influence of the compactness parameter for the algorithms with compactness
parameter not discussed in detail in Section \ref{subsec:experiments-qualitative}.
It can be seen, that a compactness parameter allows to gradually trade boundary
adherence for compactness in all of the presented cases.
However, higher \K also induces higher compactness for algorithms not providing a compactness parameter
such as \CIS, \RW, \W or \MSS to name only a few examples.
Overall, compactness benefits from higher~\K.

Overall, higher \K induces both better boundary adherence and higher compactness independent of whether a
compactness parameter is involved.

\subsection{Quantitative}
\label{subsec:appendix-experiments-quantitative}

\begin{figure*}
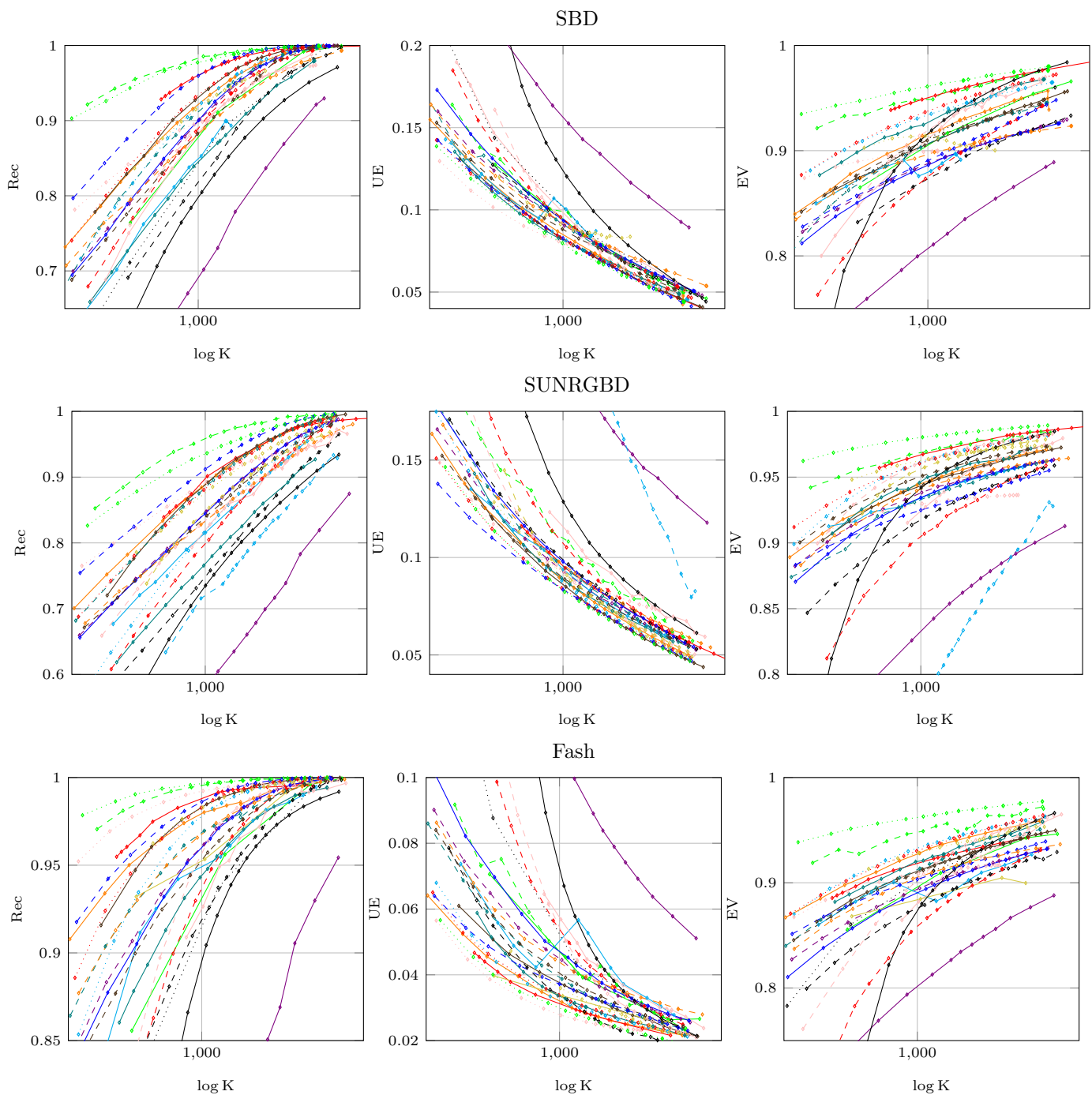

	\centering
	\input{plots/appendix-sbd-avg}
	\input{plots/appendix-sunrgbd-avg}
	\input{plots/appendix-fash-avg}
	\caption{\Rec, \UE and \EV on the \SBD, \SUNRGBD and \Fash datasets. Similar to the results
	presented for the \BSDS and \NYU datasets (compare Figures \ref{fig:experiments-quantitative-bsds500}
	and \ref{fig:experiments-quantitative-nyuv2}), \Rec and \UE give a roguh overview of
	algorithm performance with respect to ground truth. Concerning \Rec, we observe
	similar performance across the three datasets, while algorithms may show different
	behavior with respect to \UE. Similarly, \EV gives a ground truth independent
	overview of algorithm performance where algorithms show similar performance across datasets.
	\textbf{Best viewed in color.}}
	\label{fig:appendix-experiments-sbd-sunrgbd-fash}
	\vskip 12px
	{\scriptsize
		\begin{tabularx}{\textwidth}{X X X X X X X X X l}
			\ref{plot:w} \W &
			\ref{plot:eams} \EAMS &
			\ref{plot:nc} \NC &
			\ref{plot:fh} \FH &
			\ref{plot:rw} \RW &
			\ref{plot:qs} \QS &
			\ref{plot:pf} \PF &
			\ref{plot:tp} \TP &
			\ref{plot:cis} \CIS \\
			\ref{plot:slic} \SLIC &
			\ref{plot:crs} \CRS &
			\ref{plot:ers} \ERS &
			\ref{plot:pb} \PB &
			\ref{plot:dasp} \DASP &
			\ref{plot:seeds} \SEEDS &
			\ref{plot:tps} \TPS &
			\ref{plot:vc} \VC &
			\ref{plot:ccs} \CCS \\
			\ref{plot:vccs} \VCCS &
			\ref{plot:cw} \CW &
			\ref{plot:ergc} \ERGC &
			\ref{plot:mss} \MSS &
			\ref{plot:preslic} \preSLIC &
			\ref{plot:wp} \WP &
			\ref{plot:etps} \ETPS &
			\ref{plot:lsc} \LSC &
			\ref{plot:poise} \POISE \\
			\ref{plot:seaw} \SEAW & & & & & & & &
		\end{tabularx}
	}

\end{figure*}
\begin{figure*}
	\centering
	\input{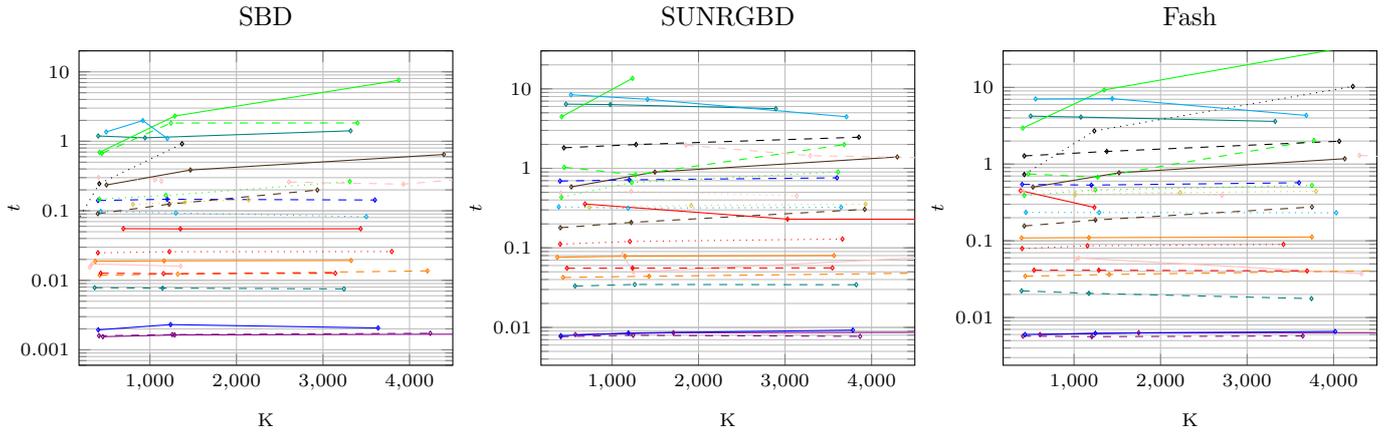}
	\caption{Runtime in seconds $t$ on the \SBD, \SUNRGBD and \Fash datasets. The results allow
	to get an impression of how runtime of individual algorithms scales with the size
	of the image. In particular, we deduce that most algorithm's runtime scales linear
	in the input size, while the number of generated superpixels does have little influence.
	\textbf{Best viewed in color.}}
	\label{fig:appendix-experiments-runtime}
\end{figure*}
\begin{figure*}
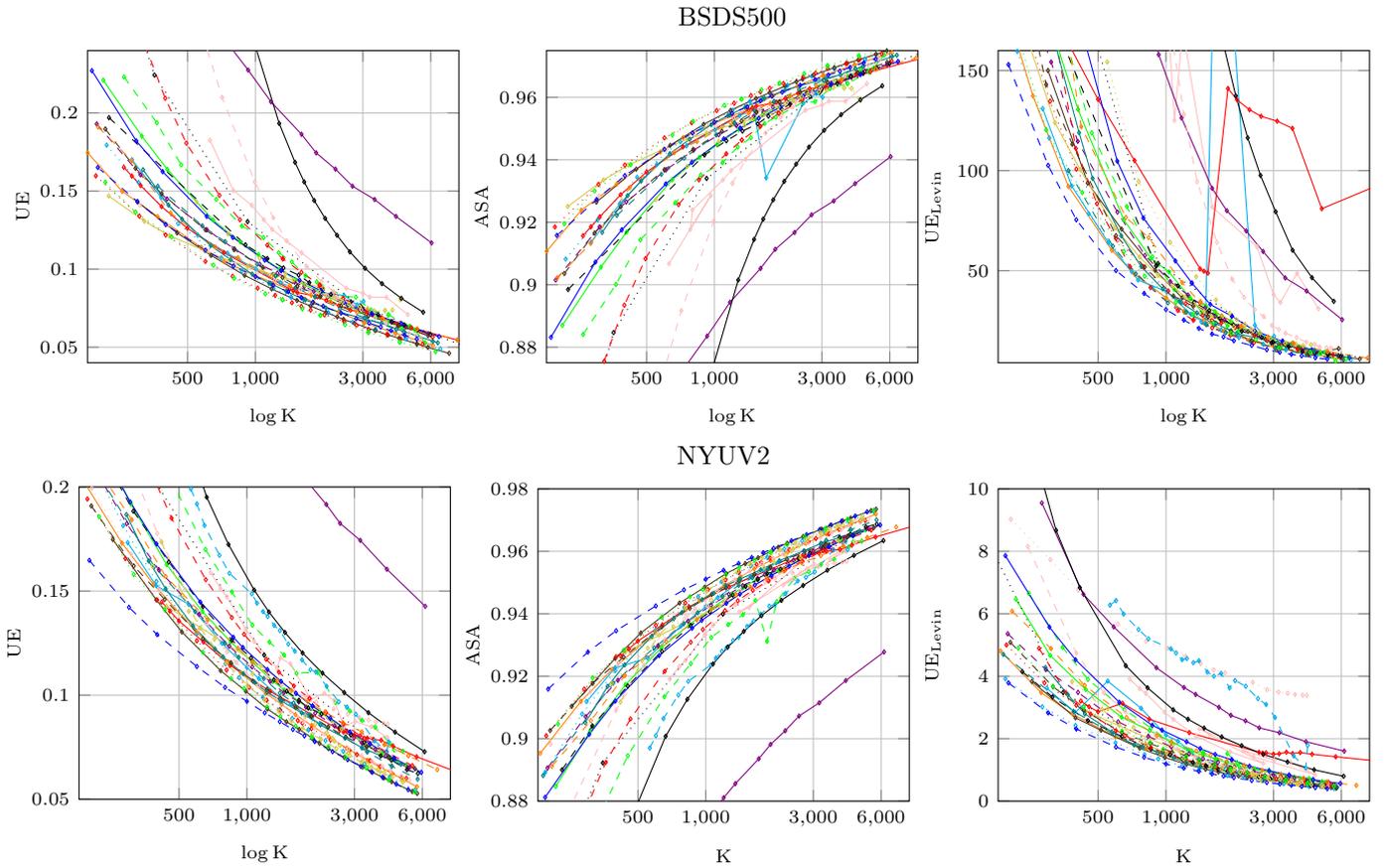

	\centering
	\input{plots/appendix-bsds500-ue-asa-uel}\\
	\input{plots/appendix-nyuv2-ue-asa-uel}
	\caption{\UE, \ASA and \UEL on the \BSDS and \NYU datasets. We find that \ASA does not
	provide new insights compared to \UE, as it closely reflects $(1 - \UE)$ except
	for a minor absolute offset. \UEL, in contrast, provides a different point view
	compared to \UE. However, \UEL is harder to interpret and strongly varies across datasets.
	\textbf{Best viewed in color.}}
	\label{fig:appendix-experiments-bsds500-nyuv2}
	\vskip 12px
	{\scriptsize
		\begin{tabularx}{\textwidth}{X X X X X X X X X l}
			\ref{plot:w} \W &
			\ref{plot:eams} \EAMS &
			\ref{plot:nc} \NC &
			\ref{plot:fh} \FH &
			\ref{plot:rw} \RW &
			\ref{plot:qs} \QS &
			\ref{plot:pf} \PF &
			\ref{plot:tp} \TP &
			\ref{plot:cis} \CIS \\
			\ref{plot:slic} \SLIC &
			\ref{plot:crs} \CRS &
			\ref{plot:ers} \ERS &
			\ref{plot:pb} \PB &
			\ref{plot:dasp} \DASP &
			\ref{plot:seeds} \SEEDS &
			\ref{plot:tps} \TPS &
			\ref{plot:vc} \VC &
			\ref{plot:ccs} \CCS \\
			\ref{plot:vccs} \VCCS &
			\ref{plot:cw} \CW &
			\ref{plot:ergc} \ERGC &
			\ref{plot:mss} \MSS &
			\ref{plot:preslic} \preSLIC &
			\ref{plot:wp} \WP &
			\ref{plot:etps} \ETPS &
			\ref{plot:lsc} \LSC &
			\ref{plot:poise} \POISE \\
			\ref{plot:seaw} \SEAW & & & & & & & &
		\end{tabularx}
	}

\end{figure*}

The following experiments complement the discussion in Section \ref{subsec:experiments-quantitative}
in two regards. First, we present additional experiments considering both \ASA and
\UEL on the \BSDS and \NYU datasets. Then, we consider \Rec, \UE and \EV in more details for
the remaining datasets, \ie the \SBD, \SUNRGBD and \Fash datasets. We begin by discussing \ASA
and \UEL, also in regard to the observations made in Sections \ref{subsec:benchmark-correlation} and \ref{sec:appendix-benchmark}.

As observed on the \BSDS and \NYU datasets in Section \ref{subsec:experiments-quantitative},
\Rec and \UE can be used to roughly asses superpixel algorithms based on ground
truth. However, for large \K, these metrics are not necessarily
sufficient to discriminate between the superpixel algorithms. Considering Figure
\ref{fig:appendix-experiments-sbd-sunrgbd-fash}, in particular with regard to
\Rec, we can identify algorithms showing above-average performance such as \ETPS
and \SEEDS. These algorithms perform well on all three datasets. Similarly,
\PF, \QS, \SEAW and \TPS perform poorly on all three datasets. Regarding \UE,
in contrast, top-performer across all three algorithms are not identified as easily.
For example, \POISE demonstrates low \UE on the \SBD and \Fash datasets, while performing
poorly on the \SUNRGBD dataset. Similarly, \ERS shows excellent performance on the \SUNRGBD dataset,
while being outperformed by \POISE as well as \ETPS on the \SBD and \Fash datasets.
Overall, \Rec and \UE do not necessarily give a consistent
view on the performance of the superpixel algorithms across datasets. This may also
be explained by the ground truth quality as already discussed in Section \ref{subsec:experiments-quantitative}.

The above observations also justify the use of \EV to judge superpixel algorithms
independent of ground truth. Considering Figure \ref{fig:appendix-experiments-sbd-sunrgbd-fash},
in particular, with regard to \EV, we can observe a more consistent view across the datasets.
Both, top-performing algorithms such as \ETPS and \SEEDS, as well as poorly performing
algorithms such as \PF, \PB or \TPS can easily be identified. In between these two extremes,
superpixel algorithms are easier to discriminate compared to \Rec and \UE. Furthermore,
some superpixel algorithms such as \QS, \FH or \CIS are performing better compared to
\Rec or \UE. This confirms the observations that ground truth independent assessment
is beneficial but cannot replace \Rec or \UE.

We find that \ASA closely mimicks the behavior of $(1 - \UE)$ while \UEL may
complement our discussion with an additional viewpoint which is, however, hard to interpret.
We consider Figure \ref{fig:appendix-experiments-bsds500-nyuv2}
showing \UE, \ASA and \UEL for both the \BSDS and \NYU datasets. Focussing on \UE and
\ASA, we easily see that \ASA nearly reflects $(1 - \UE)$ while being a small
constant off. In particular, all algorithms exhibit nearly the same behavior,
while absolutely the algorithms show higher \ASA
compared to $(1 - \UE)$. This demonstrates that \ASA does not give new insights
with respect to the quantitative comparison of superpixel algorithms. In contrast,
the algorithms show different behavior considering \UEL. This is mainly due to the
unconstrained range of \UEL (compared to $\UE \in [0,1]$). In particular,
for algorithms such as \EAMS and \FH, \UEL reflects the behavior of $\max\UE$
as shown in Figure \ref{subfig:experiments-quantitative-bsds500-ue_np.max_max}.
The remaining algorithms lie more closely together.
Still, algorithms such as \ERS, \SEEDS or \PB show better \UEL than \UE (seen relatively to the remaining algorithms).
In the case of \EAMS and \FH, high \UEL may indeed
be explained by the considerations of Neubert and Protzel \cite{NeubertProtzel:2012}
arguing that \UEL unjustly penalizes large superpixels slightly overlapping with multiple ground truth segments.
For the remaining algorithms,
the same argument can only be applied in smaller scale as these algorithms usually
do not generate large superpixels. In this line of throught, the excellent
performance of \ERS may be explained by the employed regularizer for enforcing
uniform superpixel size. Overall, \ASA does not contribute to an insightful discussion,
while \UEL may be considered in addition to \UE to complete the picture of algorithm performance.

\subsection{Runtime}
\label{subsec:appendix-experiments-robustness}

We briefly discuss runtime on the \SBD, \SUNRGBD and \Fash datasets allowing to get
more insights on how the algorithms scale with respect to image size and the
number of generated superpixels.

We find that the runtime of most algorithms scales roughly linear in the input size, while the
number of generated superpixels has little influence. We first remember that
the average image size of the \SBD, \SUNRGBD and \Fash datasets is: $314 \times 242 = 75988$,
$660 \times 488 = 322080$ and $400 \times 600 = 240000$. For $\K \approx 400$, \W
runs in roughly $1.9\text{ms}$ and $7.9\text{ms}$ on the \SBD and \SUNRGBD datasets, respectively.
As the input size for the \SUNRGBD dataset is roughly $4.24$ times larger compared to the
\SBD dataset, this results in roughly linear scaling of runtime with respect to the input size.
Similar reasoning can be applied to most of the remaining algorithms, especially
fast algorithms such as \CW, \PF, \preSLIC, \MSS or \SLIC. Except for \RW, \QS and \SEAW
we also notice that the number of generated superpixels does not influence runtime significantly.
Overall, the results confirm the claim of many authors that algorithms scale
linear in the input size, while the number of generated superpixels has little influence.

\end{appendix}
\end{document}